\documentclass[conference]{IEEEtran}
\IEEEoverridecommandlockouts
\usepackage{amsmath,amsfonts}
\usepackage{array}
\usepackage{amsthm}
\usepackage{textcomp}
\usepackage{stfloats} 
\usepackage{url}
\usepackage{verbatim}
\usepackage[table]{xcolor}
\usepackage{xcolor}
\usepackage{graphicx}
\usepackage[tight,small]{subfigure}
\usepackage{cite}
\usepackage{bm}   
\usepackage{bbm}
\usepackage{amsthm,amsmath,amssymb,amsfonts}
\usepackage{mathrsfs}
\usepackage{newunicodechar}
\usepackage[linesnumbered,ruled,vlined]{algorithm2e}
\usepackage{algpseudocode}
\usepackage{enumitem}
\usepackage{adjustbox}
\newtheorem{remark}{Remark}
\newtheorem{assumption}{Assumption}
\newtheorem{theorem}{Theorem}
\newtheorem{lemma}{Lemma}
\newtheorem{corollary}{Corollary}
\newtheorem{proposition}{Proposition}
\newtheorem{definition}{Definition}
\usepackage{ragged2e} 
\usepackage{booktabs,makecell, multirow, tabularx}
\usepackage{graphicx}
\definecolor{verylightgray}{rgb}{0.5, 0.6, 0.85}
\definecolor{veryverylightblue}{rgb}{0.765, 0.824, 0.918}
\usepackage{caption}
\usepackage{fancyhdr}
\usepackage[margin=0.65in, bottom=0.98in]{geometry}

\def\BibTeX{{\rm B\kern-.05em{\sc i\kern-.025em b}\kern-.08em
		T\kern-.1667em\lower.7ex\hbox{E}\kern-.125emX}}
\algnewcommand{\LineComment}[1]{\State \(\triangleright\) #1}
\def\BibTeX{{\rm B\kern-.05em{\sc i\kern-.025em b}\kern-.08em
		T\kern-.1667em\lower.7ex\hbox{E}\kern-.125emX}}
\graphicspath{{Figures/}}
\begin{document}

\title{HEART: Achieving Timely Multi-Model Training for Vehicle-Edge-Cloud-Integrated Hierarchical Federated Learning}
\author{
	\IEEEauthorblockN{
		Xiaohong Yang,
		Minghui Liwang, \textit{Senior Member}, \textit{IEEE},
            Xianbin Wang, \textit{Fellow}, \textit{IEEE}, 
		Zhipeng Cheng, \\ 
            Seyyedali Hosseinalipour, \textit{Senior Member}, \textit{IEEE}, 
		Huaiyu Dai, \textit{Fellow}, \textit{IEEE}, Zhenzhen Jiao
	}   
    \thanks{This work was supported in part by Shanghai Pujiang Programme under Grant no. 24PJD117; the National Natural Science Foundation of China under Grant nos. 62271424, 62088101, 72171172, 92367101; Shanghai Municipal Science and Technology Major Project under Grant no. 2021SHZDZX0100; the Chinese Academy of Engineering, Strategic Research and Consulting Program under Grant no. 2023-XZ-65; U.S. National Science Foundation (NSF) under Grant nos. ECCS-2512911ZX, ECCS-2203214; Chinese Academy of Engineering, Strategic Research and Consulting Projects, under Grant 2025-DFZD-27; Shanghai ``Explorer Plan" project under Grant 24TS1400900. Xiaohong Yang (xiaohongyang@stu.xmu.edu.cn) is with the School of Informatics, Xiamen University, Fujian, China. Minghui Liwang (minghuiliwang@tongji.edu.cn) is with the Department of Control Science and Engineering, Shanghai Institute of Intelligent Science and Technology, the State Key Laboratory of Autonomous Intelligent Unmanned Systems, Shanghai Key Laboratory of Intelligent Autonomous Systems, and Frontiers Science Center for Intelligent Autonomous Systems, Ministry of Education, Tongji University, Shanghai, China. Xianbin Wang (xianbin.wang@uwo.ca) is with the Department of Electrical and Computer Engineering, Western University, Ontario, Canada. Zhipeng Cheng (chengzp\_x@163.com) is with the School of Future Science and Engineering, Soochow University, Suzhou, China. Seyyedali Hosseinalipour (alipour@buffalo.edu) is with the Department of Electrical Engineering, University at Buffalo-SUNY, Buffalo, NY USA. Huaiyu Dai (hdai@ncsu.edu) is with the Department of Electrical and Computer Engineering, NC State University, Raleigh, NC, USA. Zhenzhen Jiao (jiaozhenzhen@gmail.com) is with Aeromind Technology Co., Ltd., Shenzhen, China. Corresponding author: Minghui Liwang.}
}
\maketitle


\begin{abstract}
The rapid growth of AI-enabled Internet of Vehicles (IoV) calls for efficient Machine Learning (ML) solutions that can handle high vehicular mobility and decentralized data. This has motivated the emergence of Hierarchical Federated Learning over vehicle-edge-cloud architectures (VEC-HFL). Nevertheless, one aspect which is underexplored in the literature on VEC-HFL is that vehicles often need to execute multiple ML tasks simultaneously, where this multi-model training environment introduces crucial challenges. First, improper aggregation rules can lead to model obsolescence and prolonged training times. Second, vehicular mobility may result in inefficient data utilization by preventing the vehicles from returning their models to the network edge. Third, achieving a balanced resource allocation across diverse tasks becomes of paramount importance as it majorly affects the effectiveness of collaborative training. We take one of the first steps towards addressing these challenges via proposing a framework for multi-model training in dynamic VEC-HFL with the goal of minimizing global training latency while ensuring balanced training across various tasks, a problem that turns out to be NP-hard. To facilitate timely model training, we introduce a hybrid synchronous-asynchronous aggregation rule. Building on this, we present a novel method called Hybrid Evolutionary And gReedy allocaTion (HEART). The framework operates in two stages: first, it achieves balanced task scheduling through a hybrid heuristic approach that combines improved Particle Swarm Optimization (PSO) and Genetic Algorithms (GA); second, it employs a low-complexity greedy algorithm to determine the training priority of assigned tasks on vehicles. Experiments on real-world datasets demonstrate the superiority of HEART over existing methods.
\end{abstract}


\section{Introduction}
\noindent \IEEEPARstart{T}{he} rapid advancement of Artificial Intelligence (AI), communication and computing technologies, and evolving vehicle design requirements has enabled the integration of Machine Learning (ML)-driven applications in the Internet of Vehicles (IoV) \cite{SplitLearning,JointClient}, improving vehicular services and driving experiences. However, the distributed data and high mobility in IoV significantly hinder efficient ML training, as conventional methods rely on centralized paradigms that require uploading raw data from vehicles to a central cloud server, leading to high latency, heavy communication overhead, and privacy risks. Federated Learning (FL) mitigates these issues by enabling distributed training at data sources, thus preserving privacy and reducing communication costs \cite{AdaptiveTraining, GHPFL,FedCDac}. However, in highly dynamic vehicular environments, FL still suffers from inefficient communication and frequent failures due to vehicle mobility and unstable connectivity \cite{HierFedML,TranWGAN,DeviceS}. To address these limitations, Hierarchical Federated Learning (HFL) has been proposed as a multi-layer extension of FL \cite{11202380, ConvergenceL, WirelessP}, introducing edge servers as intermediate aggregators between vehicles and the cloud. This hierarchical design improves communication efficiency and system robustness, making HFL a promising solution for ML training in IoV systems.

\subsection{Investigations on the Integration of FL and HFL within the IoV Ecosystem}
Recent investigations of conventional FL and HFL over IoV systems predominantly emphasize single-task/model FL \cite{MOBFL,DistributedP,EfficientVehicle}, where all vehicles are engaged in training a single ML model. However, in real-world vehicular networks, vehicles often need to execute multiple ML tasks simultaneously, such as object detection for autonomous driving, traffic prediction for route optimization, and driver behavior analysis for safety enhancements. This necessitates the use of multi-model training approaches that can handle diverse ML models concurrently. Despite its promising potential, there exists very limited literature addressing multi-model FL over wireless networks \cite{FedICT, Asynchronous, CommunicationEfficient}, and these works often rely on conventional FL training methods, which face significant communication overhead. Consequently, reducing communication costs while accelerating the overall multi-model training process is a primary concern that demands critical attention for both IoV systems and wireless networks more broadly.

Some early efforts have been dedicated to address these issues, while leveraging the conventional FL architecture. For instance, recent studies have considered the unique characteristics of diverse ML models by effectively utilizing vehicle-side resources to reduce the required number of global iterations, thereby ensuring accelerated model training \cite{MatchingGame}. In these approaches, local models are collectively transmitted only once when their training is finalized on vehicles, effectively minimizing the frequency of communication with the global aggregator (e.g., a cloud server)\cite{AdaptiveP}. Additionally, \cite{MASS} highlighted the importance of selecting appropriate clients for training and effectively aggregating results from various tasks in FL, \cite{TowardsOP} proposed a mobility-aware multi-task decentralized FL framework capable of efficiently lowering training costs and \cite{ManyTask} introduced a task similarity metric based on client vector alignment, reducing communication overhead. While these contributions have advanced the new field, they often overlook the impact of balanced scheduling of ML tasks on global model convergence when handling multiple tasks with a HFL over dynamic IoV systems. To address this issue, \cite{tang2023fedml, TaskSelection, MobilityBased} focus on solving balanced multi-task allocation problem while reducing the global train time cost in FL. However, in such systems, risks may arise due to model obsolescence if synchronous aggregation mechanism, dynamic nature of vehicles, and the trained models are not promptly aggregated upon completion \cite{HiFlash, FedFetch, OptimizingM}. Further, improper prioritization of task training on vehicles can lead to prolonged waiting times for model aggregation, further exacerbating potential model obsolescence.

\subsection{Motivation and Overview of Our Methodology}

This work is among the first to investigate multi-model training in VEC-HFL. In our setting, vehicles serve as training clients, multiple edge servers (ESs) act as intermediate aggregators, and a cloud server (CS) functions as the global aggregator. A key challenge is to mitigate model obsolescence while improving the utilization of distributed and dynamic vehicular data. Existing FL/HFL aggregation schemes are typically either synchronous or asynchronous. Synchronous aggregation waits for all clients before updating, ensuring consistency but incurring high latency and increasing model obsolescence risk. In contrast, asynchronous aggregation reduces latency by updating with partial client participation, but may lead to model inconsistency and degraded convergence. To effectively utilizing all hierarchical data during training process, we introduce a \textit{hybrid synchronous-asynchronous aggregation rule for VEC-HFL}, operating on a per-task basis: at the edge layer, under the constraint of maximum waiting time (detailed in Section IV-A), each ES performs aggregation only after receiving all local models of a specific task from vehicles (loose synchronous approach); while at the cloud layer, the CS conducts global aggregation after receiving a subset of ES models (asynchronous approach). This hybrid design combines the advantages of both paradigms. Edge-layer synchronous aggregation improves consistency and fully exploits local vehicular data, enhancing accuracy and convergence speed. Meanwhile, cloud-layer asynchronous aggregation reduces latency by avoiding waiting for all ESs, which may experience heterogeneous delays due to different vehicle distributions and channel conditions. Overall, this mechanism balances timeliness and model quality, making it well suited for dynamic IoV environments. Table \ref{relate work comparison} summarizes related studies and highlights key differences.

Building on this hybrid aggregation rule, we aim to minimize the global latency while ensuring balanced training across diverse ML tasks over resource-limited vehicles. We formulate this as a min-max optimization problem, which is NP-hard and thus challenging in dynamic IoV systems. To address this, we propose a stagewise framework, \textit{Hybrid Evolutionary And gReedy allocaTion method (HEART)}. In Stage 1 of HEART, we optimize the assignment between training tasks and heterogeneous resource-limited vehicles. Leveraging the global search capability and efficiency of Particle Swarm Optimization (PSO) and Genetic Algorithms (GA), we develop an improved PSO–GA hybrid to obtain a near-optimal and balanced task allocation. In Stage 2 of HEART, we optimize the task execution order on each selected vehicle to accelerate decision-making, using a greedy-based task ranking strategy.

\begin{table}[t!]
	\centering
	\scriptsize
	\vspace{-0.2 cm}
	\caption{A summary of related works}
	\setlength{\tabcolsep}{2 pt} 
	\centering
	\vspace{-0.2 cm} 
	\begin{tabular}{|c|c|c|c|c|c|c|c|c|}
		\hline
		  \multirow{2}{*}{\textbf{Related studies}} & \multicolumn{2}{c|}{\makecell{\textbf{Environmental}\\ \textbf{attributes}}} & \multicolumn{2}{c|}{\textbf{Task count}} & \multicolumn{2}{c|}{\makecell{\textbf{Scene} \\ \textbf{attributes}}} & \multicolumn{1}{c|}{\makecell{\textbf{Training} \\ \textbf{sequence}}} \\ \cline{2-8} 
		& Stable & Dynamic & Single & Multiple & IoT & VEC &If optimized\\
		\hline
		[11][12][13][18] &  & $\checkmark$ & $\checkmark$ &  &  & $\checkmark$ &  \\
		\hline
		[14][15][16][19][22][23] & $\checkmark$ &  &  &  $\checkmark$ & $\checkmark$ &  & \\
		\hline
		[17][20][24] &  & $\checkmark$ &  & $\checkmark$ &  & $\checkmark$ & \\
		\hline
		[8][25][28][30][32] & $\checkmark$ &  & $\checkmark$ &  & $\checkmark$ &  & \\
		\hline
		\rowcolor{veryverylightblue} Our Paper&  & $\checkmark$ &   & $\checkmark$ &  & $\checkmark$ & $\checkmark$ \\
		\hline
	\end{tabular}
	\label{relate work comparison}
\end{table}

\subsection{Summary of Contributions}
Our major contributions can be summarized as follows:

\noindent $\bullet$ \textit{Hybrid Synchronous-Asynchronous Aggregation Rule for Multi-Model Training in VEC-HFL:} We introduce an aggregation rule tailored for VEC-HFL. This hybrid approach performs synchronous aggregation at the ES and asynchronous aggregation at the CS. By doing so, it effectively leverages the rich data generated by vehicles while significantly mitigating the adverse effects of model obsolescence, thereby enhancing the overall timeliness of the training process across ML tasks.

\noindent $\bullet$ \textit{Efficient Two-Stage Optimization Method (HEART) to Minimize Global Time Cost:} Addressing the NP-hard nature of minimizing the overall time cost in the VEC-HFL process, we develop an efficient two-stage optimization method named HEART. In Stage 1, we optimize task scheduling for vehicles by designing a hybrid Particle Swarm Optimization-Genetic Algorithm (PSO-GA), which combines the global search capabilities of PSO with the robust convergence properties of GA. In Stage 2, we optimize the task training sequence on the vehicle side by introducing a low-complexity greedy algorithm, which sequentially assigns tasks to maximize overlap and minimize uploading time. This two-pronged approach ensures both effective task distribution and efficient training sequence/rank planning, benefiting from a low computational complexity while maintaining high performance.

\noindent $\bullet$ \textit{Superior Performance Demonstration through Extensive Simulations on Real-World Datasets:} We conduct comprehensive simulations using real-world datasets to evaluate the performance of HEART. The results demonstrate its superior performance in terms of time efficiency and the overall communication costs compared to baseline methods. These findings highlight the effectiveness of HEART in dynamic IoV systems, showcasing its potential for practical deployment and scalability in real-world scenarios.

The remaining of this paper is structured as follows. Section II provides relevant background and description. Section III presents the framework for multi-model task training and scheduling in VEC-HFL. Section IV provides problem formulation and solution description. Section V provides simulation results, and Section VI concludes this work and outlines interesting future work directions.

\section{Background and Preliminaries}
\noindent The VEC-HFL architecture of our interest (see Fig. 1) comprises one CS as the global aggregator, multiple ESs as middle-layer aggregators, and multiple vehicles as model training clients moving between ESs. We collect the ESs via the set $\mathcal{M}=\{1,...,M\}$, where each ES $m\in\mathcal{M}$ has a certain communication coverage. Also, we denote the set of vehicles in the IoV system by $\mathcal{N}=\{1,...,N\}$, where each vehicle $n\in \mathcal{N}$ is assumed to be located inside the considered IoV region during the training process, moving between ESs at varying speeds. Moreover, we denote the set of all ML tasks in the system by $\mathcal{J}=\{1,...,J\}$, which are expected to be scheduled and trained/executed over moving vehicles.
\begin{figure}[!t]
    \vspace{-4mm}
    \centering
    \includegraphics[trim=0.1cm 0.1cm 0.1cm 0.1cm, clip, width=1\columnwidth]{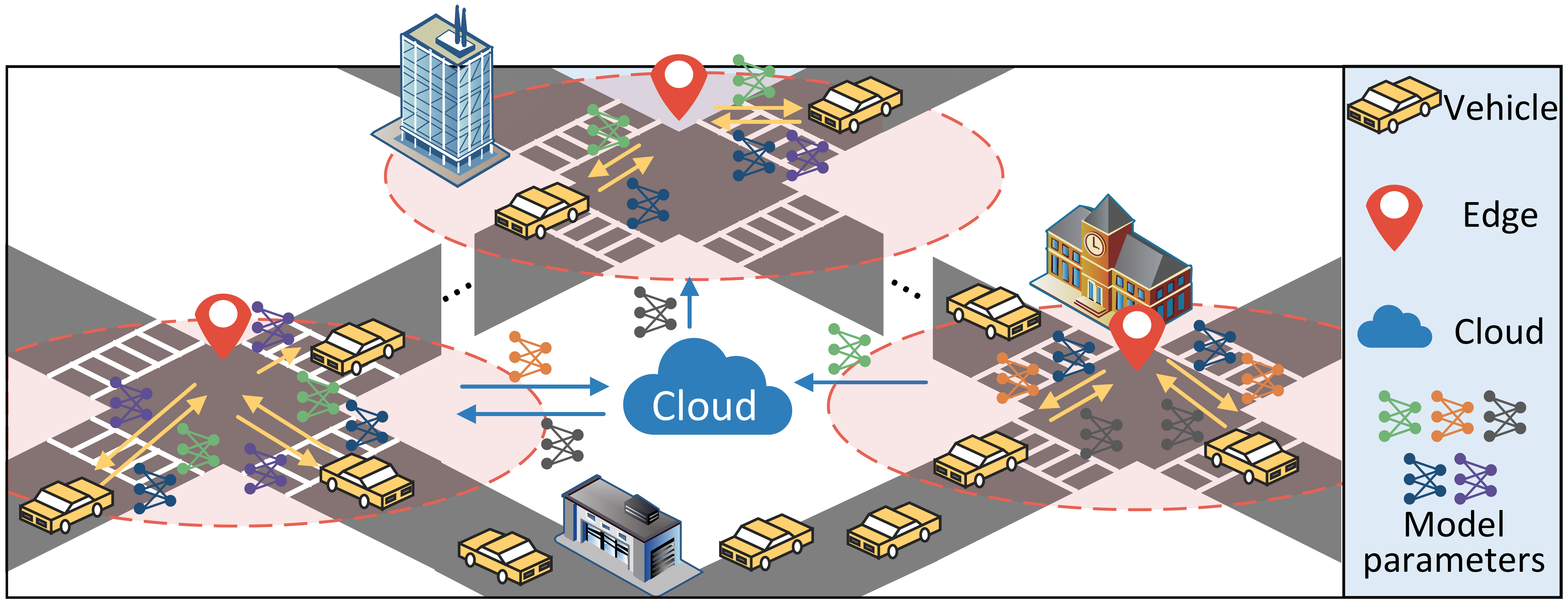}
    \caption{ A schematic of multi-model training over VEC-HFL.}
    \vspace{-3mm}
\end{figure}

To enhance data utilization and mitigate model obsolescence, we introduce a hybrid synchronous-asynchronous aggregation rule for VEC-HFL as illustrated in Fig. 2. Presuming a specific task/model training rank across the vehicles (further concertized in Section III and optimized in Section IV), once vehicle $n$ completes the local training of a task $j\in\mathcal{J}$, it transmits the local model to its associated ES, and then proceeds to training the next task according to the task ranking. Here, for each task $j$, an ES waits for local models from vehicles within coverage who train the task until a predefined deadline is reached, rather than indefinitely waiting for all participants (loose synchronous mechanism), before conducting the local model aggregation. Once the models for task $j$ are aggregated at each ES, they then broadcast the updated edge model back to associated vehicle to prompt the start of the next local training round at the vehicles. Differently, the CS employs an asynchronous aggregation rule where once it receives a subset of edge models (e.g., from $Q^{\left \langle j \right \rangle}$ ESs) for task $j$, it performs global aggregation and broadcasts the updated model to the ESs. 

Subsequently, the processes taken place in VEC-HFL can be divided into \textit{local iterations} (i.e., local ML training using stochastic gradient descent), \textit{edge iterations} (i.e., model transfer from the vehicles to the ESs followed by model aggregation), and \textit{global iterations} (i.e., model transfer from a subset of ESs to the CS followed by model aggregations), where we use notation $h$, $k$ and $g$ to index them, respectively. Besides, for different tasks (e.g., task $j$), we also set corresponding maximum number of local iteration and edge iterations to be $H^{\left \langle j \right \rangle}$ and $K^{\left \langle j \right \rangle}$, which implies $h\in \{1,\cdots, H^{\left \langle j \right \rangle}\}$ and $k\in \{1,\cdots, K^{\left \langle j \right \rangle}\}$ (the configuration of $H^{\left \langle j \right \rangle}$ and $K^{\left \langle j \right \rangle}$ follow the foundational concepts of \cite{optimes} and is specifically optimized to align with the VEC-HFL scenario proposed in this paper). We assume that the training of each task $j$ starts with broadcasting a unified global model $\varpi_{[0]}^{\left \langle j \right \rangle}$ across all the ESs and vehicles and further assume that each task $j$ continues its global iterations until its global model satisfies its convergence condition. Consequently, the maximum number of global iterations $g$ is not predetermined. Next, we provide a mathematical formalization of the VEC-HFL architecture. 

For each task $j$, we model the dataset of the $n^{\textrm{th}}$ vehicle within the coverage of ES $m$ as $\mathcal{D}_{m,n}^{\left \langle j \right \rangle}=\{(x_i,y_i):1\leq i\leq|\mathcal{D}_{m,n}^{\left \langle j \right \rangle}|\}$, where $x_{i}$ and $y_{i}$ describe the feature vector and the label of the $i^{\text{th}}$ data point and $|\mathcal{D}_{m,n}^{\left \langle j \right \rangle}|$ represents the size of the dataset. Accordingly, the loss function of task $j$ at this vehicle is given by
\begin{align}
L^{\left \langle j \right \rangle}_{m,n}\left(\omega_{m,n;[g,k,h]}^{\left \langle j \right \rangle}\right)=\sum_{i=1}^{|\mathcal{D}_{m,n}^{\left \langle j \right \rangle}|}l^{\left \langle j \right \rangle}_i(\omega_{m,n;[g,k,h]}^{\left \langle j \right \rangle}),
\end{align}
where $l^{\left \langle j \right \rangle}_i\left(\omega_{m,n;[g,k,h]}^{\left \langle j \right \rangle}\right)$ denotes the loss function for the $i^{\text{th}}$ data point given the instantaneous local model parameters $\omega_{m,n;[g,k,h]}^{\left \langle j \right \rangle}$ for the $n^{\text{th}}$ vehicle covered by ES $m$ during the $h^{\text{th}}$ local iteration of the $k^{\text{th}}$ edge iteration of global iteration $g$. In particular, the update process of the instantaneous local model on this vehicle is carried out using Stochastic Gradient Descent (SGD) iterations as follows $(h\in \{1,\cdots, H^{\left \langle j \right \rangle}\})$: 
\begin{multline}
        \omega_{m,n;[g,k,h]}^{\left \langle j \right \rangle}=\omega_{m,n;[g,k,h-1]}^{\left \langle j \right \rangle}-\\
        \eta^{\left \langle j \right \rangle}\nabla L_{m,n}^{\left \langle j \right \rangle}(\omega_{m,n;[g,k,h-1]}^{\left \langle j \right \rangle}, \mathcal{B}_{m,n;[g,k,h]}^{\left \langle j \right \rangle}),
\end{multline}where $\eta^{\left \langle j \right \rangle}$ is the learning rate, and $\mathcal{B}_{m,n;[g,k,h]}^{\left \langle j \right \rangle}$ is a mini-batch of data sampled randomly from the local dataset $\mathcal{D}_{m,n}^{\left \langle j \right \rangle}$. The local model of the edge iteration $k$ for each global iteration $g$ is initialized with $w^{\left \langle j \right \rangle}_{m,n;[g,k,0]}=\tilde{w}^{\left \langle j \right \rangle}_{m;[g,k-1]}$ which is the latent model received from the edge. Also, the latest local model $w^{\left \langle j \right \rangle}_{m,n;[g,k,H^{\left \langle j \right \rangle}]}$ is sent to the ES and used to obtain the next edge model as discussed below.

Considering the $k^{\text{th}}$ edge iteration of global iteration $g$ for an ES $m$, we denote the set of vehicles under its coverage who also train task $j$ as $\mathcal{N}_{m;[g,k]}^{\left \langle j \right \rangle}$. However, due to potential abrupt changes in channel conditions or vehicle dropouts in practical scenarios, vehicles may fail to transmit model parameters to the ES in a timely manner. Therefore, within the coverage of ES $m$, the set of vehicles that actually complete the training of task within the maximum waiting time is denoted as $\mathcal{N}_{m;[g,k]}^{Act, \left \langle j \right \rangle}$ (detailed in Section IV-A). Letting $\tilde{\omega}_{m;[g,k]}^{\left \langle j \right \rangle}$ represent the edge model for task $j$ at the $k^{\text{th}}$ edge iteration of global iteration $g$, its update can be expressed as follows:
\begin{align}
    \widetilde{\omega}_{m;[g,k]}^{\left \langle j \right \rangle}=\sum_{n\in\mathcal{N}_{m;[g,k]}^{Act,\left \langle j \right \rangle}}\frac{|\mathcal{D}_{m,n}^{\left \langle j \right \rangle}|}{|\widetilde{\mathcal{D}}_{m;[g,k]}^{\left \langle j \right \rangle}|}\omega_{m,n;[g,k,H^{\left \langle j \right \rangle}]}^{\left \langle j \right \rangle}, \label{3}
\end{align}where $\widetilde{ \mathcal{D}}_{m;[g,k]}^{\left \langle j \right \rangle}=\cup_{n\in\mathcal{N}_{m;[g,k]}^{\left \langle j \right \rangle}}\mathcal{D}_{m,n}^{\left \langle j \right \rangle}$, and $|\widetilde{\mathcal{D}}_{m;[g,k]}^{\left \langle j \right \rangle}|$ represents the overall data volume for task $j$ during the $k^{\text{th}}$ edge iteration possessed by the vehicles covered by ES $m$. The latest edge model $\widetilde{\omega}_{m;[g,K^{\left \langle j \right \rangle}]}^{\left \langle j \right \rangle}$ is sent to the CS and used to obtain the next global model as discussed below.

For each $j\in\mathcal{J}$, once the CS receives a predetermined number of edge models (i.e., $Q^{\left \langle j \right \rangle}$), it performs the global aggregation. Let $\mathcal{M}_{[g]}^{\left \langle j \right \rangle}$, where $|\mathcal{M}_{[g]}^{\left \langle j \right \rangle}|=Q^{\left \langle j \right \rangle}$, represent the set of ESs who deliver the edge model of task $j$ to the CS at the $g^{\text{th}}$ global aggregation. The update process of global model for task $j$ is as follows:
\begin{multline}
    \varpi_{[g]}^{\left \langle j \right \rangle}=\alpha^{\left \langle j \right \rangle}\varpi_{[g-1]}^{\left \langle j \right \rangle}+\\
    (1-\alpha^{\left \langle j \right \rangle})\sum_{m\in\mathcal{M}_{[g]}^{\left \langle j \right \rangle}}\frac{|\widetilde{\mathcal{D}}_{m;[g,K^{\left \langle j \right \rangle}]}^{\left \langle j \right \rangle}|}{|\overline{\mathcal{D}}_{[g]}^{\left \langle j \right \rangle}|}\widetilde{\omega}_{m;[g,K^{\left \langle j \right \rangle}]}^{\left \langle j \right \rangle}, \label{4}
\end{multline}
where $\alpha^{\left \langle j \right \rangle}$ represents the weighting coefficient between the current global model and the updated edge models received from the ESs. $\overline{\mathcal{D}}_{[g]}^{\left \langle j \right \rangle}=\cup_{m\in\mathcal{M}_{[g]}^{\left \langle j \right \rangle}}\widetilde{\mathcal{D}}_{m;[g,K^{\left \langle j \right \rangle}]}^{\left \langle j \right \rangle}$, and $|\overline{\mathcal{D}}_{[g]}^{\left \langle j \right \rangle}|$ represents the overall number of data points involved in training task $j$ during the $g^{\text{th}}$ global aggregation. The global model $\varpi_{[g]}^{\left \langle j \right \rangle}$ is then sent back to these ESs, which relay it back to their covered vehicles to synchronize their local models and start the next round of local training. We consider that the global iterations for each task $j$ ends at iteration $g$ when its following convergence criterion
\begin{align}
    \|\varpi_{[g]}^{\left \langle j \right \rangle} - \varpi_{[g-1]}^{\left \langle j \right \rangle}\| \leq \zeta^{\left \langle j \right \rangle}, \label{5}
\end{align}
where $\|\cdot\|$ represents the Euclidean 2-norm, and $\zeta^{\left \langle j \right \rangle}$ represents a small positive value used to control the convergence criterion. Alg. 1 provides an overview of our designed hybrid synchronous-asynchronous aggregation rule for VEC-HFL.
\begin{figure}[htbp]
    \vspace{-4mm}
    \centering
    \includegraphics[trim=0.0cm 0.0cm 0.0cm 0.0cm, clip, width=1\columnwidth]{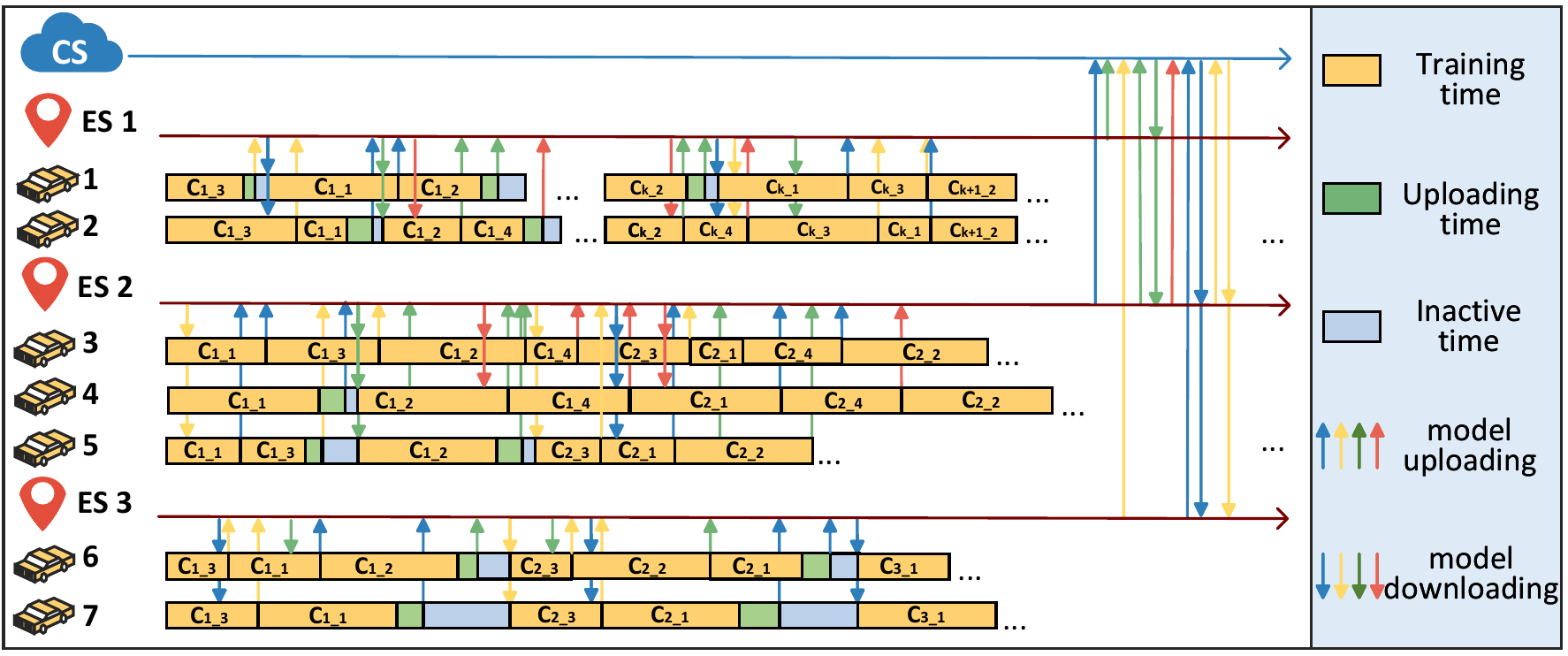}
    \caption{A schematic of multi-task scheduling and training VEC-HFL architecture of our interest.}
    \vspace{-4mm}
\end{figure}

\section{Multi-Task Scheduling over VEC-HFL Architecture}
\noindent In this section, to better analyze the time cost of the proposed VEC-HFL architecture, we first analyze the local and edge iterations, deriving the local training time for vehicles and the time required for ESs to complete an entire edge iteration. Then, we provide an analysis for the global iteration.
\begin{algorithm}[htb]
	{
		\small
		
		
		\caption{Overview of the VEC-HFL of our interest}
		
		\SetKwInOut{Input}{Input}\SetKwInOut{Output}{Output}
		
		\Input{$H^{\left \langle j \right \rangle}$, $K^{\left \langle j \right \rangle}$, $\eta^{\left \langle j \right \rangle}$;}
		
		\Output{Global model $\varpi_{[g]}^{\left \langle j \right \rangle}$;}

            {\bf{Initialization :}} 
		$ w_{m,n}^{\left \langle j \right \rangle}=\varpi^{\left \langle j \right \rangle}_{[0]} $;

            \If{for task $j$, where $j\in\mathcal{J}$, the global model parameter changes does not satisfy (5)}{
                \textit{\textbf{Edge process:}} // running at each ES

                \ForEach{ $m\in\mathcal{M}$ in parallel}{
                
                    \For{$k=\{1,2,\cdots,K^{\left \langle j \right \rangle}\}$}{
                
                        \textit{\textbf{Vehicle process:}} // $\overline{x}_{m,n;[g]}^{\left \langle j \right \rangle}$ obtained from Alg. 2

                        \ForEach{ $n\in\mathcal{N}$ in parallel}{

                        \For{j $\in \{\overline{x}_{m,n;[g]}^{\left \langle j \right \rangle}=1\}$ in turn}{
                            Synchronize the local model based on edge model: $w_{m,n;[g,k,0]}^{\left \langle j \right \rangle} \xleftarrow{} \widetilde{w}_{m;[g,k-1]}^{\left \langle j \right \rangle}$

                            vehicle $n$ undergoes local training in sequence based on the task training rank obtained from Alg. 3

                            \For{$h=\{1,2,\cdots,H^{\left \langle j \right \rangle}\}$}{

                                Update local model based on (2), and after training, send the final local model $\omega_{m,n;[g,k,H^{\left \langle j \right \rangle}]}^{\left \langle j \right \rangle}$ to ES     
                                }
                            }
                        }

                        \If{models received from all $n\in \mathcal{N}_{m;[g,k]}^{Act,\left \langle j \right \rangle}$}{
                        Update edge model $\tilde{\omega}_{m;[g,k]}^{\left \langle j \right \rangle}$ based on (3)
                    }

                    }

                    Send edge model $\widetilde{\omega}_{m;[g]}^{\left \langle j \right \rangle}$ to CS   
                }

                \textit{\textbf{Cloud process:}} // for task $j$
                
                \If{Received $Q^{\left \langle j \right \rangle}$ edge models }{

                    Update the global model $\varpi_{[g]}^{\left \langle j \right \rangle}$ according to (4)

                    Broadcast $\varpi_{[g]}^{\left \langle j \right \rangle}$ to the appropriate vehicle based on Alg. 2
                }

                \If{the global model parameter changes satisfy (5)}{
        
                    $\mathcal{J} \xleftarrow{} \mathcal{J} \setminus \{j\}$
                }
	    }
}
\end{algorithm}

\subsection{Modeling the Time Overhead of Local and Edge Iterations}
The time consumed to perform one local training round (i.e., one SGD iteration) at the $n^{\text{th}}$ vehicle covered by ES $m$ for task $j$ can be calculated as follows:
\begin{align}
    t_{m,n;[g,k,h]}^{\langle j\rangle}=\frac{|\mathcal{B}_{m,n;[g,k,h]}^{\left \langle j \right \rangle}| c^{\langle j\rangle}_{m,n}}{f^{\langle j\rangle}_{m,n} }+b^{\langle j\rangle}_{m,n},
\end{align}
where $c^{\langle j\rangle}_{m,n}$ represents the number of cycles that it takes to process one data point of task $j$, $f^{\langle j\rangle}_{m,n}$ denotes the allocated CPU clock frequency of the vehicle to conduct ML training for task $j$, and $|\mathcal{B}_{m,n;[g,k,h]}^{\left \langle j \right \rangle}|$ is the number of data points contained in the mini-batch of SGD. Moreover, $b^{\langle j\rangle}_{m,n}$ is a constant value, representing the migration time of the model between the CPU and GPU, or instantiating the model parameters. Harder tasks (e.g., those that use deeper neural networks) are often associated with larger values of $c^{\langle j\rangle}_{m,n}$ and $b^{\langle j\rangle}_{m,n}$.

We model the configuration of task processing (i.e., task-to-vehicle assignment) via a binary indicator variable $\overline{x}_{m,n;[g]}$: $\overline{x}_{m,n;[g]}^{\left \langle j \right \rangle}=1$ indicates that task $j$ is processed on the $n^\text{th}$ vehicle of ES $m$ during the $g^{\text{th}}$ global iteration; and $\overline{x}^{\left \langle j \right \rangle}_{m,n;[g]}=0$  otherwise. Additionally, since the ESs employ a synchronous aggregation rule for each task, when the task sequence of vehicles (i.e., the order in which the vehicles execute their local tasks) within the ES coverage exhibit a higher degree of overlap, the local models of all tasks can be aggregated at shorter intervals to mitigate the effects of model obsolescence. Furthermore, in the dynamic IoV channels considered in this work, uploading local models at optimal times during each edge iteration, which can be achieved through optimizing the task sequence of vehicles, can significantly reduce communication latency and global communication costs. For instance, uploading tasks with larger model parameters when vehicles are closer to the ES dramatically decreases both model upload time and communication delays compared to uploading when vehicles are farther away.

To formally describe this process, we introduce $\lambda_{m,n;[g,k]}$ to represent the training sequence of tasks on the $n^\text{th}$ vehicle covered by ES $m$ during the $k^{\text{th}}$ edge iteration of global iteration $g$. For example, if this vehicle is assigned tasks 1, 3, and 4, then the training sequence \(\lambda_{m,n;[g,k]}\) is \([\text{4}, \text{1}, \text{3}]\). We further use $\mathcal{I}\bigl(\lambda_{m,n;[g]}^{\langle j \rangle}\bigr)$ to denote the position of task \(j\) on the \(n^\text{th}\) vehicle within the coverage of ES \(m\). Continuing the previous example, if \(\lambda_{m,n;[g,k]} = [\text{4}, \text{1}, \text{3}]\), then $\mathcal{I}\bigl(\lambda_{m,n;[g]}^{\langle 1 \rangle}\bigr) = \text{2}$.  This notation helps model the completion time for each task during edge iterations, which in turn provides a tractable framework to address the aforementioned challenges by optimizing the task training sequence \(\lambda_{m,n;[g,k]}\) across various vehicles. Below, we integrate task training sequence in modeling the task completion time and will later  provide the details of the optimization of the task training sequence in Section~IV. 

Due to the complexities involved in modeling the timeline of the processes that take place at each vehicle in our hybrid synchronous-asynchronous aggregation rule, we divide the time span dedicated to conducting local iterations of a vehicle into two parts: \textit{task-training time} and \textit{non-task-training time}. For the former, the task-training time of the $n^{\text{th}}$ vehicle covered by ES $m$ to complete a task (e.g., task $j$) during the $k^{\text{th}}$ edge iteration of the $g^{\text{th}}$ global iteration can be calculated by considering the number of local iterations ($H^{\langle j\rangle}$), and the time consumed by each local training round (i.e., $t_{m,n;[g,k,h]}^{\langle j\rangle}$ given by (6)), which can be defined as
\begin{align}
    t_{m,n;[g,k]}^{Cmp,{\langle j\rangle}}=\sum_{h=1}^{H^{\left \langle j \right \rangle}}t_{m,n;[g,k,h]}^{\langle j\rangle}. \label{7}
\end{align}

The non-task-training time comprises \textit{(i)} local model upload and \textit{(ii)} the idle interval between consecutive tasks\footnote{We have disregarded the downlink communication delays (from CS to ESs and ESs to vehicles) because: \textit{(i)} the high broadcasting power of the CS and ESs typically results in negligible latency as in \cite{MatchingGame, luo2020hfel}; and \textit{(ii)} such delays are generally constant across different tasks\cite{MOBFL} (i.e., determined by server-side capacity rather than individual vehicle states), thereby not affecting the optimization of scheduling and training sequences. For conciseness, they are omitted from our model.}. During the $g^{\text{th}}$ global iteration, vehicles upload immediately after local training, yielding two cases:

$\bullet$ \textit{Case (i)}: \textit{Immediate local execution}. If the next edge model has been received before the current task finishes, training proceeds without interruption; upload and training overlap, so non-task-training time is negligible.

$\bullet$ \textit{Case (ii)}: \textit{Delayed local execution}. Otherwise, non-task-training time equals the upload duration plus the idle time until the next edge model is received.

Fig. 2 illustrates these cases using four training tasks, seven vehicles, and three ESs. Colored arrows (blue, green, yellow, red) represent uploading and broadcasting for Tasks 1–4, while $\mathsf{C_{i\_j}}$ denotes the $j^\text{th}$ task in the $i^\text{th}$ edge iteration of a global iteration (e.g., $\mathsf{C_{1\_2}}$ is the second task in the first edge iteration). Vehicle 1 completes Task 3 without an immediately available next task (i.e., \textit{case (ii)}), and thus remains idle until Task 1 is broadcast by the ES. In contrast, Vehicle 2 receives Task 1 before finishing Task 3 (i.e., \textit{case (i)}), enabling seamless transition without idle time. After collecting two edge models from ESs, the CS performs global aggregation and re-broadcasts the updated global model to ESs (see arrows between ESs and CS, right side of Fig. 2).

Next, we provide a more formal description of the non-task-training time process. Firstly, the upload time of task $j$ of the $n^\text{{th}}$ vehicle covered by ES $m$ during the $k^{\text{th}}$ edge iteration of the $g^{\text{th}}$ global iteration can be expressed by 
\begin{align}
    t_{m,n;[g,k]}^{V2E,\left \langle j \right \rangle}=I_{m,n}^{V2E,\left \langle j \right \rangle}/r_{m,n;[g,k]}^{V2E}, \label{8}
\end{align}
where $r_{m,n;[g,k]}^{V2E,\left \langle j \right \rangle}$ represents the vehicle-to-edge data rate of vehicle $n$ under the coverage of ES $m$ during the $k^{\text{th}}$ edge iteration, which is related to the distance between the vehicle and ES and the channel conditions \cite{Platoon}. Besides, the size of the local model (in bits) is denoted by $I_{m,n}^{V2E,\left \langle j \right \rangle}$.

Let $t^{inact,\langle j\rangle}_{m,n;[g,k]}$ denote the inactive time of the $n^\text{th}$ vehicle covered by ES $m$ after completing its local iteration at edge iteration $k$ of global aggregation $g$. It can be observed that this inactive time is a function of the completion time of edge iteration $k-1$ within global aggregation $g$ for a task $j$, which we denote it by $t_{m,n;[g,k-1]}^{end,\langle j\rangle}$. Mathematically, we obtain this inactive time as follows:
\begin{multline}
    t^{inact,\left \langle j \right \rangle}_{m,n;[g,k]}=\max\Bigg\{\underbrace{\max_{n'\in\mathcal{N}_{m,g}^{\left \langle j \right \rangle}} \Big\{ t^{end,\left \langle j \right \rangle}_{m,n';[g,k-1]} \Big\}}_{(a)}-\\
     \underbrace{t^{end,\left \langle j^\prime \right \rangle}_{m,n;[g,k]} }_{(b)}, 0 \Bigg\},j^\prime:~\mathcal{I}(\lambda_{m,n;[g]}^{\langle j^\prime \rangle}) = \mathcal{I}(\lambda_{m,n;[g]}^{\left \langle j \right \rangle})-1,
\end{multline}
where term $(a)$ represents the time where all the other vehicles send their local models of task $j$ to ES $m$ at the $(k-1)^\text{th}$ edge iteration, which creates the updated edge model for starting the next edge iteration (i.e., the $k^\text{th}$ within the global iteration $g$), and $(b)$ indicates the time that vehicle $n$ will start processing task $j$ during the $k^\text{th}$ edge iteration in global iteration $g$. We will later obtain $t^{{end},\left \langle j \right \rangle}_{m,n;[g,k]}$ as a function of $t^{{inact},\left \langle j \right \rangle}_{m,n;[g,k]}$ in (12), which combined with (9) forms a recursive expression between these two quantities.

Subsequently, the non-task training time of the $n^\text{th}$ vehicle covered by ES $m$ is given by
\begin{align}
    t_{m,n;[g,k]}^{Ntt,{\langle j\rangle}}=\begin{cases}~~0& \text{if}~case~(i)\\~~t_{m,n;[g,k]}^{V2E,\left \langle j \right \rangle}+t^{inact,{\langle j\rangle}}_{m,n;[g,k]}& \text{if}~case~(ii)\end{cases}.
\end{align}

Therefore, for the $n^\text{th}$ vehicle covered by ES $m$, the time required to complete location iterations for task $j$ in the $k^\text{th}$ edge iteration of global iteration $g$ can be expressed as
\begin{align}
    t_{m,n;[g,k]}^{\langle j\rangle}=t_{m,n;[g,k]}^{Cmp,{\langle j\rangle}}+t_{m,n;[g,k]}^{Ntt,{\langle j\rangle}}.
\end{align}

Since the ESs utilize synchronous aggregation rule, it needs to wait for all vehicles assigned to the task to upload their local models before performing edge aggregation. Nevertheless, different vehicles have different training sequences for tasks, and their training and uploading times are also different. Thus, the completion of edge iteration $k$ of global aggregation $g$ for a task $j$ at the $n^\text{th}$ vehicle covered by ES $m$ is a function of the execution times of the tasks that have lower execution rank as compared to this task (i.e., get processed before task $j$); mathematically, we have 
\begin{align}
    t_{m,n;[g,k]}^{end,\langle j\rangle}= \hspace{-4mm}\sum_{j^\prime:~\mathcal{I}(\lambda_{m,n;[g]}^{\langle j^\prime \rangle}) \leq \mathcal{I}(\lambda_{m,n;[g]}^{\langle j\rangle})} \hspace{-4mm}t_{m,n;[g,k]}^{\langle j^\prime\rangle}\overline{x}_{m,n;[g]}^{\left \langle j^\prime \right \rangle},j\in\mathcal{J},
\end{align}
where the summation is taken over all the tasks that get executed before task $j$, where the execution time of task $j$ is included as well. Then, combined with the synchronous aggregation rule, the overall time cost/overhead of execution of task training for task $j$ at ES $m$, which is dictated by the last vehicle that uploads it model to the ES, at the end of edge iteration $k$ of global aggregation $g$ can be obtained as
\begin{align}
    T_{m;[g,k]}^{\langle j\rangle}=\max_{n\in\mathcal{N}_{m;[g]}^{Act, \langle j\rangle}}\left\{t_{m,n;[g,k]}^{end,\langle j\rangle}\right\}.
\end{align}

\subsection{Modeling the Time Overhead of Global Iterations} 
When an ES completes the edge training of task $j$, the edge model will be transmitted to CS for global aggregation. We denote the upload time from ES $m$ to the CS as
\begin{align}
    t^{E2C,\left \langle j \right \rangle}_{m}=I_{m}^{E2C,\left \langle j \right \rangle}/r_{m}^{E2C},
\end{align}
where $I_{m}^{E2C,\left \langle j \right \rangle}$ represents the size of edge model, and $r_{m}^{E2C}$ represents the data rate of ES $m$ to CS, which is assumed to be time-invariant as these communications often take place over the backhaul network wired links. Then, we can get the time required for each task to complete one round of global iteration at ES $m$, as given by
\begin{align}
    T_{m;[g]}^{\left \langle j \right \rangle}=t^{E2C,\left \langle j \right \rangle}_{m} + \sum_{k=1}^{K^{\langle j\rangle}}T_{m;[g,k]}^{\langle j\rangle}.
\end{align}

\begin{remark}(Scope and Scalability)
\textbf{Focus on Core Bottlenecks:} This work focuses on task balancing and sequence optimization to minimize global training latency. While inter-vehicle or inter-ES cooperation could be considered, our framework requires only minimal control information exchange, which is well-supported by current hardware. Explicitly modeling cooperative communication would significantly increase problem dimensionality without offering substantial latency gains, thus it is omitted to maintain architectural conciseness. \textbf{Extensibility for Large Models:} To address the computational burden of training multiple models, especially large-scale ones (e.g., Large Language Model (LLM)), HEART is designed as a modular framework. It can seamlessly integrate model-level efficiency techniques such as Federated Dropout\cite{Federatedl}, Split Learning\cite{thapasplitfed}, or Low-Rank Adaptation (LoRA)\cite{hulorae}. These methods complement our system-level scheduling, ensuring efficient execution on resource-constrained vehicles and underscoring the framework's strong scalability.
\end{remark}

\section{Problem Formulation and Design of HEART}
\noindent  This section first provides key constraints involved. Then, we present problem formulation and explain solution method.

\subsection{Key Constraint Design} 
In the IoV context, one key concern is ensuring that vehicles complete their training tasks and transmit the results to the corresponding ES promptly, mitigating potential negative impacts on training performance caused by network dynamics. Accordingly, we introduce a constraint given by 
\begin{align}
    \sum_{j \in \mathcal{J}}\sum_{k \in K^{\langle j\rangle}}t_{m,n;[g,k]}^{\langle j\rangle}\overline{x}_{m,n;[g]}^{\left \langle j \right \rangle} < t_{m,n;[g]}^{stay}, \label{16}
\end{align}
where $t_{m,n;[g]}^{stay}$ represents the dwell time of the $n^{\text{th}}$ vehicle within its associated ES $m$ coverage, where $\overline{x}_{m,n;[g]}^{\left \langle j \right \rangle}\in\{0,1\}$ is defined in Section III-A. 

Moreover, training multiple models and uploading them to the ES incurs substantial energy consumption at vehicles. Let $\overline{p}_{n}^{\left \langle j \right \rangle}$ denote the average power consumed by vehicle $n$ during the training and transmission of task $j$ \cite{DistributedP,energybb,energyll}. We impose an energy constraint to ensure that the total energy consumption of vehicle $n$ in each global iteration does not exceed its available energy budget, defined as follows:
\begin{align}
    \sum_{j \in \mathcal{J}}\sum_{k \in K^{\langle j\rangle}}\overline{p}_{n}^{\left \langle j \right \rangle}t_{m,n;[g,k]}^{\langle j\rangle}\overline{x}_{m,n;[g]}^{\left \langle j \right \rangle} < \text{e}_{n;[g]}^{ini}, \label{117}
\end{align}
where $\text{e}_{n;[g]}^{ini}$ represents the initial energy of vehicle n before the start of each global iteration.

Although constraints \eqref{16} and \eqref{117} guarantee task completion within edge iterations, channel instability may cause delays (i.e., stragglers) or effectively infinite latency (i.e., dropouts), rendering strict synchronous aggregation inefficient. We therefore adopt a deadline-based mechanism: once the ES receives the first local model parameter, it starts a timer; if a predefined threshold is exceeded, aggregation proceeds (i.e., relaxed/loose synchronous aggregation). Let $T^{Sta,\langle j \rangle}_{m;[g,k]}$ denote the time instant when the ES initiates the training request (i.e., when the first vehicle has uploaded its model parameters to the associated ES), and let $t^{ini,\langle j \rangle}_{m,n;[g,k]}$ denote the timestamp when vehicle $n$ starts training task $j$. For each $n\in\mathcal{N}_{m;[g,k]}^{\left \langle j \right \rangle}$, the theoretical completion time is expressed as:
\begin{equation}
		t^{The, \langle j \rangle}_{m,n;[g,k]} = t^{ini,\langle j \rangle}_{m,n;[g,k]} + t_{m,n;[g,k]}^{Cmp,{\langle j\rangle}} + t_{m,n;[g,k]}^{V2E,\left \langle j \right \rangle},
\end{equation}
where $t_{m,n;[g,k]}^{Cmp,{\langle j\rangle}}$ and $t_{m,n;[g,k]}^{V2E,\left \langle j \right \rangle}$ are the computation and communication latencies defined in \eqref{7} and \eqref{8}, respectively.

Consequently, the theoretical latest completion time $T^{The, \langle j \rangle}_{m;[g,k]}$, representing the moment the ES expects to receive all local updates under ideal conditions (i.e., without random channel interference), is determined by:
\begin{equation}
	T^{The, \langle j \rangle}_{m;[g,k]} = \max_{n \in \mathcal{N}_{m;[g,k]}^{\left \langle j \right \rangle}} \left\{ t^{The, \langle j \rangle}_{m,n;[g,k]} \right\}.
\end{equation}

To enhance system robustness, the actual edge aggregation trigger time satisfies:
\begin{equation}
	T^{Agg, \langle j \rangle}_{m;[g,k]} = \min \{T^{The, \langle j \rangle}_{m;[g,k]}, T^{Fin, \langle j \rangle}_{m;[g,k]}, T^{Sta, \langle j \rangle}_{m;[g,k]} + \Delta_{\max} \}, \label{19}
\end{equation}
where $T^{Fin, \langle j \rangle}_{m;[g,k]}$ is the actual time when all assigned vehicles have successfully uploaded model parameters and $\Delta_{\max}= (1 + \epsilon) \mathbb{E}[T^{Agg, \langle j \rangle}_{m;[g,k]} - T^{Sta, \langle j \rangle}_{m;[g,k]}]$ represents the preset \textit{maximum waiting threshold}, with $\epsilon$ being the fault tolerance ratio (e.g., $\epsilon = 0.1$).

As for the handling of laggards and dropouts, we have formulated the following rules. If vehicles remain incomplete at time $T^{Sta, \langle j \rangle}_{m;[g,k]} + \Delta_{\max}$ (i.e., maximum waiting time), the ES executes the following:

$\bullet$ \textit{Stop waiting:} Immediately disconnect from ``laggards'' to avoid slowing overall training progress.

$\bullet$ \textit{Partial aggregation:} Perform weighted aggregation only on received model parameters to update the edge model (i.e., \eqref{3}), where $\mathcal{N}_{m;[g,k]}^{Act,\left \langle j \right \rangle} \subset \mathcal{N}_{m;[g,k]}^{\left \langle j \right \rangle}$ denotes the set of vehicles that successfully deliver model parameters within $\Delta_{\max}$.

\begin{remark}(Decoupled Deadline Mechanism)
Notably, the maximum waiting time is imposed solely at the aggregation level and is decoupled from task allocation and training rank optimization. Vehicle selection remains unchanged, with the deadline applied only during edge aggregation, thereby preserving design modularity and enhancing robustness to dynamic uncertainties.
\end{remark}

Next, we focus on imposing balance across the training of different tasks. To reach this, we let $\sum_{m \in \mathcal{M}_{[g]}^{\left \langle j \right \rangle}}\sum_{n\in 
\mathcal{N}_{m;[g,k]}^{\left \langle j \right \rangle}}{\overline{x}_{m,n;[g]}^{\left \langle j \right \rangle}}$ denote the number of times that task $j$ has been assigned/trained at the $g^{\text{th}}$ global iteration. Subsequently, we impose the training balance across the tasks via the following constraint
\begin{align}
    \hspace{-0.28cm}\frac{N}{J}-\xi_1\leq \sum_{m \in \mathcal{M}_{[g]}^{\left \langle j \right \rangle}}\sum_{n\in 
\mathcal{N}_{m;[g,k]}^{\left \langle j \right \rangle}}  \hspace{-2mm}{\overline{x}_{m,n;[g]}^{\left \langle j \right \rangle}}\leq\frac{N}{J}+\xi_2, \forall j\in\mathcal{J}. \label{17}
\end{align}
In \eqref{17}, $\xi_1,\xi_2\in\mathbb{N}^{+}$ are adjustment coefficients. Enforcing strict fairness, i.e., assigning each task to exactly $\frac{N}{J}$ vehicles per global iterationi, is impractical due to heterogeneous vehicle sojourn times and task-dependent training durations. Instead, we restrict the assignment frequency of each task within a range centered at $\frac{N}{J}$, where the left-hand side and right-hand side specify the lower and upper bounds, respectively. This design promotes fairness while accommodating dynamic vehicle availability.

\subsection{Optimization Problem of Our Interest}
We next focus on optimizing the latency of the entire process taken place in each round of global iteration (e.g., the $g^{\text{th}}$ round). \textit{Our major goal is to obtain task scheduling at each vehicle $\overline{x}_{m,n;[g]}^{\left \langle j \right \rangle}$ and training sequence $\lambda_{m,n;[g,k]}$, upon considering mobility of vehicles and task training balance}, which we formulate through the following optimization problem
\begin{equation}
    \bm{\mathcal{P:}}\underset{\overline{x}_{m,n;[g]}^{\left \langle j \right \rangle},~\lambda_{m,n;[g,k]}}{\operatorname{argmin}}\left\{\max_{j\in\mathcal{J},~m\in \mathcal{M}_{[g]}^{\left \langle j \right \rangle}}\{T_{m;[g]}^{\left \langle j \right \rangle}\}\right\} \label{22}
\end{equation}
\vspace{-5mm} 
\begin{subequations}\label{p}{
        \begin{align}
            \text{s.t.}~~~~
            &\text{\eqref{16}, \eqref{117}, \eqref{19}, \eqref{17}} \notag
    \end{align}}    
\end{subequations}


Integrating asynchronous aggregation rule in the CS, in this formulation, we aim to optimize $\overline{x}_{m,n;[g]}^{\left \langle j \right \rangle}$ and $\lambda_{m,n;[g,k]}$ for all the vehicles $n$, ESs $m$, and tasks $j$ so as to minimize the cumulative time required for all the tasks to undergo a complete round of global iteration, which is captured by the objective function, thereby reducing the time cost of the entire VEC-HFL process.

\subsection{Proposed Stagewise Scheduling through Hybrid Evolutionary And gReedy allocaTion (HEART)}

Due to the discrete nature of decision variables, $\bm{\mathcal P}$ is formulated as an NP-hard integer programming (IP) problem. In particular, beyond the binary variable $\overline{x}_{m,n;[g]}^{\left \langle j \right \rangle}$, the task-ordering variable $\lambda_{m,n;[g,k]}$ further increases the combinatorial complexity. Moreover, the hybrid synchronous–asynchronous aggregation, mobility-induced timing constraints, and cross-task fairness requirements jointly complicate the problem structure, rendering classical methods (e.g., branch-and-bound and dynamic programming) computationally prohibitive. RL-based approaches also suffer from unstable convergence as the scale of tasks and vehicles grows. To address this, we propose a stagewise scheduling framework, HEART, which achieves near-optimal performance with efficient runtime.

In HEART, problem $\bm{\mathcal P}$ is solved in two stages. In \textit{Stage 1}, we handle task scheduling, including vehicle assignment and load balancing. We propose a PS-GA hybrid heuristic, where GA crossover and mutation are embedded into PSO to enhance global exploration and avoid premature convergence in high-dimensional search spaces\footnote{Detailed motivations behind the design of hybrid PSO-GA approach and our choice of this specific methodology are elaborated in \textbf{Appendix A}.}.
In \textit{Stage 2}, we optimize the task execution order on each vehicle to reduce latency. A task scheduling combination score is introduced to capture task overlap and upload time, which is maximized via a greedy sequence optimization strategy.

\noindent\textit{Stage 1: Task Scheduling via a Hybrid Evolutionary Approach}

Heuristic methods such as PSO and GA are widely adopted for scheduling and resource allocation due to their strong empirical performance \cite{OffloadedT,DynamicSR, Swarm}. However, for problems such as Stage 1 in this work, using either method alone often leads to premature convergence or entrapment in local optima. PSO tends to suffer from particle clustering and reduced exploration \cite{STATCOM, PSOAA}, while GA is sensitive to parameter settings and may also converge prematurely \cite{MADDPG,Assessment}. To address these limitations, we propose a hybrid PSO–GA that exploits their complementary strengths. Specifically, we introduce a dynamic inertia weight in PSO to improve adaptability, and enhance GA via improved mutation and strengthened constraints to promote exploration. These modifications jointly improve global search capability and reduce the risk of suboptimal convergence.

In the following, we first summarize the PSO modifications in S1.1 and GA modifications in S1.2, and then present the integrated hybrid framework in S1.3. Numerical results further demonstrate that the proposed method outperforms standard heuristics in solution quality.

\textit{S1.1 Overview and sketch of the PSO process:} In our task scheduling framework, conventional PSO performs global search over all vehicles \cite{DynamicSR,STATCOM}. Inspired by flocking behavior, particles iteratively update positions based on individual and collective experience under a fitness function. Although this global design captures system-level objectives, it leads to high complexity and limits per-vehicle adaptability in our setting. Conversely, purely per-vehicle scheduling may overlook global balance. In particular, under limited dwell time, vehicles tend to prioritize short-duration tasks to maximize the number of assignments, which results in long-duration tasks being under-selected. This imbalance slows the progress of global model training and increases overall training time. To address this trade-off, we propose an enhanced PSO that jointly balances global coordination and per-vehicle adaptability. In the following, we detail the proposed design.

\noindent
$\bullet$ \textit{Design of an enhanced fitness function.} To address the above issues, we redesign the fitness function in PSO to better balance local (per-vehicle) and global task distributions (a natural outcome will be to satisfy the task scheduling constraint \eqref{16}, \eqref{117} and \eqref{17}, while reaching a near-optimal value for the objective function in \eqref{22}). Here, we continuously optimize each vehicle’s fitness function to find an appropriate scheduling solution. We also incorporate a \textit{task weight coefficient} $\rho_j$, increasing the reward for assigning tasks with longer training durations. This ensures that shorter-duration tasks do not overshadow more time-intensive tasks. By doing so, we achieve an equitable distribution of tasks across vehicles and maintain a high overall system performance. Through the above two strategies, our design effectively balances scheduling decisions, thus improving both training efficiency and convergence toward the global model accuracy. In particular, we formalize the fitness function of vehicle $n$ under the coverage of ES $m$ in the $g^\text{th}$ global iteration as follows:
\begin{align}
    f_{m,n;[g]}=\begin{cases}-\infty&\hspace{-32mm}\text{if it can not meet \eqref{16}, \eqref{117} and \eqref{17}}\\\sum_{j=1}^J\overline{x}_{m,n;[g]}^{\left \langle j \right \rangle}-(\xi_{3}\sum_{j=1}^J|\psi_j\\~~-\chi|-\sum_{j=1}^J\rho_j|\psi_j-\chi|)&\hspace{-4mm}\text{otherwise} \tag{23}\end{cases}. \label{23}
\end{align}
In \eqref{23}, $\sum_{j=1}^J|\psi_j-\chi|$ serves as a penalty for imbalanced task assignments. Here, $\psi_j$ is a binary variable that indicates whether task $j$ has been assigned (e.g., when $\overline{x}_{m,n;[g]}^{\left \langle j \right \rangle}=1$, $\psi_j=1$, $j\in{J}$), and $\chi$ represents the target number of assignments for each task. Thus, $|\psi_j - \chi|$ measures the discrepancy between the actual assignment and the ideal allocation. Additionally, $\xi_3$ is a balance factor that influences the importance of minimizing this discrepancy. For each vehicle, we define the task scheduling corresponding to the global optimal fitness value $f_{m,n;[g]}^{global}$ as the vehicle's task scheduling (i.e., $\overline{x}_{m,n;[g]}^{global,\left \langle j \right \rangle}$). In addition, this value will be used to update the particle's task scheduling along with the local optimal fitness value $f_{m,n;[g]}^{p,local}$ of each particle. These variables are randomly initialized at the beginning of the PSO process.

\noindent
$\bullet$ \textit{Enhancing the inertia weight.} Traditional PSO uses a constant inertia weight $\pi$ to moderate the velocity of the particles, preventing them from moving too quickly or too slowly. Although this constant weight helps with extreme search bias, it also suffers from several drawbacks: it cannot adapt to dynamic changes in the environment, it is more prone to premature convergence, and it limits the algorithm's exploratory ability. To address these issues, we introduce a linear descent mechanism for updating the inertia weight. By gradually decreasing $\pi$ over the course of the iterations, the swarm initially benefits from a larger momentum that aids in exploring a wide region of the solution space. As the search progresses, the decreasing inertia weight allows the particles to fine-tune their convergence, thus reducing the risk of stagnating in suboptimal solutions. Mathematically, we modify the inertia weight as follows:
\setcounter{equation}{23}
\begin{align}
    \pi(\tau)=\pi_{\max}-\frac{\pi_{\max}-\pi_{\min}}{\tau^{*}}\tau, \label{24}
\end{align}
where, $\pi_{\max}$ and $\pi_{\min}$ define the upper and lower bounds of the inertia weight, respectively, while $\tau^{*}$ is the total number of iterations and $\tau$ represents the current iteration index. Using \eqref{24}, in the initial phase of the optimization process, the inertia weight starts at a high value to promote broad exploration of the search space. As the algorithm advances, the inertia weight gradually decreases, shifting the focus toward more refined local convergence. This adaptive behavior increases the overall efficiency of the algorithm at different stages and aids in identifying an optimal task scheduling plan. 

\noindent
$\bullet$ \textit{Update of velocity and task scheduling.} Next, we update the velocity $v$ in the $\tau^{\text{th}}$ PSO process through $\pi(\tau)$ and the task scheduling $\overline{x}_{m,n;[g]}^{p,\left \langle j \right \rangle}(\tau-1)$ for each particle in the previous iteration, and update the position of the particles accordingly (i.e., the task scheduling $\overline{x}_{m,n;[g]}^{p,\left \langle j \right \rangle}(\tau-1)$). Firstly, the velocity for particle $p$ in the $\tau^{\text{th}}$ itration is given by 
\begin{multline}
    v^{p,\left \langle j \right \rangle}_{m,n;[g]}(\tau)=\pi(\tau-1)v^{p,\left \langle j \right \rangle}_{m,n;[g]}(\tau-1)+\xi_4\{\overline{x}_{m,n;[g]}^{p,local,\left \langle j \right \rangle}-\\
    ~~\overline{x}_{m,n;[g]}^{p,\left \langle j \right \rangle}(\tau-1)\}+\xi_5\{\overline{x}_{m,n;[g]}^{global,\left \langle j \right \rangle}-\overline{x}_{m,n;[g]}^{p,\left \langle j \right \rangle}(\tau-1)\}, \label{25}
\end{multline}
where $\xi_4, \xi_5 \in [0,2]$ capture the social cognitive impact during PSO process, while $\overline{x}_{m,n;[g]}^{p,local,\left \langle j \right \rangle}$ represents the local optimal task scheduling for each particle, and $\overline{x}_{m,n;[g]}^{global,\left \langle j \right \rangle}$ is the global optimal task scheduling of the vehicle. Before the PSO process begins, we will randomly initialize these and update them later. Then, we introduce $\Phi^{p,\left \langle j \right \rangle}_{m,n;[g]}(\tau)$ to represent the probability of the current task being assigned and use it to update the task scheduling of particle $p$ as follows:
\begin{align}
    \Phi^{p,\left \langle j \right \rangle}_{m,n;[g]}(\tau)&=\frac{1}{1+e^{-v^{p,\left \langle j \right \rangle}_{m,n;[g]}(\tau)}}, \label{26}\\
    \overline{x}_{m,n;[g]}^{p,\left \langle j \right \rangle}(\tau)&=\begin{cases}1&\text{if} ~~e_1\leq\Phi^{p,\left \langle j \right \rangle}_{m,n;[g]}(\tau)\\0&\text{otherwise}\end{cases}, \label{27}
\end{align}
where $e_1$ is a random value in the range [0,1]. We obtain $\overline{x}_{m,n;[g]}^{p,\left \langle j \right \rangle}(\tau)$ through \eqref{27} and use it in \eqref{23} to calculate the fitness value $f_{m,n;[g]}^p(\tau)$ of the particle, and decide whether to update $\overline{x}_{m,n;[g]}^{p,local,\left \langle j \right \rangle}$, $\overline{x}_{m,n;[g]}^{global,\left \langle j \right \rangle}$, $f_{m,n;[g]}^{p,local}$, $f_{m,n;[g]}^{global}$ (later concertized in the algorithm description).

Despite the advancements we made to PSO, it may still encounter challenges in global exploration. To address this, we integrate a GA step -- as discussed below --  between particle updates, thereby enhancing global search capabilities. At the same time, the PSO framework bolsters the GA by continuously refining the search for the global optimum. This hybrid approach effectively balances exploration and exploitation, leading to improved task scheduling solutions in complex optimization scenarios of our interest in this work.

\textit{S1.2 Overview and sketch of the GA process:} Traditional GA generally includes steps such as fitness evaluation, selection, crossover, mutation, and other operations. However, because the PSO component in our hybrid system already handles fitness evaluation, we streamline the GA portion to just two steps at the end of each PSO cycle: \textit{(i)} Crossover and \textit{(ii)} Mutation. These operations help diversify the swarm by introducing new task scheduling patterns and reducing the risk of converging prematurely to local optima. 

\noindent
$\bullet$ \textit{Crossover enhancement.} During the task scheduling phase for each vehicle, we sequentially select two adjacent particles, $p$ and $p+1$, correspond to task scheduling $\overline{x}_{m,n;[g]}^{{p,\left \langle j \right \rangle}}(\tau)$, $\overline{x}_{m,n;[g]}^{{p+1,\left \langle j \right \rangle}}(\tau)$. We then apply crossover and mutation on these allocations repeatedly. Specifically, for crossover, $p$ and $p+1$ serve as parent solutions, and we define a genetic length $J$ along with a randomly chosen crossover point $r$ such that $1<r<J$. Exchanging the segments of the parent allocations at the crossover point creates two offspring, $c_1$ and $c_2$. Thus, new task scheduling schemes are generated, which expand the exploration space and reduce the likelihood of trapping the algorithm in suboptimal solutions. We define all task scheduling for $p$, $p+1$ as a vector $X^p_{m,n;[g]}(\tau)$ and $X^{p+1}_{m,n;[g]}(\tau)$ (e.g., for $p$ during the $\tau^\text{th}$ iteration, if the particle implies the scheduling of tasks 1, 3, and 4 out of the four available tasks in the system, we will have the vector $X^p_{m,n;[g]}(\tau)=[\text{1, 0, 1, 1}]$). To better describe this crossover process, we use $[i:j]$ to represent the elements from $i$ to $j-1$ in the vector, e.g., considering $X^p_{m,n;[g]}(\tau)=[\text{1, 0, 1, 1}]$, we have $X^p_{m,n;[g]}(\tau)[\text{0:2}]=\text{1, 0}$. We then obtain the offspring inherit allocations from the parents according to the following crossover rules:
\begin{align}
    X_{m,n;[g]}^{p,c_1}(\tau)&=[X_{m,n;[g]}^{p}(\tau)[0:r],X_{m,n;[g]}^{{p+1}}(\tau)[r:J]], \label{28}\\
    X_{m,n;[g]}^{{p+1},c_2}(\tau)&=[X_{m,n;[g]}^{{p+1}}(\tau)[0:r],X_{m,n;[g]}^{{p}}(\tau)[r:J]], \label{29}
\end{align}
where $ X_{m,n;[g]}^{p,c_1}(\tau), X_{m,n;[g]}^{{p+1},c_2}(\tau)$ represents the vector of task scheduling for the offsprings, and the single task scheduling $\overline{x}_{m,n;[g]}^{{p,\left \langle j \right \rangle,c_1}}(\tau)$ and $\overline{x}_{m,n;[g]}^{{p+1,\left \langle j \right \rangle,c_2}}(\tau)$ are the $j^\text{th}$ elements of these vectors. 

\noindent
$\bullet$ \textit{Mutation enhancement.} For the newly generated offspring $c_1$ and $c_2$ in \eqref{28} and \eqref{29}, we further expand the solution space for each particle by applying mutation operations: whether these offsprings also undergo mutation is determined by a mutation rate $\varphi$. Nevertheless, a constant $\varphi$ is less flexible and cannot accommodate different exploration needs across various search stages. To maintain the diversity of the solution space, we adopt an adaptive mutation rate $\varphi(\tau)$ that evolves with the iteration index $\tau$. Unlike conventional fixed rates, our specific functional form, a logarithmic decay model defined as $\varphi(\tau)=\frac{\varphi_{\max}}{1+\log(1+\tau)}$, is empirically determined through extensive ablation studies\footnote{We evaluated various decay trajectories (including linear, exponential, and logarithmic) and different initial values of $\varphi_{\max}$ to balance exploration and convergence. The detailed comparative analysis and experimental justifications for this selection are provided in \textbf{Appendix B}.}. During the initial phase, a higher mutation rate promotes broad global exploration, reducing the risk of convergence to local optima. As the number of iterations increases, $\varphi(\tau)$ decreases, aiding the algorithm in fine-tuning its convergence process. To introduce additional randomness and broaden the global search, we compare a random factor against $\varphi(\tau)$ to determine if a mutation occurs. For example, for offspring $c_1$, the mutation decision is governed by the following rule:
\begin{align}
    \quad\overline{x}_{m,n;[g]}^{{p,\left \langle j \right \rangle}}(\tau)\longleftarrow\begin{cases} 1-\overline{x}_{m,n;[g]}^{{p,\left \langle j \right \rangle},c_1}(\tau)& \textrm{if}~~ e_2<\varphi(\tau)\\[2ex]\overline{x}_{m,n;[g]}^{{p,\left \langle j \right \rangle},c_1}(\tau)&\text{otherwise}\end{cases}, \label{30}
\end{align}
where $e_2$ is a random value in the range [0, 1]. Although the GA stage can generate novel task scheduling schemes, these new allocations may cause the total time, energy and task balance required to complete all assigned tasks to exceed the vehicle's available time under ES $m$'s coverage, as constrained by \eqref{16}, \eqref{17} and \eqref{117}, which we will take into account in our overarching method described next.
\begin{algorithm}[htb!]
	{
		
		\small
		
		\caption{Design of hybrid heuristic by integrating improved PSO and GA in Stage 1}
		
		\SetKwInOut{Input}{Input}\SetKwInOut{Output}{Output}
		
		\Input{$t_{m,n;[g,k,h]}^{\langle j\rangle}$, $t_{m,n;[g]}^{stay}$, $p^{*}$, $\tau^{*}$, $\rho_j$, $\chi$, $\pi_{\max}, \pi_{\min}$;}
		
		\Output{$\overline{x}_{m,n;[g]}^{global,\left \langle j \right \rangle}$;}
		
		{\bf{Initialization :}} 
		$\overline{x}_{m,n;[g]}^{p,\left \langle j \right \rangle}(0)$, $v_{m,n;[g]}^{p,\left \langle j \right \rangle}(0)$, $\overline{x}_{m,n;[g]}^{p,local,\left \langle j \right \rangle}$, $\overline{x}_{m,n;[g]}^{global,\left \langle j \right \rangle}$, $f_{m,n;[g]}^{p,local}$, $f_{m,n;[g]}^{global}$;

            \ForEach{$n \in \mathcal{N}$}{

                \For{$\tau=\{1,2,\cdots,\tau^{*}\}$}{

                     \textit{\textbf{PSO process:}}

                     Dynamically adjust $\pi(\tau)$ based on \eqref{24}

                     \For{$p=\{1,2,\cdots,p^{*}\}$}{

                        Obtain the task scheduling $\overline{x}_{m,n;[g]}^{p,\left \langle j \right \rangle}(\tau)$ for the current particle based on \eqref{25}-\eqref{27}
                        
                        Based on $\overline{x}_{m,n;[g]}^{p,\left \langle j \right \rangle}(\tau)$ and \eqref{23} calculate the fitness $f_{m,n;[g]}^p(\tau)$ 

                        \If{$f_{m,n;[g]}^p(\tau) > f_{m,n;[g]}^{p,local}$}{

                            $\overline{x}_{m,n;[g]}^{p,local,\left \langle j \right \rangle} \xleftarrow[]{} \overline{x}_{m,n;[g]}^{p,\left \langle j \right \rangle}(\tau)$,
                            $f_{m,n;[g]}^{p,local} \xleftarrow[]{} f_{m,n;[g]}^p(\tau)$

                            \If{$f_{m,n;[g]}^p(\tau) > f_{m,n;[g]}^{global}$}{
                        
                            $\overline{x}_{m,n;[g]}^{global,\left \langle j \right \rangle} \xleftarrow[]{} \overline{x}_{m,n;[g]}^{p,\left \langle j \right \rangle}(\tau)$,$f_{m,n;[g]}^{global} \xleftarrow[]{} f_{m,n;[g]}^p(\tau)$
                        }
                        }
                
                     }

                    \textit{\textbf{GA process:}}

                    \For{$p$ = $1$ to $(p^*-1)$}{

                        Crossover enhancement: based on \eqref{28} and \eqref{29} get $\quad\overline{x}_{m,n;[g]}^{{p,\left \langle j \right \rangle,c_1}}(\tau)$, $\quad\overline{x}_{m,n;[g]}^{{p+1,\left \langle j \right \rangle, c_2}}(\tau)$

                        Mutation enhancement: based on \eqref{30} get $\quad\overline{x}_{m,n;[g]}^{{p,\left \langle j \right \rangle}}(\tau)$, $\quad\overline{x}_{m,n;[g]}^{{p+1,\left \langle j \right \rangle}}(\tau)$

                        \If{task scheduling  $\quad\overline{x}_{m,n;[g]}^{{p,\left \langle j \right \rangle}}(\tau)$, $\quad\overline{x}_{m,n;[g]}^{{p+1,\left \langle j \right \rangle}}(\tau)$ do not meet constraint \eqref{16}, \eqref{117}, \eqref{17}}{

                                Exclude the task with the longest training time from $\quad\overline{x}_{m,n;[g]}^{{p,\left \langle j \right \rangle}}(\tau)$, $\quad\overline{x}_{m,n;[g]}^{{p+1,\left \langle j \right \rangle}}(\tau)$  until the constraint is met
                            
                            }
                    }                 
                }
            }
    }
\end{algorithm}

\textit{S1.3 Detail of our proposed hybrid heuristic approach:} Our proposed method combines the above PSO and GA processes. Additional, we provide the description of our method in \textbf{Appendix C}, where its steps are summarized in Alg. 2.

\noindent \textit{Stage 2: Optimizing Task Training Rank via a Greedy Algorithm}

In our synchronous aggregation process at the ES, each vehicle uploads its locally trained model for task $j$ upon completion, and aggregation is performed only after all models are received. However, an improper training rank for task $j$ may delay edge updates, increasing training time and causing model staleness.
Therefore, selecting an appropriate task rank is essential to maximize temporal overlap among vehicles processing the same task and minimize \eqref{22}. However, multiple near-optimal sequence configurations and the coupling of vehicle mobility with heterogeneous model sizes make the ranking decision non-trivial due to varying transmission times. These challenges motivate a low-complexity scheduling strategy for task ranking to maximize overlap while reducing aggregation delay.

\textit{S2.1 Determining the task training sequences/ranks:} To determine the training rank of vehicles for their assigned tasks, we introduce an \emph{aggregate-score} that jointly captures task overlap and model upload time. The optimal sequence for each vehicle is defined as the one maximizing this score. This formulation resolves ambiguities from multiple feasible schedules while accounting for key factors affecting scheduling quality. In terms of \emph{task overlap}, once the set of tasks that vehicle \(n\) must train under ES \(m\) in the \(g^{\text{th}}\) global iteration is determined in the above-described Stage 1 of our method, we first randomly initialize its training rank \(\lambda_{m,n;[g,k]}\). To compute the overlap across the tasks, we define an \emph{overlap indicator function} \(\delta\bigl(\lambda_{m,n;[g,k]}^{\langle j\rangle},\,\lambda_{m,n^\prime;[g,k]}^{\langle j\rangle}\bigr)\) for each task \(j\). This function equals 1 if task \(j\) for vehicles \(n\) and \(n'\) that are both covered by ES $m$  has the same rank (i.e., \(\lambda_{m,n;[g,k]}^{\langle j\rangle} = \lambda_{m,n';[g,k]}^{\langle j\rangle}\)), and 0 otherwise. For example, consider a scenario in which there are four tasks and two vehicles, \(n\) and \(n'\). Vehicle \(n\) is assigned tasks by following the sequence/rank \([\text{1, 3, 4}]\), while vehicle \(n^\prime\) is assigned tasks with \([\text{2, 3}]\). In this situation, only the second task indices overlap (both vehicles train task 3 in their second slots), making \(\delta(\lambda_{m,n;[g,k]}^{\langle3\rangle},\,\lambda_{m,n^\prime;[g,k]}^{\langle3\rangle}) = \text{1}\), while the indicators of all other tasks are 0. During the \(k^{\text{th}}\) edge iteration, the overlap score for task \(j\), denoted as \(\mathcal{S}_{m;[g,k]}^{lap,\langle j\rangle}\), is then computed as the sum of the above indicator functions over all vehicles within each ES \(m\)'s coverage:  
\begin{align}
    \mathcal{S}_{m;[g,k]}^{lap,\left\langle j\right\rangle}=\sum_{n,n^\prime\in\mathcal{N}_{m;[g]}^{cov},n\neq n'}\delta( \lambda_{m,n;[g,k]}^{\left\langle j\right\rangle},\lambda_{m,n^{\prime};[g,k]}^{\left\langle j\right\rangle}). \label{31}
\end{align}

In principle, maximizing task overlap is desirable, as it enables temporally aligned training of the same task across vehicles, facilitating timely model aggregation and distribution at the ES, reducing \textit{case (ii)} in Section III-A, and thereby decreasing idle time. However, when multiple task sequences achieve similar maximum overlap, selecting the optimal one becomes non-trivial. Moreover, model upload time in \textit{case (ii)} must also be considered in sequence design. For instance, when a vehicle is moving away from the ES, prioritizing tasks with larger model sizes allows transmission under better channel conditions, while smaller tasks can be deferred. To capture this effect, we compute the model upload time $t_{m,n;[g,k]}^{V2E,\left \langle j\right \rangle}$ based on (8), which depends on the distance to ES $m$ and model size. Accordingly, we define the \textit{model upload score} $\mathcal{S}^{up,\left \langle j \right \rangle}_{m,n;[g,k]}$ as a function of the upload time:
\begin{align}
    \mathcal{S}^{up,\left \langle j\right \rangle}_{m,n;[g,k]}=\xi_{6}/t_{m,n;[g,k]}^{V2E,\left \langle j\right \rangle}, \label{32}
\end{align}
where $\xi_{6}$ is a positive coefficient, which adjusts the model upload score in \eqref{32} to the same level as the overlap score given by \eqref{31}. This score is then combined with the task overlap score for each task $j$ to get its aggregate-score, which is given by
\begin{align}
    \mathcal{S}_{m;[g,k]}^{all,\left\langle j\right\rangle}=\xi_{7}\mathcal{S}_{m;[g,k]}^{lap,\left\langle j\right\rangle}+(1-\xi_{7})\sum_{n\in\mathcal{N}_{m;[g]}^{Act,\left\langle j\right\rangle}}{\mathcal{S}_{m,n;[g,k]}^{up,\left\langle j\right\rangle}}, \label{33}
\end{align}
where \(\xi_7\) is a balance factor that weighs task overlap against model upload time. The aggregate-score determines the training order of tasks. However, jointly scheduling all tasks under ES \(m\) is combinatorial and computationally intractable, and would significantly increase the complexity of PSO and GA if embedded into Stage 1. To address this, we propose a low-complexity greedy algorithm that sequentially ranks tasks based on their aggregate-scores, enabling real-time implementation. 

\textit{S2.2 Detail of the greedy-based task ranking solution:}
For training sequence/rank optimization, it is necessary to maximize the aggregate-score $\mathcal{S}_{m;[g,k]}^{all,\left\langle j\right\rangle}$ under each ES $m$ coverage for each edge iteration $k$. Subsequently, maximization of aggregate-score can be decoupled and conducted in parallel across the ESs and edge iterations. As a result, we focus on a specific ES $m$ and edge iteration $k$ in the following. Details of the entire process are provided in \textbf{Appendix D} (see Alg. 3).
\begin{algorithm}[htb!]
	{
		
		\small
		
		\caption{Greedy-based algorithm for training rank determination in Stage 2}
		
		\SetKwInOut{Input}{Input}\SetKwInOut{Output}{Output}
		
    \Input{$\overline{x}_{m,n;[g]}^{global,\left \langle j \right \rangle}$;}
		
		\Output{$\lambda_{m,n;[g,k]}$;}
		
		{\bf{Initialization :}} 
		Highest aggregate-score $\mathcal{S}_{m;[g,k]}^{*}=0$, $\lambda_{m,n;[g,k]}=\emptyset$, $q=0$, $\mathcal{J}^{\prime}=\mathcal{J}$;

            \ForEach{ $m\in\mathcal{M}$ in parallel}{

                \For{$k=\{1,2,\cdots,K^{\left \langle j \right \rangle}\}, where~j \in \mathcal{J}$}{
                    
                    \For{$j\in\mathcal{J^{\prime}}$}{

                        Calculate $\mathcal{S}_{m;[g,k]}^{lap,\left\langle j\right\rangle}$ based on $\overline{x}_{m,n;[g]}^{global,\left \langle j \right \rangle}$ and \eqref{31}

                        \For{$n \in \mathcal{N}_{m;[g,k]}^{\left \langle j \right \rangle}$}{

                            Use (8) to obtain the model upload time $t_{m,n;[g,k]}^{V2E,\left \langle j\right \rangle}$ 
                                        
                                 Calculate $\mathcal{S}_{m,n;[g,k]}^{up,\left \langle j \right \rangle}$ based on \eqref{32}
                                
                            }

                        Calculate $\mathcal{S}_{m;[g,k]}^{all,\left \langle j \right \rangle}$ based on \eqref{33}

                        \If{$\mathcal{S}_{m;[g,k]}^{all,\left \langle j \right \rangle} > \mathcal{S}_{m;[g,k]}^{*}$}{

                            $\mathcal{S}_{m;[g,k]}^{*} \xleftarrow{} \mathcal{S}_{m;[g,k]}^{all,\left \langle j \right \rangle}$ 

                            $q \xleftarrow{} j$

                        }
                    }

                        \For{$n \in \mathcal{N}_{m;[g,k]}^{\left \langle q \right \rangle}$}{

                                Add task $q$ at the end of training sequence $\lambda_{m,n;[g,k]}$
                            
                            }

                         $\mathcal{J^{\prime}} \xleftarrow{} \mathcal{J^{\prime}} \setminus \{q\}$

                         $\mathcal{S}_{m;[g,k]}^{*}=0$ 
        }
    }
}
\end{algorithm}

\subsection{Theoretic Justification and Convergence Analysis of HEART}
To ensure training stability under the hybrid synchronous-asynchronous mechanism, we analyze the impact of gradient staleness induced by asynchronous cloud aggregation on global convergence. While edge synchronization reduces local variance, the cloud layer manages the staleness-accuracy trade-off. We further compare the proposed method with fully synchronous HFL (Sync-HFL). A detailed convergence analysis, including assumptions, derivations, and comparison with Sync-HFL, is provided in \textbf{Appendix E.}

\section{Numerical Evaluations}
\noindent The VEC-HFL architecture considered in our simulations contains 1 CS, 5 ESs, and 25 to 135 vehicles depending on the specific simulation setup (i.e., 25, 50, 85 and 135), where the CS with the coverage of 5 km is the radius in located at the center of the area. We consider an area with 4 roads and 4 intersections where an ES with the coverage of 1 km is located at each intersection. Each vehicle maintains an effective average speed ranging from 10 to 30 m/s along the road segments to model the speed changes of vehicles in real-world scenarios to the maximum extent possible (e.g., traffic jams, sudden stops, acceleration, etc.), and at each intersection it takes a turn at a direction chosen randomly\cite{MOBFL,Multiagent12}. To capture diverse training tasks, we consider 4 different ML tasks: CIFAR-10 (trained on VGG16 neural network \cite{VGG16}), MNIST (trained on a convolutional neural network with 4 layers \cite{HiFlash}), Driver Yawning (trained on ResNet-18 neural network \cite{Resnet18}) and 20 Newsgroups (trained on LSTM neural network \cite{LSTM}). In particular, CIFAR-10 and MNIST are colored and grayscale image datasets for ten-category image classification, respectively\cite{Privacy,HiFlash}, Driver Yawning represents a dataset for yawning detection during driving\cite{JHPFA}, while 20 Newsgroups is a text dataset in natural language processing, with data spanning 20 distinct topics. The data rates of vehicle-to-edge and edge-to-vehicle communications are set according to the channel modeling in \cite{Platoon}. Other key parameters used in the simulations are listed in Table II\footnote{The local and edge iterations (i.e., $H^{\left \langle j\right \rangle}$ and $K^{\left \langle j\right \rangle}$) for each task are optimized following the schemes in \cite{optimes} to achieve peak performance.}.
\begin{table}[!t]
	\small
	\caption{Parameter settings\cite{luo2020hfel,KHEF}}
	\centering
	\begin{tabular}{|>{\centering\arraybackslash}m{2.3cm}|>{\centering\arraybackslash}m{1.0cm}|@{\hskip 3pt}|>{\centering\arraybackslash}m{2.3cm}|>{\centering\arraybackslash}m{1.0cm}|}
		\hline
		\rowcolor{verylightgray}
		\bf{Parameter} & \bf{Value} & \bf{Parameter} & \bf{Value}\\
		\hline
		\hline
		Number of  training tasks $J$ & 4-25 &	Learning rate $\eta^{\left\langle j\right\rangle}$ &  0.001-0.005\\
		\hline
		\rowcolor{veryverylightblue}
		Local iterations $H^{\left \langle j\right \rangle}$ & 4-6 & Edge iterations $K^{\left \langle j\right \rangle}$  & 8-10\\
		\hline
		Number of CPU cycles $c^{\langle j\rangle}_{n}$ & 	 20-30	 &  Clock frequency	$f^{\langle j\rangle}_{m,n}$ &	[1,10] GHz\\
		\hline
        \rowcolor{veryverylightblue}
        Number of particles $p^*$ & 30 & Iteration times $\tau^{*}$ & 100 \\
        \hline
        Crossover point $r$ & 2 & Initial mutation rate $\varphi_{\max}$ & 0.3\\
        \hline
	\end{tabular}
    \vspace{-5mm}
\end{table}

We evaluate the system performance from three key perspectives: \textit{(i)} balanced task training; \textit{(ii)} model convergence for  different tasks; and \textit{(iii)} time and energy efficiency. Furthermore, to provide a comprehensive evaluation, two categories of benchmark methods are considered: \textit{i) General  baselines}, which focus on fundamental scheduling and ranking mechanisms (i.e., \textit{TSSO}, \textit{TSPSO}, \textit{TSGA} and \textit{TSGD}); and \textit{ii) Task-relationship-aware baselines}, including \textit{FedSame}, \textit{FedSS}, and \textit{FM$^2$}, which specifically leverage task correlations or shared representations for Federation Multi-Task Learning (FMTL). The detailed configurations are as follows.

\noindent 
$\bullet$ \textbf{\textit{Two-Stage Stochastic Optimization (TSSO):}} Randomly scheduling training tasks under time constraints and assigns a random seed to each task, determining the training rank of the task. 

\noindent 
$\bullet$ \textit{\textbf{Two-Stage Particle Swarm Optimization (TSPSO):}} Task scheduling and training rank are determined by traditional PSO under time constraints and task training balance \cite{DynamicSR}. 

\noindent 
$\bullet$ \textbf{\textit{Two-Stage Genetic Algorithm (TSGA):}} Task scheduling and training rank and are determined by traditional GA under time constraints and task training balance\cite{Swarm, OffloadedT}. 

\noindent 
$\bullet$ \textbf{\textit{Two-Stage Greedy Algorithm (TSGD):}} On the premise of ensuring that the task training time does not exceed the vehicles' dwell times, the task scheduling score is calculated for each vehicle based on the task weight coefficient $\rho_j$ and the number of times it is allocated. The task with a higher score is selected and assigned to the vehicle, and then the task is removed. The above operation continues until the task scheduling is completed. Afterwards, Alg. 3 is employed to further refine the solution \cite{tang2023fedml}.

\noindent 
$\bullet$ \textbf{\textit{FedSame:}} Adopting a Bayesian similarity-aware framework to dynamically model inter-task correlations through Beta distribution updates \cite{FedSame}.

\noindent 
$\bullet$ \textbf{\textit{FedSS:}} Employing a sparse sharing scheme to implement a clustered FMTL mechanism, focusing on optimizing communication and computation efficiency during multi-task joint training \cite{FedSS}.

\noindent 
$\bullet$ \textbf{\textit{FM$^2$:}} Introducing a dual management strategy that utilizes task-related dual variables to balance shared and personalized representations, leveraging distributed optimization for scalable model updates \cite{FM2}.

To underscore the advantages of our HEART compared to lightweight/general baselines, we provide a comprehensive analysis of computational complexity and decision-making latency in \textbf{Appendix F}.

\begin{figure}[htbp]
    \vspace{-3mm}
	\centering
	\subfigure{
		\includegraphics[trim=0cm 0cm 0cm 0cm, clip, width=0.9\columnwidth]{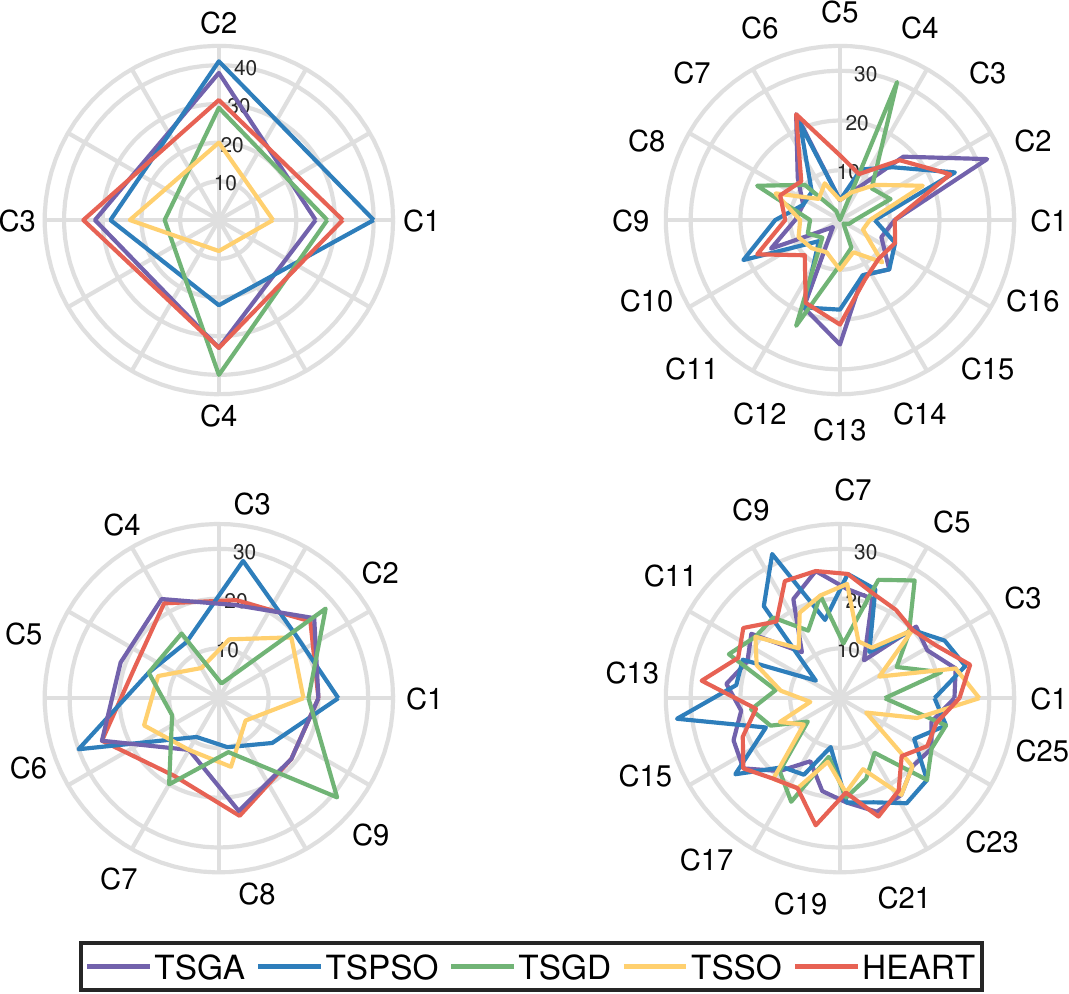}
	}
	\caption{The average number of times that a task has been executed on vehicles for training across different methods. Top-left plot: 4 tasks where C1= CIFAR task; C2= MNIST task; C3= Driver Yawning task; C4= 20 Newsgroups task; Bottom-left plot, Top-right plot and Bottom-right plot: represent 9, 16, and 25 tasks, respectively, all generated by combining varied data volumes and label partitions based on C1-C4.}
    \vspace{-3mm}
\end{figure}

\noindent \textbf{Evaluation of Task Execution Balance.} We evaluate task execution balance by calculating the average task execution frequency per global iteration across scenarios with 4, 9, 16, and 25 tasks\footnote{For the 4-task setting, each vehicle possesses an equal distribution of all data labels. In scenarios with 9, 16, and 25 tasks, we introduce data heterogeneity by assigning random subsets of labels and varying sample sizes to each vehicle (e.g., tasks 7 and 9 use CIFAR on CNN with only 5-7 labels and randomized sample counts).}. As illustrated in Fig. 3, traditional \textit{TSPSO} and \textit{TSGA} tend to converge toward local optima, resulting in skewed task distributions. \textit{TSGD} exhibits suboptimal performance (notably for Task 3) as its sequential search lacks a global backtracking mechanism, while the random selection-based \textit{TSSO} yields the poorest results. In contrast, our proposed \textit{HEART} framework consistently achieves superior distribution balance through its specialized scheduling constraints. To further validate robustness, we applied varied data partitioning across 9, 16, and 25 tasks scenarios, rendering unique training times for each task. The results indicate that \textit{HEART} effectively maximizes throughput while minimizing variance as complexity increases. Specifically, in the 16-task scenario, \textit{HEART} executes 9 more tasks than the strongest baseline with a 67.69\% reduction in variance; in the 25-task scenario, it completes 19 additional tasks and reduces variance by 73.08\%\footnote{Note that task relation-aware baselines, which exclude allocation or sequence optimization, are evaluated separately regarding convergence, time, and energy costs.}.

\begin{figure*}[!t]
    \centering
    \subfigure[]{
        \includegraphics[trim=0.2cm 0cm 0.2cm 0.2cm, clip, width=0.45\columnwidth]{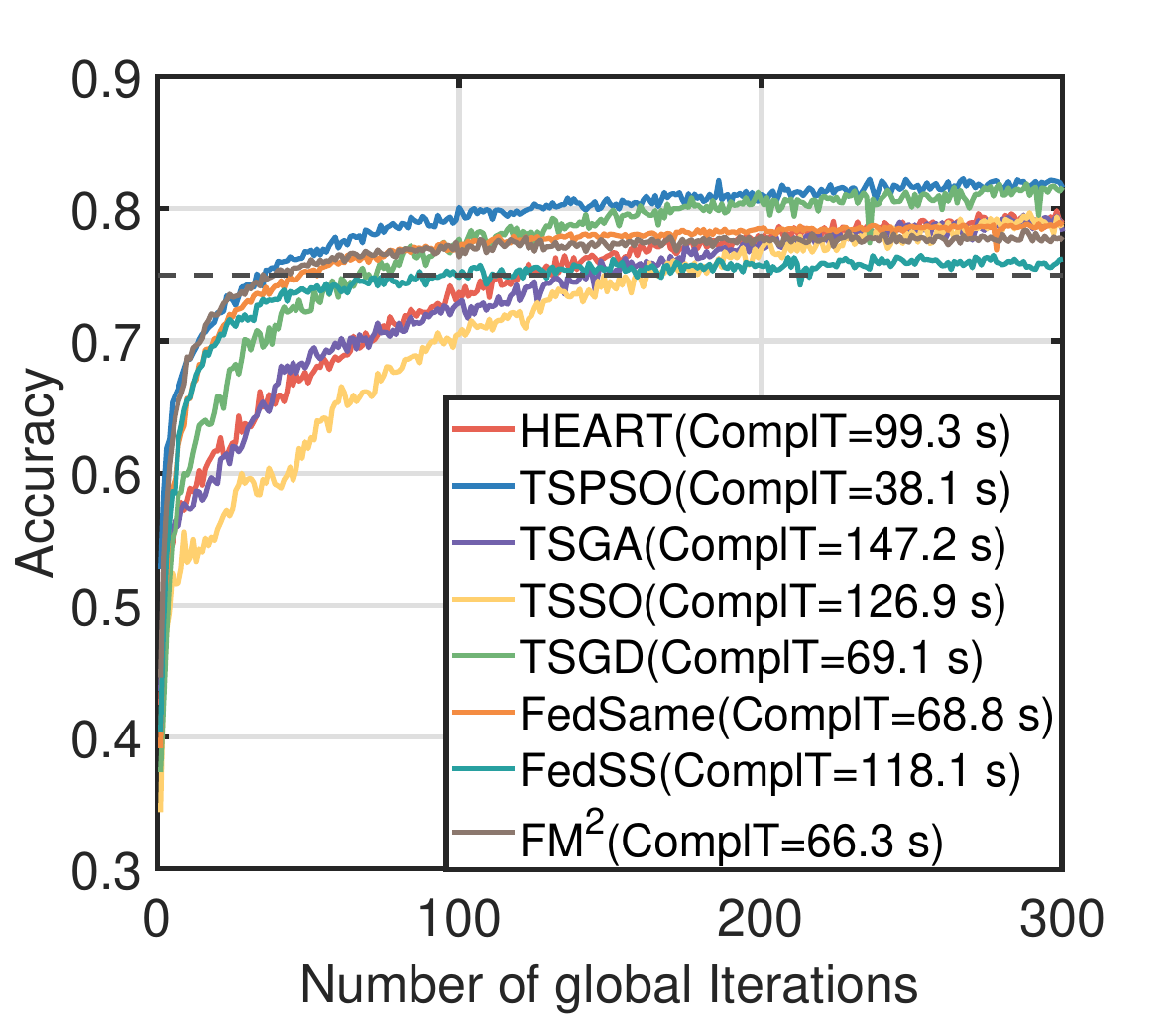}
    }
    \hspace{0.01\columnwidth} 
    \subfigure[]{
        \includegraphics[trim=0.2cm 0cm 0.2cm 0.2cm, clip, width=0.45\columnwidth]{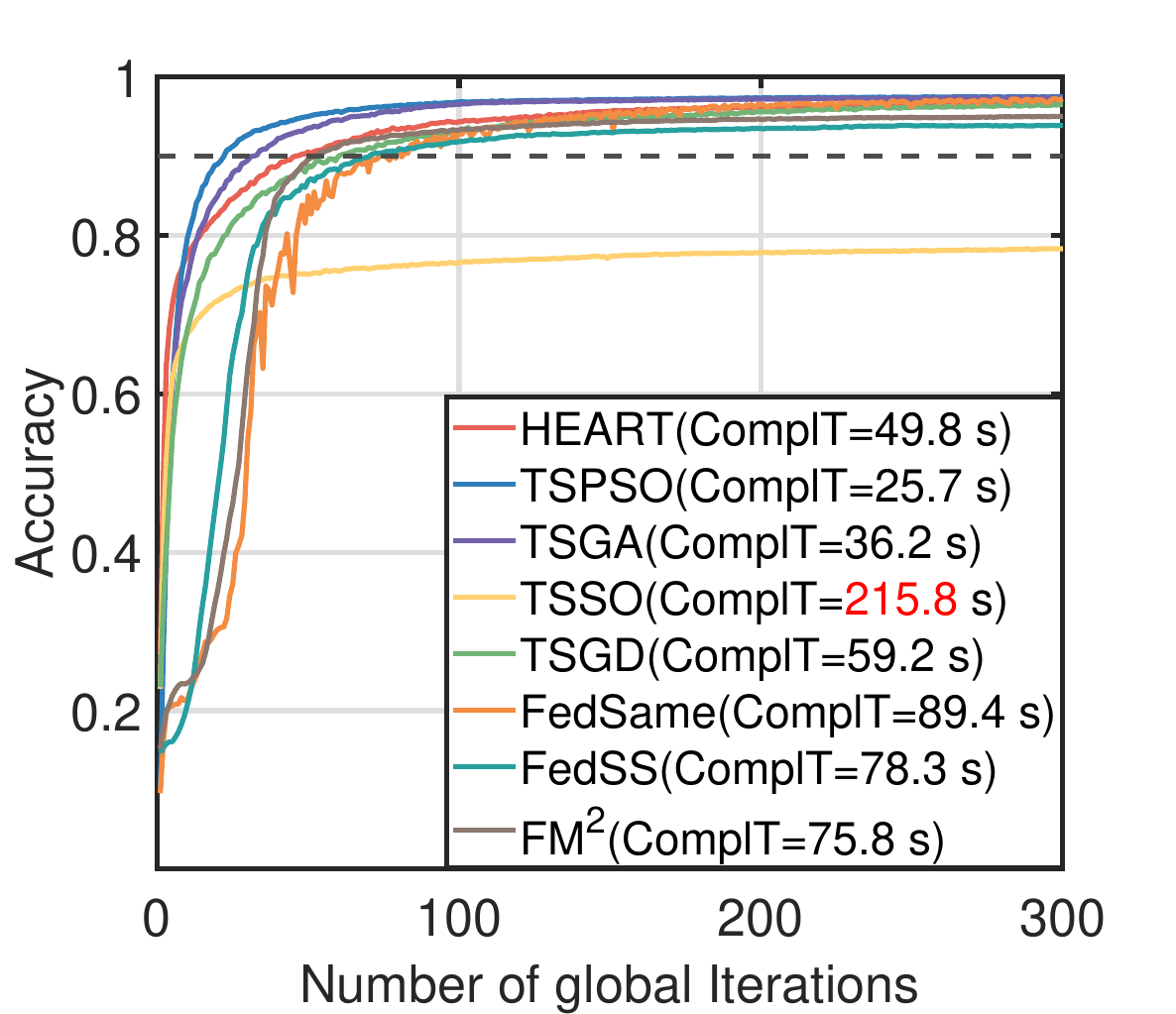}
    }
    \subfigure[]{
        \includegraphics[trim=0.2cm 0cm 0.2cm 0.2cm, clip, width=0.45\columnwidth]{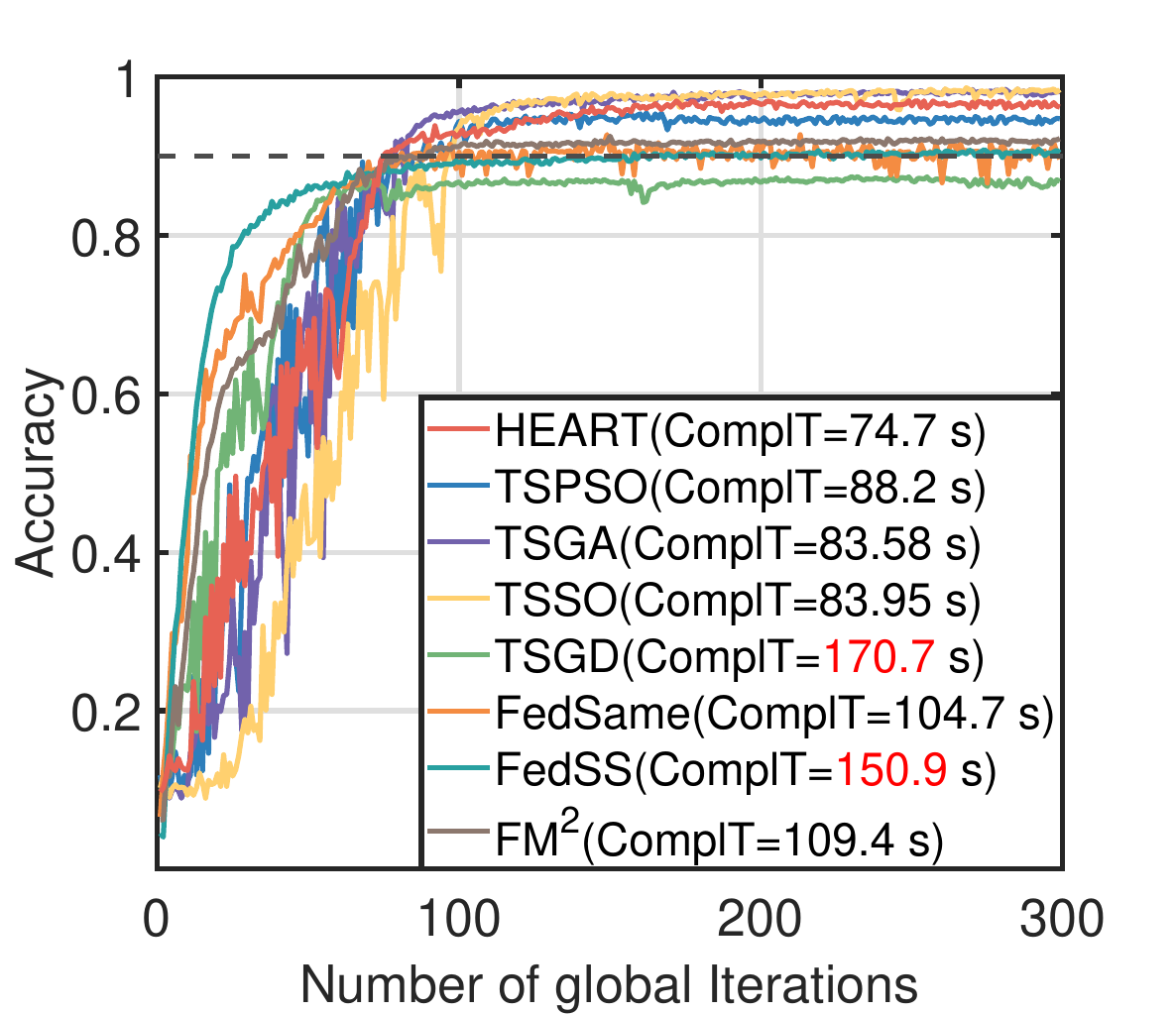}
    }
    \hspace{0.01\columnwidth} 
    \subfigure[]{
        \includegraphics[trim=0.2cm 0cm 0.2cm 0.2cm, clip, width=0.45\columnwidth]{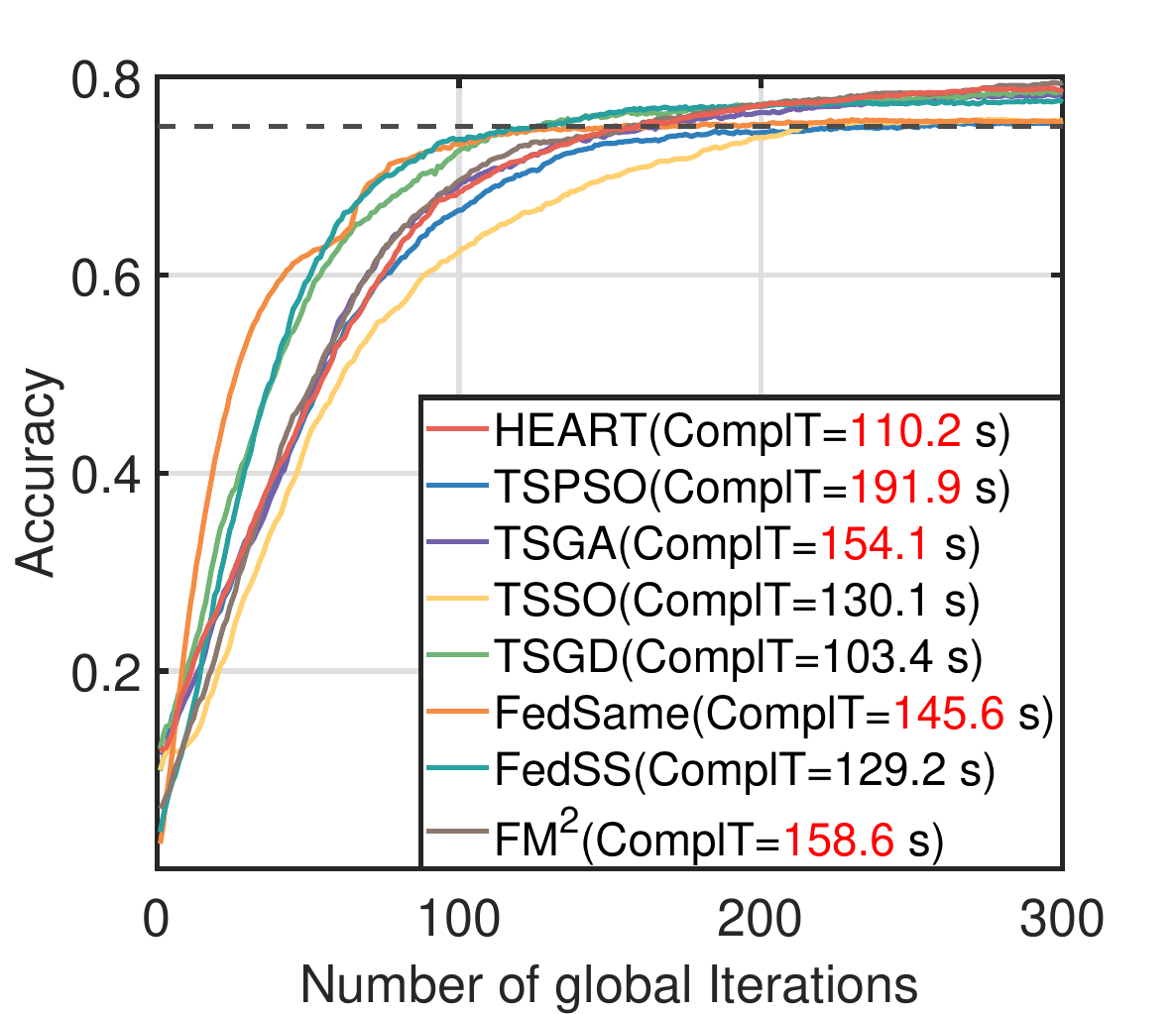}
    }
    
    \caption{Test accuracy of the global models for different tasks under a 50-vehicle network: (a) CIFAR; (b) MNIST; (c) Driver Yawning; (d) 20 Newsgroups.}
    \vspace{-3mm}
    \label{fig_2}
\end{figure*}

\begin{figure*}[!t]
    \centering
    \subfigure[]{
        \includegraphics[trim=0.2cm 0cm 0.2cm 0.2cm, clip, width=0.45\columnwidth]{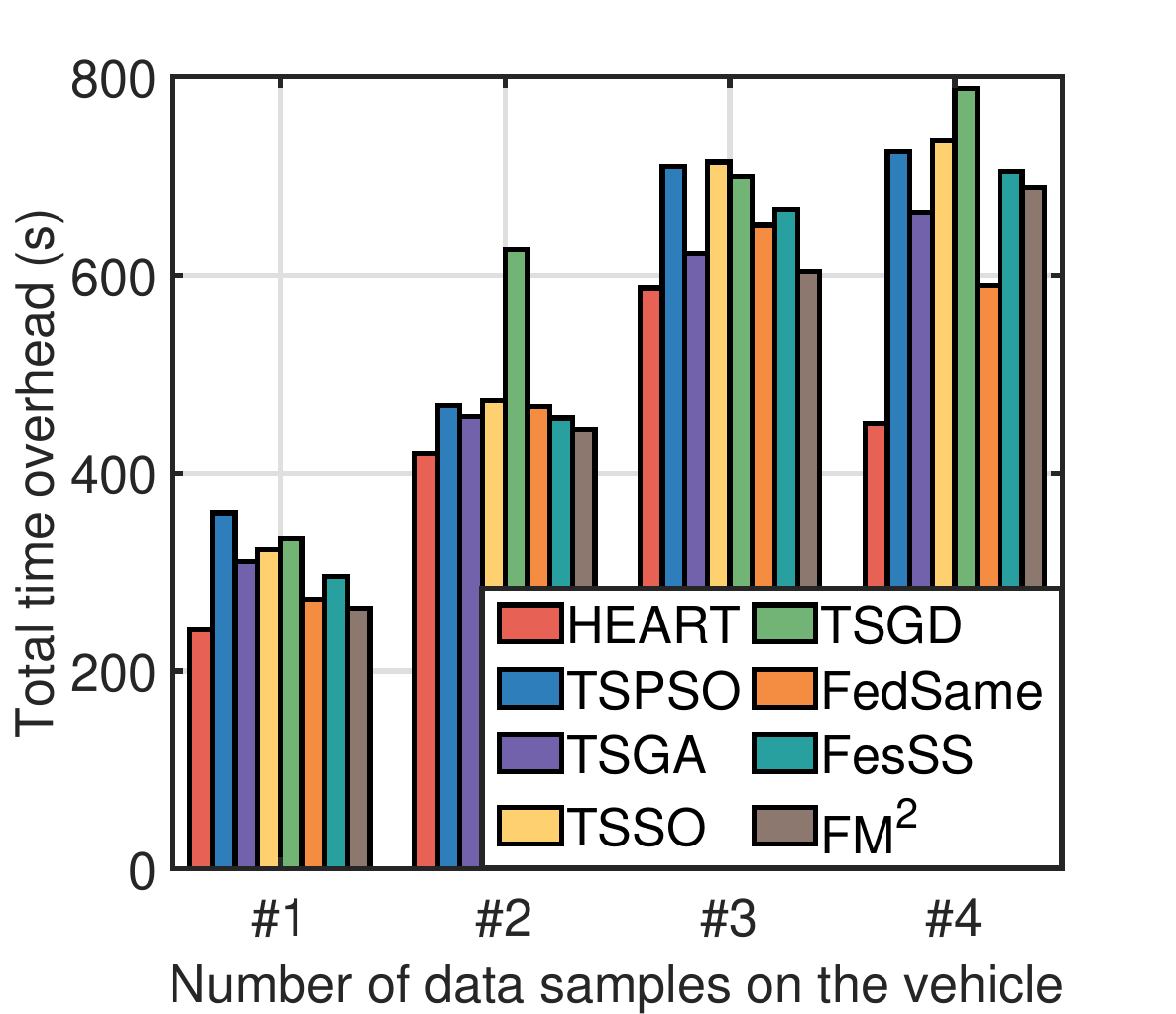}
    }
    \hspace{0.01\columnwidth} 
    \subfigure[]{
        \includegraphics[trim=0.2cm 0cm 0.2cm 0.2cm, clip, width=0.45\columnwidth]{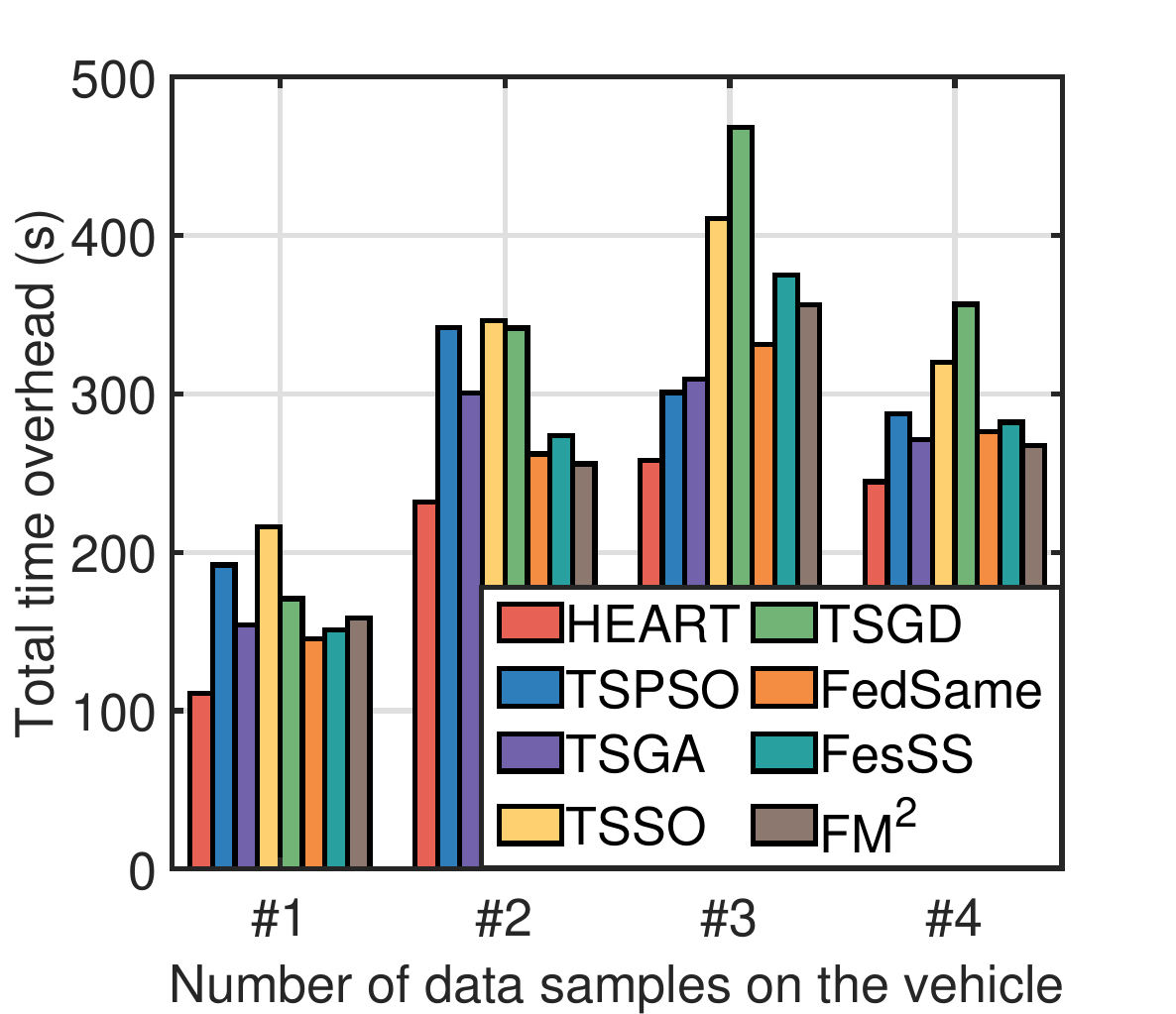}
    }
    \subfigure[]{
        \includegraphics[trim=0.2cm 0cm 0.2cm 0.2cm, clip, width=0.45\columnwidth]{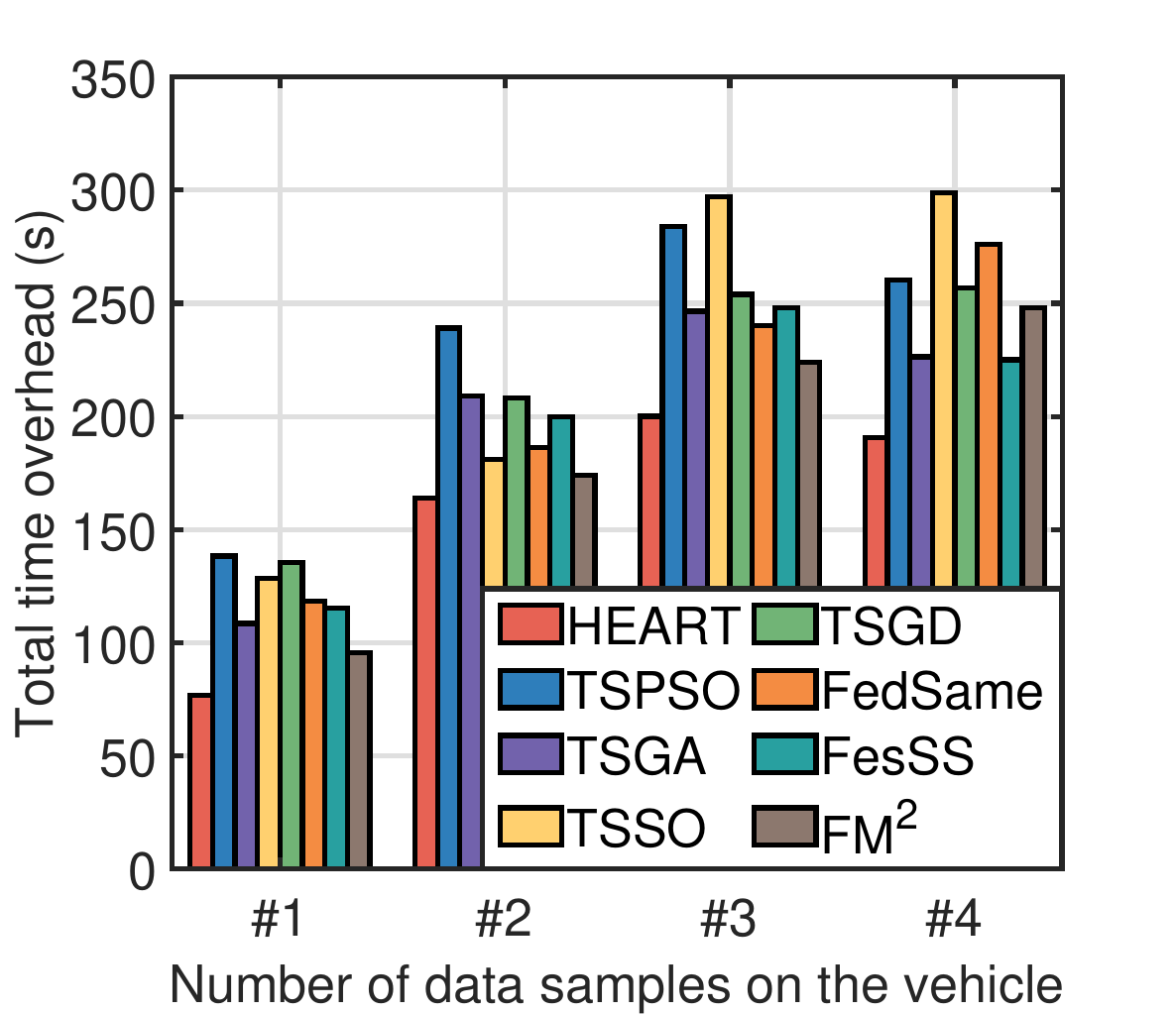}
    }
    \hspace{0.01\columnwidth} 
    \subfigure[]{
        \includegraphics[trim=0.2cm 0cm 0.2cm 0.2cm, clip, width=0.45\columnwidth]{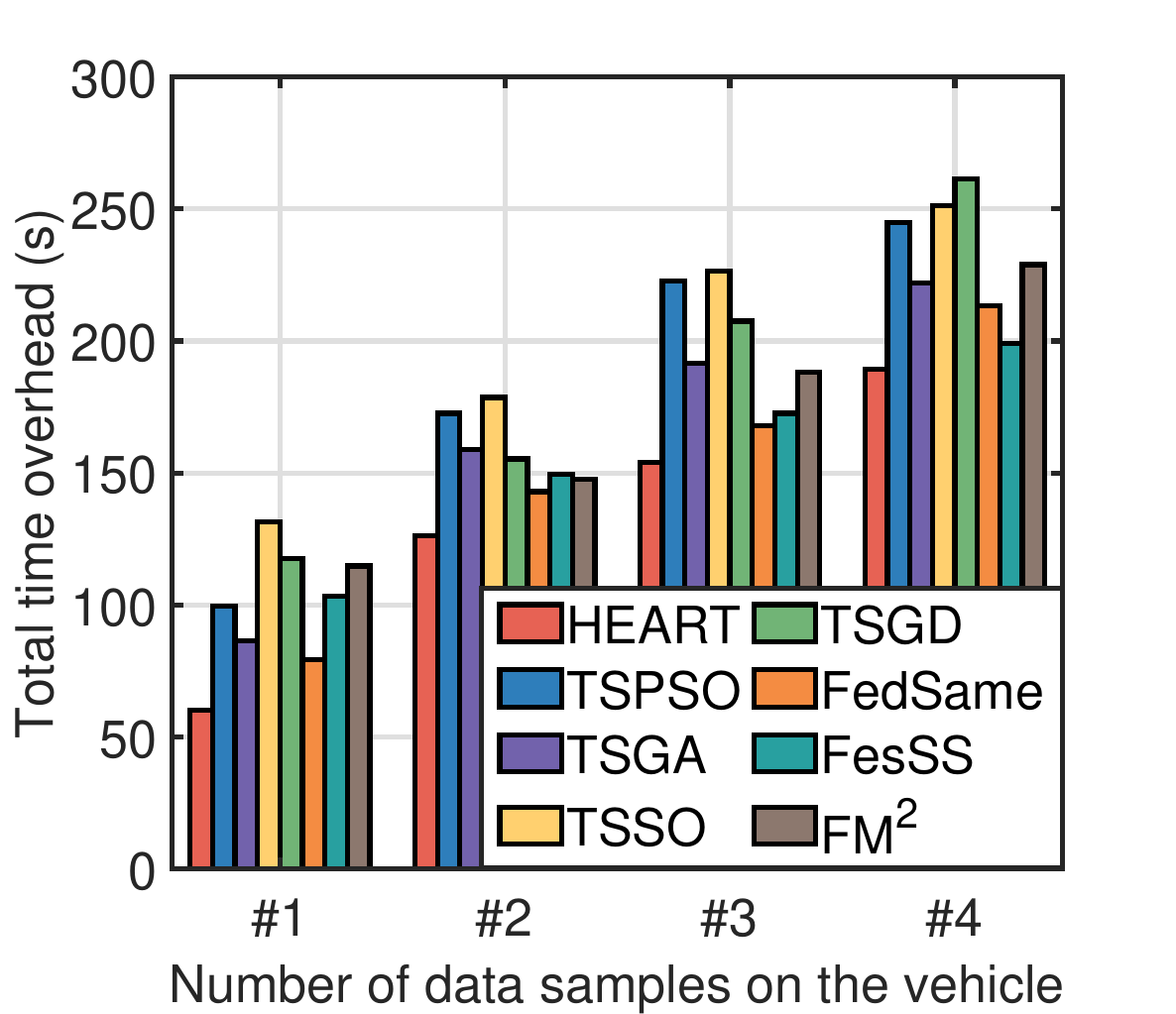}
    }
    \caption{The time that it takes for the global model of all tasks to achieve the fixed accuracy under different vehicle numbers, data samples per vehicles, and different methods: (a) 25 vehicles, (b) 50 vehicles, (c) 85 vehicles, and (d) 135 vehicles. Note that \#1-\#4 represent the number of 4 different data samples that the vehicle has for each task, which are 200, 400, 600, 800.}
    \label{fig_3}
    \vspace{-3mm}
\end{figure*}

\noindent \textbf{Evaluation of Learning Efficiency and Model Performance.} 
Fig. 4 illustrates the model accuracy performance across various methods, with target thresholds for the four tasks set at 0.75, 0.90, 0.90, and 0.75 (represented by dotted horizontal lines). In the legends, \textit{ComplT} denotes the wall-clock time at which each task (e.g., CIFAR in Fig. 4(a)) achieves its target accuracy. The total system completion time, defined as the moment the final task finishes, is highlighted in red (e.g., \textit{TSSO} completes all tasks at 215.8s upon finishing MNIST). Our proposed method completes MNIST, Driver Yawning, and CIFAR at 49.8s, 74.7s, and 99.3s, respectively, with all four tasks reaching convergence by 110.2s (upon completion of 20 Newsgroup). From Fig. 4, several key insights emerge. First, specialized FMTL methods (\textit{FedSame}, \textit{FedSS}, and \textit{FM$^2$}) demonstrate competitive per-iteration convergence due to superior task-relationship modeling, resulting in lower total time costs compared to baselines like \textit{TSPSO} and \textit{TSGD}. Second, while our method may not exhibit the fastest convergence for any single task in terms of accuracy vs. global iterations, it ensures a uniform convergence rate across all tasks. This is attributed to our strategic resource balancing, which prevents individual task bottlenecks and optimizes overall system efficiency.

As highlighted by the red values in Fig. 4, \textit{HEART} achieves the minimum system completion time (110.2s), substantially outperforming \textit{TSPSO} (191.9s), \textit{TSGA} (154.1s), \textit{TSSO} (215.8s), and \textit{TSGD} (170.7s). This efficiency stems from synchronized convergence rates across tasks, which minimizes the "long-tail" effect in total execution time. In contrast, the rapid convergence of specific tasks in \textit{TSPSO} and \textit{TSGD} (Fig. 4(a)) typically necessitates disproportionate resource prioritization, inducing a systemic imbalance that compromises overall performance. A representative trade-off is observed in \textit{TSGD}, where suboptimal scheduling for the Driver Yawning task (Fig. 3, left) leads to its failure in reaching the target accuracy (Fig. 4(c)). Ultimately, \textit{HEART} demonstrates a superior ability to bridge the gap between wall-clock convergence speed and scheduling equilibrium across all scenarios..

Furthermore, \textit{HEART} consistently achieves the minimum latency across all scenarios; for instance, with 50 vehicles, it reduces time costs by an average of 16.7\% compared to the optimal FMTL baseline. This superiority stems from \textit{HEART}'s two-stage optimization: the first stage (hybrid PSO-GA) employs global search to maximize task assignment and distribution balance, while the second stage (Greedy Sequential Optimization) maximizes training overlap to synchronize task completion and minimize idle waiting, a critical capability lacking in standard FMTL methods within dynamic IoV environments. While baselines like \textit{TSGA} and \textit{TSPSO} also optimize assignment and sequencing, their susceptibility to local optima results in performance comparable to, or even lower than, FMTL methods. This underscores a fundamental distinction: while \textit{FedSame}, \textit{FedSS}, and \textit{FM$^{2}$} prioritize model-level collaboration (e.g., parameter sharing), \textit{HEART} focuses on system-level orchestration under mobility and resource constraints. In dynamic VEC-HFL, timely completion within the residence window is the primary bottleneck; consequently, balanced resource allocation across heterogeneous tasks proves more decisive than fine-grained representation alignment.

\noindent \textbf{Evaluation of Time Efficiency and Energy Consumption.} The performance in terms of time efficiency captured via the overall time overhead (the time for all task models to reach the aforementioned desired accuracies) is shown in Fig. 5, upon testing various numbers of vehicles, where our method \textit{HEART}, outperforms other methods. For example, we can achieve an average reduction of overall time overhead by 17.2\%, 24.9\%, 24.4\%, 30.6\%, 14.2\%, 20.0\% and 15.1\% as compared to \textit{TSGA}, \textit{TSPSO}, \textit{TSSO}, \textit{TSGD}, \textit{FedSame}, \textit{FedSS} and \textit{FM$^2$} in Fig. 5(a) with 25 vehicles; and a reduction in overall time overhead by 18.3\%, 24.7\%, 34.6\%, 36.8\%, 16.7\%, 21.8\% and 18.6\% in Fig. 5(b) when considering 50 vehicles. Furthermore, when scaled to larger networks with 85 and 135 clients, \textit{HEART} continues to outperform the strongest baselines, achieving additional time reductions of 12.83\% and 12.16\%, respectively. This consistent superiority underscores the robust scalability of our proposed \textit{HEART} in handling increasingly dense and complex vehicular environments. Interestingly, Fig. 5 shows that time cost is not always affected by the increase in data volume per vehicle. This is because the increase of data volume increases local training time, but also reduces the number of global iterations required to hit the target accuracy.

\begin{figure}[!t]
    \centering
    \subfigure[]{
        \includegraphics[trim=0cm 0.2cm 0.6cm 1.2cm, clip, width=0.455\columnwidth]{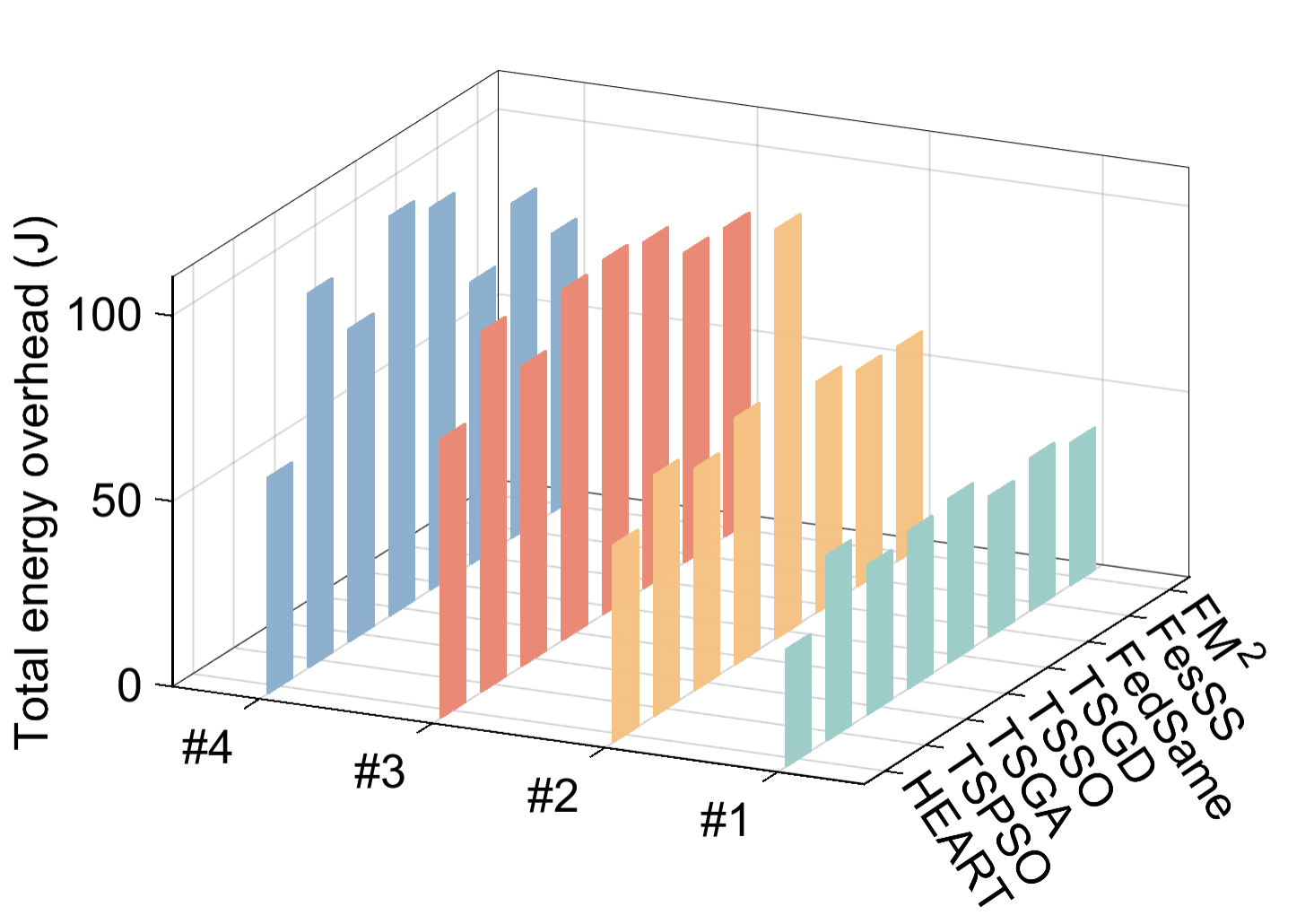}
    }
    \hspace{0.01\columnwidth} 
    \subfigure[]{
        \includegraphics[trim=0cm 0.2cm 0.61cm 1.2cm, clip, width=0.455\columnwidth]{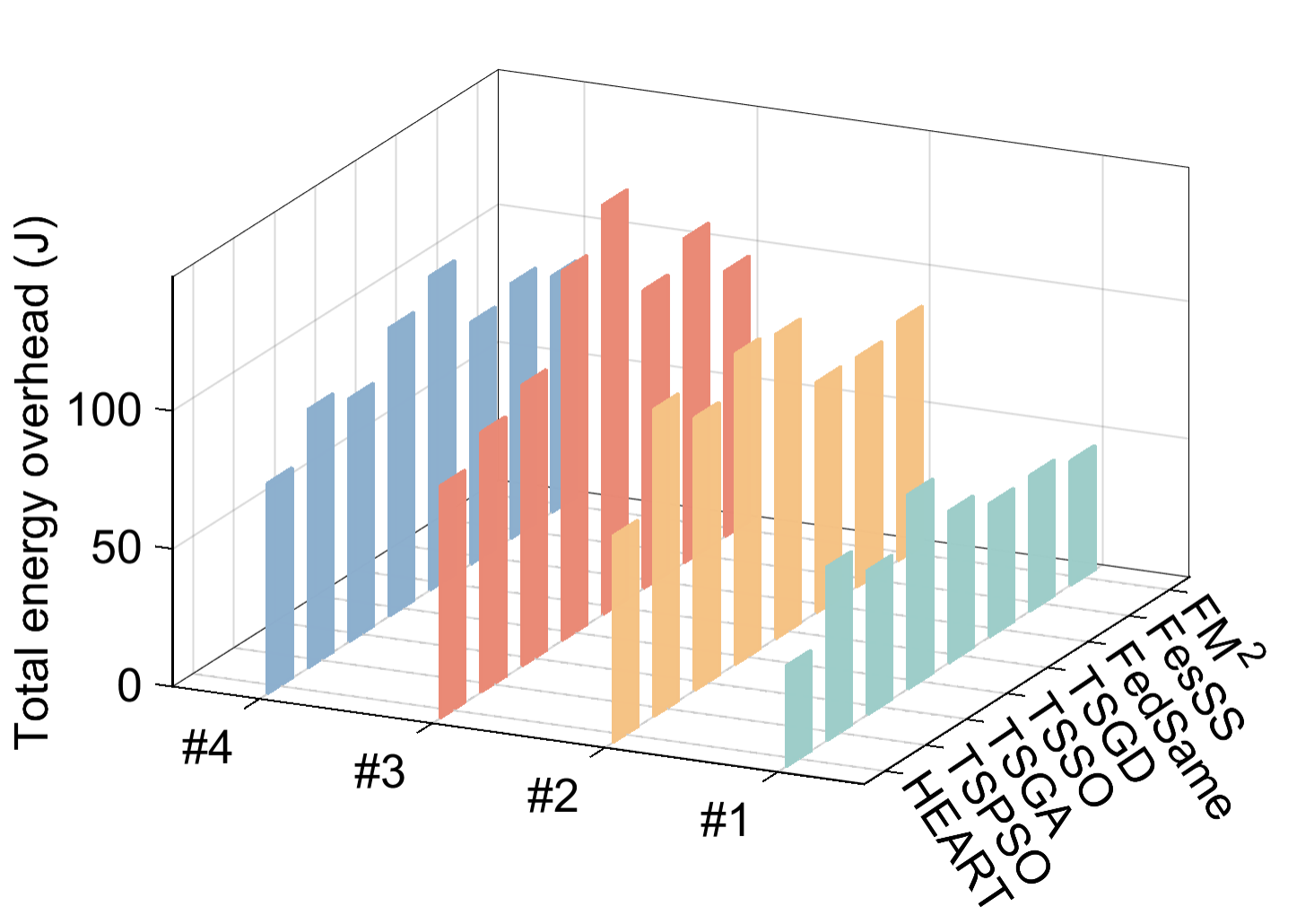}
    }
    \subfigure[]{
        \includegraphics[trim=0cm 0.2cm 0.6cm 1.2cm, clip, width=0.455\columnwidth]{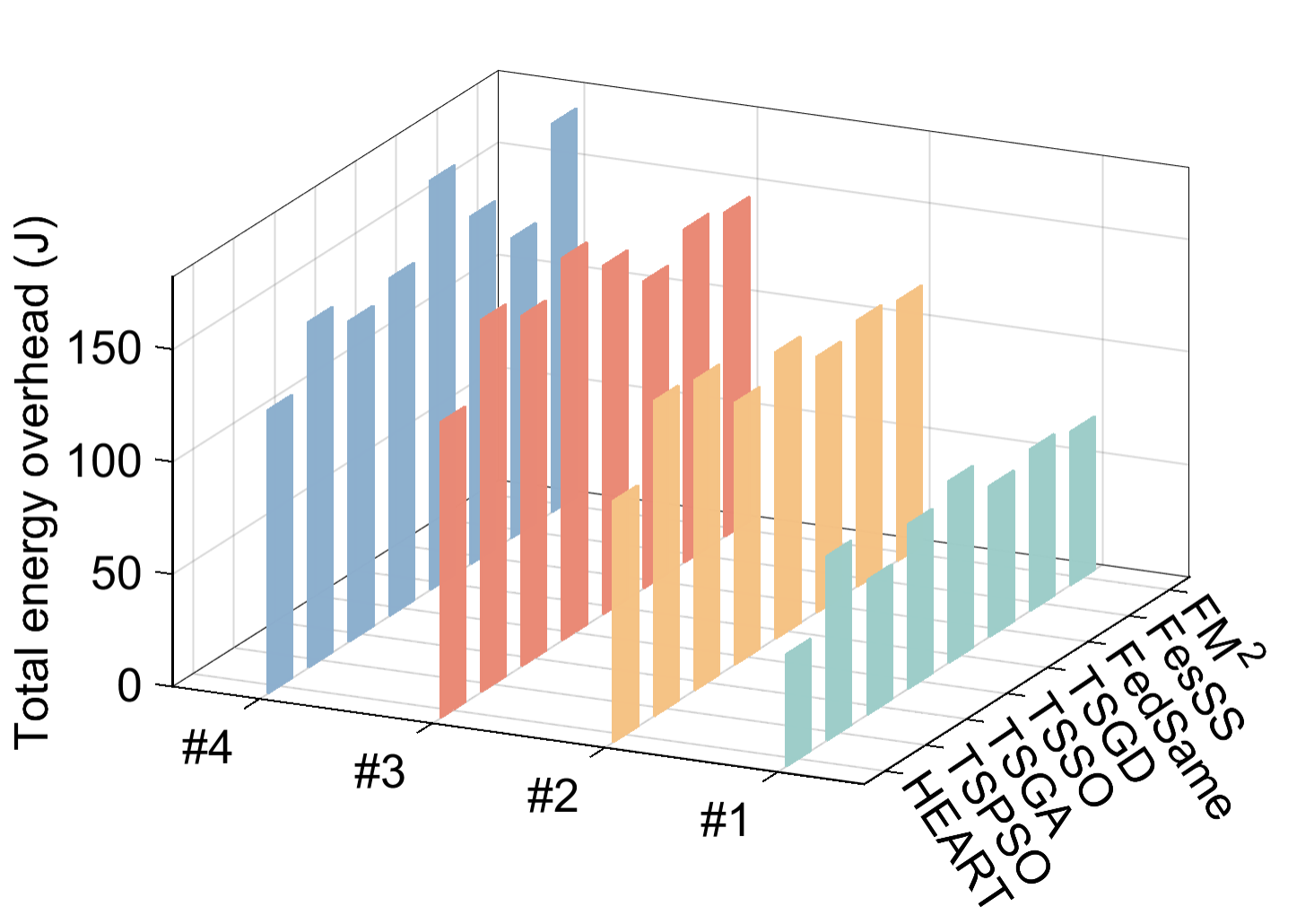}
    }
    \hspace{0.01\columnwidth} 
    \subfigure[]{
        \includegraphics[trim=0cm 0.2cm 0.61cm 1.2cm, clip, width=0.455\columnwidth]{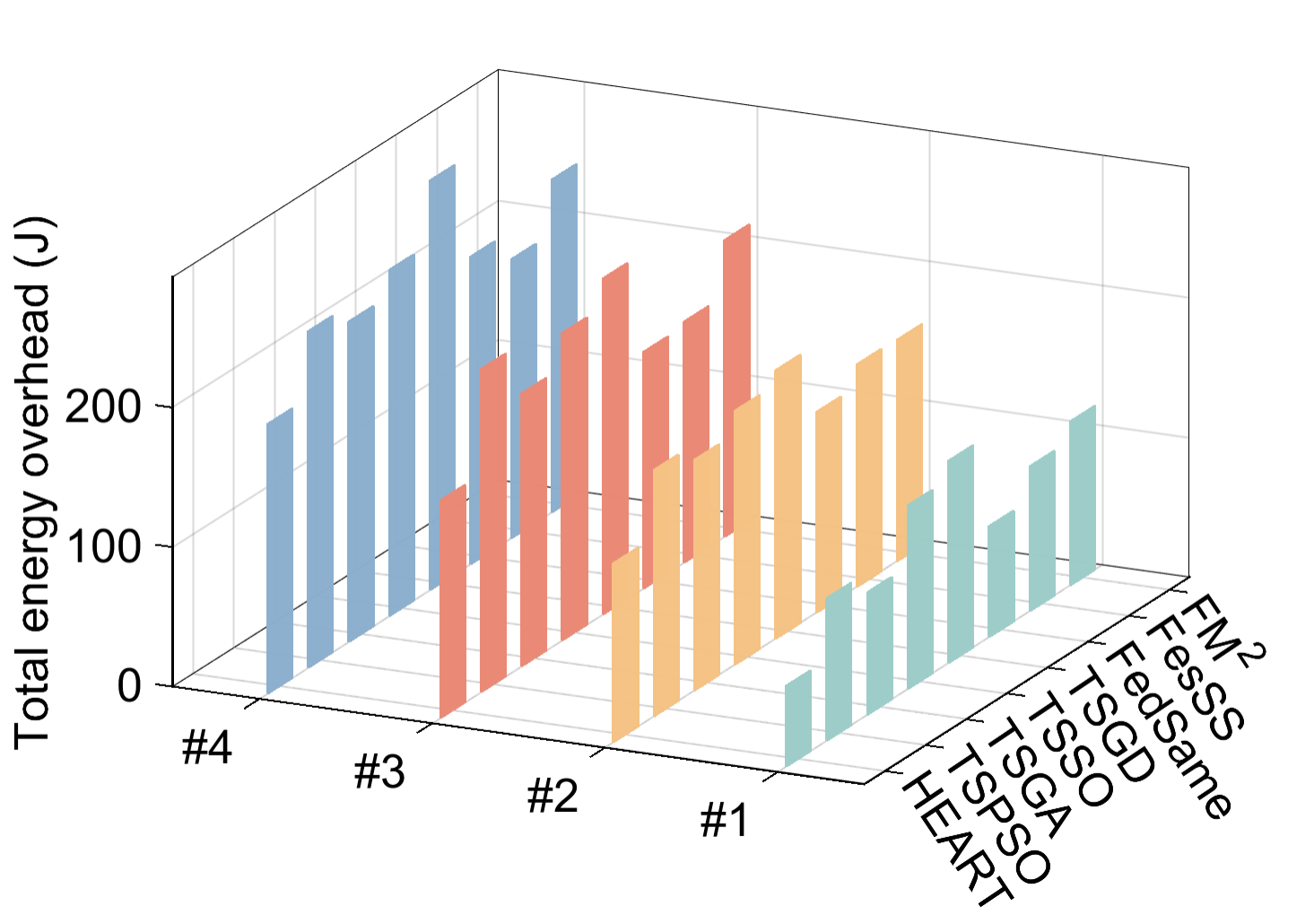}
    }
    \caption{The energy that it takes for the global model of all tasks to achieve the fixed accuracy under different vehicle numbers, data samples per vehicles, and different methods: (a) 25 vehicles, (b) 50 vehicles, (c) 85 vehicles, and (d) 135 vehicles.}
    \label{fig:energy}
    \vspace{-2mm}
\end{figure}

Although energy consumption in VEC-HFL is inherently coupled with execution time (i.e., $E = P \times T$), minimizing the overall training time, as formulated in our optimization objective, implicitly optimizes the total energy expenditure. However, the actual energy consumed during the training process is determined by the specific local training duration and the computational capacity of individual vehicles, rather than merely the maximum latency of the entire process. To provide a rigorous assessment, we adopt the energy model from to incorporate key vehicular parameters (e.g., transmission power, effective capacitance coefficients) \cite{DistributedP,energybb,energyll}. As illustrated in Fig. \ref{fig:energy}, extensive simulations across varying vehicle densities demonstrate that \textit{HEART} consistently achieves the lowest energy consumption. Specifically, compared to the strongest baseline, \textit{HEART} reduces energy overhead by 13.62\%, 12.83\%, 7.90\%, and 12.36\% in scenarios with 25, 50, 85, and 135 vehicles, respectively. These results validate that our sophisticated task scheduling and sequence optimization not only curtail training latency but also significantly minimize the overall energy footprint in dynamic VEC-HFL environments.

\begin{figure}[!t]
    \centering
    \subfigure[]{
        \includegraphics[trim=0.2cm 0cm 0.05cm 0.2cm, clip, width=0.45\columnwidth]{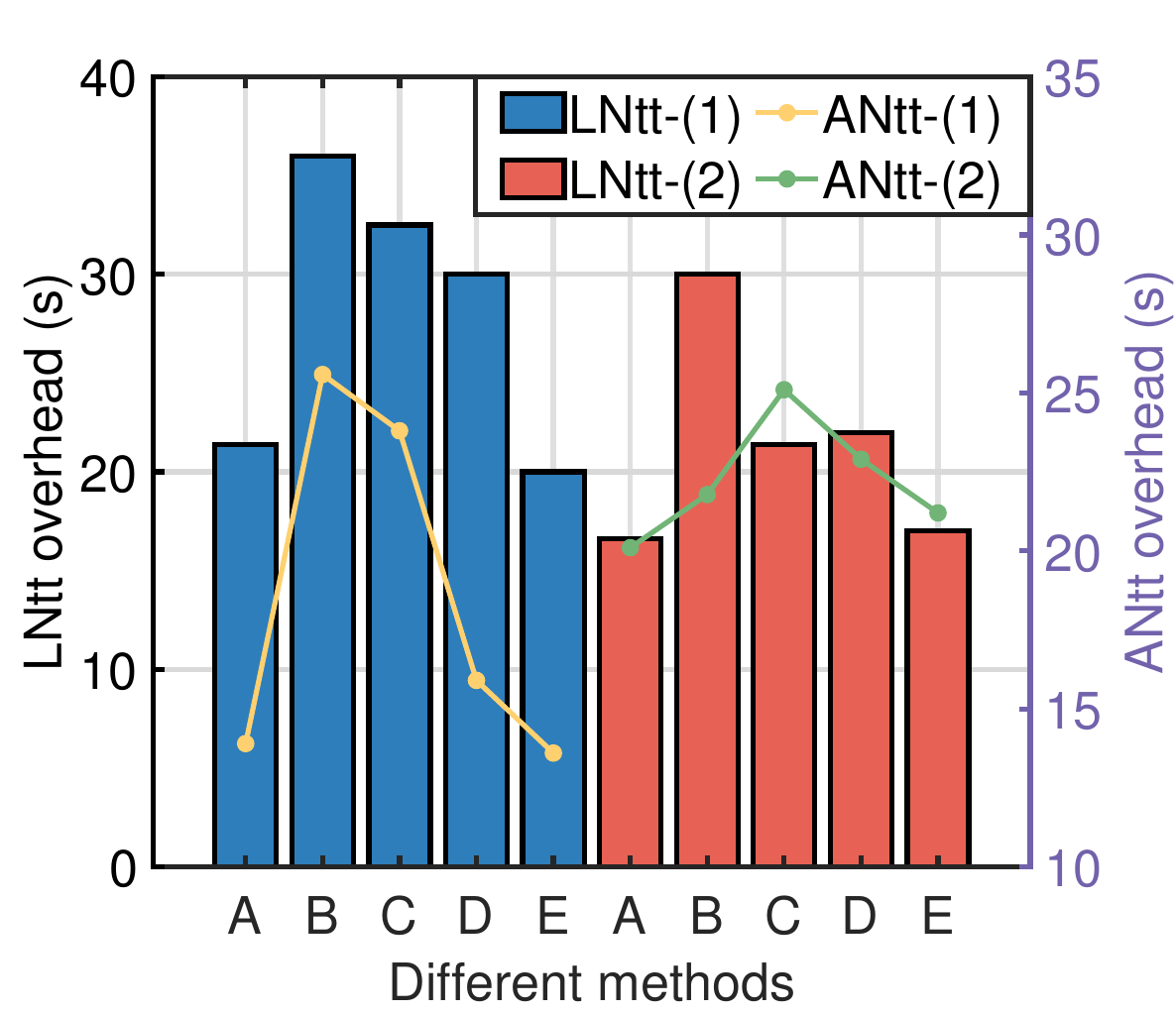}
    }
    \hspace{0.01\columnwidth} 
    \subfigure[]{
        \includegraphics[trim=0.2cm 0cm 0.2cm 0.2cm, clip, width=0.45\columnwidth]{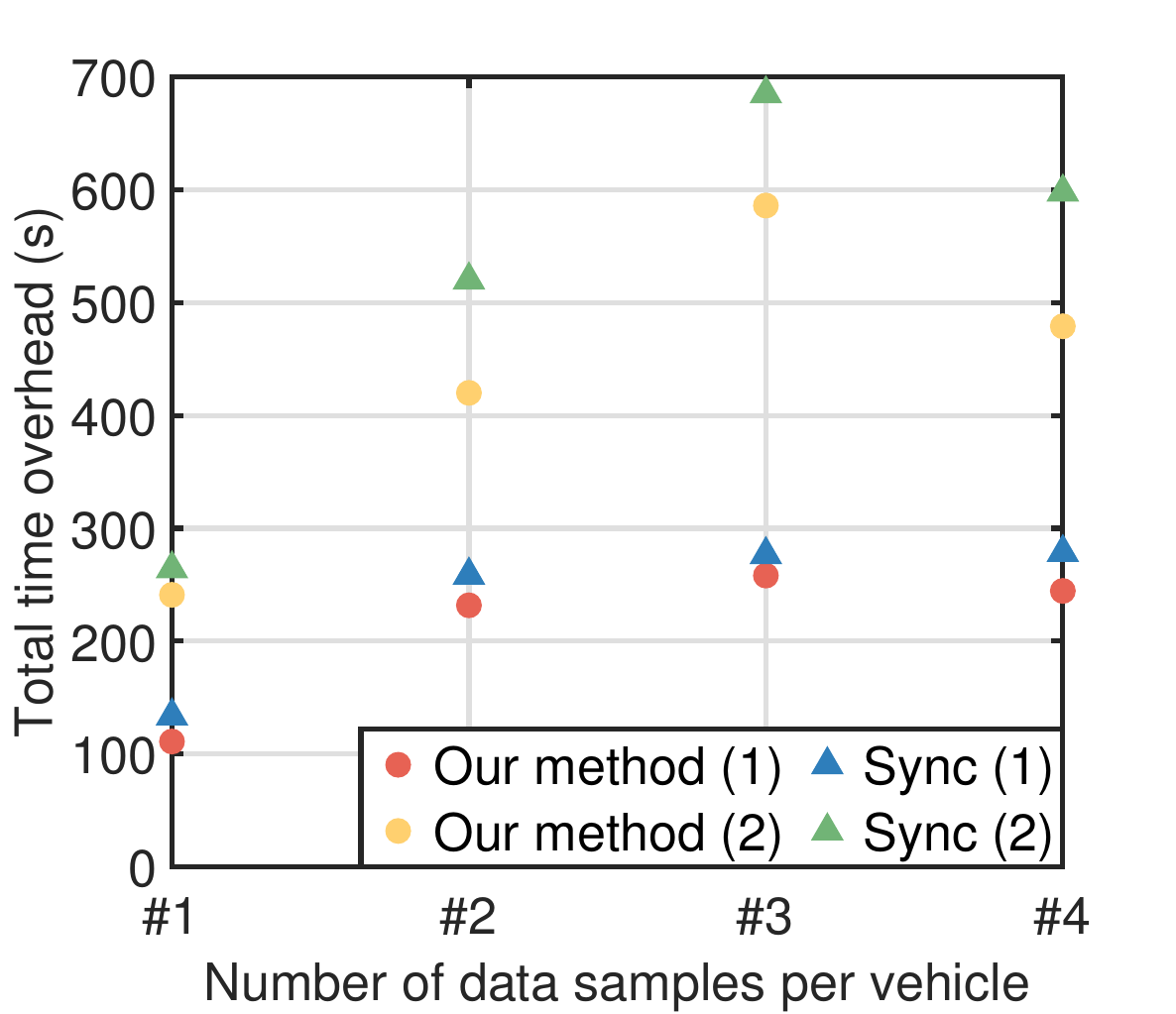}
    }
    \caption{Evaluations on non-task-training time overhead:(a) The average and the longest non-task-training time among all ESs; (b) The overall time overhead of hybrid aggregation rule and synchronous aggregation rule. Specifically, A, B, C, D and E represent \textit{HEART}, \textit{TSGA}, \textit{TSPSO}, \textit{TSSO}, \textit{TSGD}, (1) and (2) represent cases with 25 vehicles and 50 vehicles. }
    \label{fig_6}
    \vspace{-3mm}
\end{figure}

Fig. \ref{fig_6}(a) illustrates the performance in terms of the non-task-training time (abbreviated as Ntt for simplicity) across different methods, including the average (abbreviated as ANtt) and the longest non-task-training time (abbreviated as LNtt) for all ESs per global iteration. Upon considering 25 vehicles, compared to \textit{TSGA}, \textit{TSPSO}, and \textit{TSSO}, the ANtt of \textit{HEART} is lower by 51.7$\%$, 41.5$\%$, and 12.6$\%$, and its LNtt is lower by 40.1$\%$, 34.2$\%$, and 28.6$\%$. For 50 vehicles, our proposed \textit{HEART} achieves 7.7$\%$, 19.9$\%$, and 12.2$\%$ reduction of ANtt; as well as 44.6$\%$, 22.4$\%$, and 24.5$\%$ reduction of LNtt when compared with the 4 baseline methods. We note that, since \textit{HEART} and \textit{TSGD} both employ the same optimization algorithm in Stage 2, their performance outcomes are relatively similar, however, as shown in Fig. 3 and 5 our method exhibits a better time efficiency and task execution balance. These results demonstrate that our approach effectively reduces overall non-task-training time and accelerates the completion of edge iterations through efficient utilization of time for model training across the tasks. Finally, in Fig. \ref{fig_6}(b), we compare the overall execution time cost of our proposed method against the traditional synchronous aggregation rule (i.e., using synchronized aggregations for both edge and cloud/global model aggregations (i.e., Sync-HFL)). When the global model of all tasks reaches the aforementioned desired accuracies, with 25 vehicles, our method reduces the average time cost by 15.57$\%$ across various training data samples; while with 50 vehicles, it achieves a 10$\%$ reduction. These results underscore the notable time savings offered by our hybrid synchronous-asynchronous aggregation rule compared to conventional synchronous aggregation methods used in existing HFL and VEC-HFL methods \cite{Swarm, TowardRobust}.

\section{Conclusion and Future Work}
\noindent This paper took one of the first steps towards addressing the challenges of unbalanced task scheduling, model obsolescence, and low data utilization in multi-model training within the VEC-HFL architecture. We introduced a hybrid synchronous-asynchronous aggregation rule to guide the training process, combining the advantages of both synchronous and asynchronous methods. Building upon this rule, we proposed a novel methodology called the Hybrid Evolutionary And gReedy allocaTion method (HEART) to achieve balanced task scheduling and optimal task training prioritization. Extensive simulations on real-world datasets demonstrated that HEART significantly improves the efficiency of multi-model training, reducing the model training time and communication overhead compared to benchmark methods. For future work, extending HEART to incorporate additional factors is promising, including variable vehicle arrival and departures, fluctuating vehicular computing availability, data storage constraints at the vehicles, and the freshness and temporal variations of on-board datasets at vehicles. Additionally, it is highly promising to extend HEART for large-scale model training (e.g., LLM) by integrating efficiency-oriented techniques such as Federated Dropout, Split Learning, and LoRA. Furthermore, exploring multi-tier cooperative architectures, including V2V and ES-to-ES collaboration, can further facilitate localized knowledge sharing and enhance system scalability. Another critical direction involves developing context-aware resource optimization to handle extreme traffic scenarios (e.g., high-congestion urban centers or remote rural areas), by dynamically adjusting bandwidth allocation and offloading strategies. Finally, addressing further uncertainties, such as unexpected disconnections between vehicles and edge servers during model training, can be an interesting future direction.

\vspace{-0.1 cm} 
\bibliographystyle{ieeetr}
\bibliography{reference.bib}

\newpage
\clearpage
\section*{Appendix}
\subsection{Motivation of the design of hybrid PSO-GA approach}

\subsubsection{Why PSO-GA Hybridization is Particularly Suitable for Multi-Model Training}
We identify three problem-specific characteristics that make this hybridization especially appropriate: \textit{(i) Coupled discrete-continuous search space:} Task assignment variables $\overline{x}_{m,n;[g]}^{\left \langle j \right \rangle}$ are binary (discrete), while the fitness landscape involves continuous temporal dynamics (vehicle mobility, heterogeneous training times). Pure PSO excels at continuous optimization but struggles with discrete constraints; pure GA handles discrete spaces well but lacks fine-grained convergence speed. Our hybrid leverages PSO's velocity-based navigation for the continuous temporal dimension while employing GA's crossover-mutation for discrete task combinations. \textit{(ii) Multi-objective coupling with hard constraints:} The problem requires simultaneous optimization of \textit{(A)} min-max latency across tasks, \textit{(B)} balanced task distribution (constraint \eqref{17}), and \textit{(C)} dwell time feasibility (constraint \eqref{16}). Standard PSO particles tend to cluster prematurely when facing multiple competing constraints, while standalone GA suffers from slow convergence in high-dimensional spaces. The hybrid design allows PSO to rapidly identify promising regions of the Pareto front, with GA operators providing the diversity needed to maintain feasible solutions across all constraint boundaries. \textit{(iii) Time-varying solution requirements:} Vehicle mobility causes the optimal task assignment to evolve across global iterations. The adaptive inertia weight in our enhanced PSO (Eq. \eqref{24}) tracks these temporal changes, while GA's population-based diversity prevents catastrophic forgetting of good solutions when vehicle distributions shift.

\textit{Problem-Specific Advantage:}
Unlike generic scheduling problems, multi-model training involves heterogeneous task complexities (e.g., CIFAR-10 on VGG16 vs. MNIST on shallow CNN). The task weight coefficient $\rho_j$ in our fitness function (Eq. \eqref{23}) creates a non-uniform importance landscape. PSO's social learning mechanism naturally propagates high-$\rho_j$ task assignments through the swarm, while GA's crossover enables recombination of task bundles that balance computational loads across heterogeneous models.

\subsubsection{Why PSO Before GA (Not Reverse)} 
This ordering is deliberately designed based on information-theoretic properties: \textit{(i) PSO first---rapid localization of feasible regions:} PSO uses velocity vectors guided by cognitive ($\overline{x}^{p,local}$) and social ($\overline{x}^{global}$) components, creating a \textit{gradient-like} descent that efficiently identifies regions where dwell time constraints \eqref{16} are satisfied and task balance \eqref{17} is approximately achieved. Crucially, PSO's social learning propagates high-quality assignments to the entire swarm within few iterations. If GA were applied first, random crossover/mutation would generate many infeasible solutions (violating \eqref{16}), wasting computational budget before any gradient information is established. \textit{(ii) GA second---structured diversification:} After PSO identifies high-fitness regions, GA's operators serve to \textit{explore combinatorial variations of good solutions} rather than blind random search. Crossover recombines task patterns from different particles that independently discovered feasible solutions; mutation injects controlled diversity to prevent premature convergence to suboptimal distributions caused by vehicle mobility dynamics. Reversing the order would subject random GA solutions to PSO's velocity updates, effectively ``pulling'' randomly initialized particles toward potentially infeasible attractors, causing chaotic oscillations in early iterations. \textit{(iii) Constraint handling efficiency:} Our fitness function (Eq. \eqref{23}) assigns $-\infty$ to infeasible solutions. PSO's deterministic updates quickly learn to avoid these regions through velocity adaptation. GA's stochastic operators, if applied first, would repeatedly generate infeasible candidates consuming evaluation budget without informative gradient feedback. The PSO$\to$GA order ensures GA operates primarily on the feasible subset discovered by PSO. \textit{(IV) Temporal alignment:} Multi-model training involves heterogeneous task complexities ($c_{m,n}^{\left \langle j \right \rangle}$, $b_{m,n}^{\left \langle j \right \rangle}$ in Eq. (6)). PSO's velocities naturally encode ``task difficulty awareness'' through the fitness gradient, guiding toward balanced assignments. GA subsequently fine-tunes the \textit{combinatorial structure} of these balanced assignments, a discrete optimization problem that GA handles better than PSO's continuous updates.
	
\textit{Information-flow perspective:} PSO performs \textit{exploitation} (refining good solutions) while GA performs \textit{structured exploration} (recombining refined solutions). Optimization theory favors exploitation-before-exploration when: \textit{(A)} feasible regions are small relative to search space (our case: few assignments satisfy dwell time constraints), and \textit{(B)} solution quality correlates with distance in search space (similar task bundles have similar latency). The reverse order would waste exploration on low-quality regions.

\subsection{Selection and Sensitivity Analysis of the Adaptive Mutation Rate}
In this section, we provide the empirical justification for the functional form and parameter configuration of the adaptive mutation rate $\varphi(\tau)$. The selection of the logarithmic decay model is not arbitrary but is based on a rigorous performance comparison and sensitivity analysis.

\subsubsection{Theoretical Justification for Logarithmic Decay}
The adaptive mutation rate $\varphi(\tau) = \varphi_{\max} / (1 + \log(1+\tau))$ is designed based on exploration-exploitation trade-off theory: \textit{(i) Exploration-exploitation balance:} Unlike linear decay (which reduces mutation probability excessively in early-middle phases) or exponential decay (which drops too rapidly), logarithmic decay provides a smoother transition. It maintains sufficient mutation probability during middle iterations to prevent premature stagnation in complex constraint spaces, while ensuring eventual convergence. \textit{(ii) Search phase alignment:} In multi-model training resource scheduling, the solution space (encompassing bandwidth allocation, computation frequencies, vehicle selection) is highly non-linear and non-convex. Linear decay's constant reduction rate may lose the ability to escape local optima too early; exponential decay's rapid initial drop risks premature convergence to pure local search. \textit{Logarithmic decay}'s characteristic is ``fast initial descent for direction convergence, extremely slow middle descent for diversity maintenance''. This ``long-tail'' effect ensures more opportunities to correct early search deviations when handling complex VEC-HFL constraints such as dynamic topologies caused by vehicle mobility. \textit{(iii) Mathematical intuition:} The derivative of $\log(1+\tau)$ decreases over time, aligning with the ``gradual cooling while maintaining exploration resilience'' principle in simulated annealing or stochastic search.

\subsubsection{Sensitivity Analysis via Ablation Experiments}
we evaluate 50 vehicles with 4 and 9 tasks respectively, comparing logarithmic, linear, and exponential decay under varying $\varphi_{\max} \in \{\text{0.2}, \text{0.3}, \text{0.4}, \text{0.5}\}$. We use the total number of tasks assigned to vehicles (Total Tasks) and the variance between different task quantities (Variance) to evaluate different methods. For instance, Tasks 1 to 4 are allocated to 6, 9, 2, and 7 vehicles, respectively, totaling 24 training assignments with a mean of 6.0 and a variance of 6.5. A higher number of allocated vehicles per task typically enhances the positive impact on global model convergence. Meanwhile, a lower variance indicates a more uniform task distribution, which is crucial for ensuring load balancing and accelerating the overall training completion.
\begin{table}[htbp]  
	\centering
	\scriptsize
	\setlength{\tabcolsep}{4pt}
     \renewcommand{\arraystretch}{0.8}
     \caption{Sensitivity Analysis of Decay Strategies across Different $\varphi_{\max}$ and Task Scenarios}
	\begin{adjustbox}{max width=\columnwidth} 
	\label{tab:full_decay_sensitivity}
	\begin{tabular}{llcccc}
		\toprule
		\multirow{2}{*}{\textbf{$\varphi_{\max}$}} & \multirow{2}{*}{\textbf{Strategy}} & \multicolumn{2}{c}{\textbf{4 Tasks Scenario}} & \multicolumn{2}{c}{\textbf{9 Tasks Scenario}} \\
			\cmidrule(r){3-4} \cmidrule(l){5-6}
			& & Variance $\downarrow$ & Total Tasks $\uparrow$ & Variance $\downarrow$ & Total Tasks $\uparrow$ \\
			\midrule
			\multirow{3}{*}{0.2} 
			& Logarithmic & 2.50 & 120 & 14.25  & 184 \\
			& Linear      & 10.69         & 115          & 12.84          & 182 \\
			& Exponential & 6.25         & 118          & 7.56          & 180 \\
			\midrule
			\multirow{3}{*}{0.3} 
			& Logarithmic & \textbf{2.19} & \textbf{131} & \textbf{6.54}  & \textbf{190} \\
			& Linear      & 5.00          & 124          & 13.73          & 187 \\
			& Exponential & 19.19         & 119          & 6.91           & 185 \\
			\midrule
			\multirow{3}{*}{0.4} 
			& Logarithmic & 5.29 & 127 & 8.39  & 188 \\
			& Linear      & 6.19         & 125         & 9.33          & 186 \\
			& Exponential & 10.39         & 116         & 15.05           & 188 \\
			\midrule
			\multirow{3}{*}{0.5} 
			& Logarithmic & 5.51 & 124 & 10.77 & 188 \\
			& Linear      & 9.25         & 122          & 12.67         & 189 \\
			& Exponential & 12.25         & 122          & 14.69          & 184 \\
		\bottomrule
	\end{tabular}
    \end{adjustbox}
\end{table}

According to the ablation experiment results in Table \ref{tab:full_decay_sensitivity}, at $\varphi_{\max}$ = 0.3, logarithmic decay achieves the optimal trade-off between balanced task distribution and resource utilization, attaining the lowest variance (2.19 for 4 tasks scenario, 6.54 for 9 tasks scenario) that is critical for satisfying constraint \eqref{17}, while simultaneously achieving the highest total task allocation (131 for 4 tasks scenario, 190 for 9 tasks scenario) to maximize resource utilization under dwell time constraints \eqref{16}. This advantage is most pronounced at $\varphi_{\max}$ = 0.3, as higher values (0.4, 0.5) yield diminishing returns due to excessive mutation rates disrupting refined solutions, whereas lower values (0.2) insufficiently explore the feasible space.
	
These results validate that logarithmic decay with $\varphi_{\max}$=0.3 optimally balances exploration (to find feasible assignments) and exploitation (to refine balanced distributions), directly supporting our design choice for the dynamic VEC-HFL environment.

\subsection{Detailed Implementation of the Hybrid PSO-GA Algorithm}
In the next, we provide a comprehensive breakdown of the proposed hybrid heuristic approach used for task scheduling optimization.

\noindent \textbf{Step 1.} \textbf{Initialization: }Our method starts with randomly initializing \textit{(i)} the task scheduling $\overline{x}_{m,n;[g]}^{p,\left \langle j \right \rangle}(0)$ for each particle $p$ while satisfying constraint \eqref{16} and the velocity $v_{m,n;[g]}^{p,\left \langle j \right \rangle}(0)$, \textit{(ii)} the local optimal task scheduling $\overline{x}_{m,n;[g]}^{p,local,\left \langle j \right \rangle}$ \textit{(iii)} the local optimal fitness $f_{m,n;[g]}^{p,local}$ of each particle $p$, \textit{(iv)} the global optimal task scheduling $\overline{x}_{m,n;[g]}^{global,\left \langle j \right \rangle}$ (i.e., the task scheduling of the vehicle), and \textit{(v)} the global optimal fitness $f_{m,n;[g]}^{global}$ of all particles.

\noindent \textbf{Step 2.} \textbf{Implementing PSO Process: }In each iteration, the adaptive inertia weight $\pi(\tau)$ is obtained according to \eqref{24}. For each particle $p$, the velocity $v_{m,n;[g]}^{p,\left \langle j \right \rangle}(\tau)$ is first calculated based on \eqref{25}. Then, the probability of each task being assigned $\Phi^{p,\left \langle j \right \rangle}_{m,n;[g]}(\tau)$ is calculated using \eqref{26}, and the task scheduling $\overline{x}_{m,n;[g]}^{p,\left \langle j \right \rangle}(\tau)$ is obtained via \eqref{27} (lines 7-8, Alg. 2). Next, the fitness value $f_{m,n;[g]}^p(\tau)$ of the particle is computed based on \eqref{23} and compared with $f_{m,n;[g]}^{p,local}$ of the particle and $f_{m,n;[g]}^{global}$ of all particles, according to which the following conditions may occur (lines 10-12, Alg. 2):

\textbf{Condition 1.} If $f_{m,n;[g]}^p(\tau)$ is greater than $f_{m,n;[g]}^{p,local}$ and $f_{m,n;[g]}^{global}$, it implies the high quality of the task scheduling associated with particle $p$. Subsequently, we perform the following updates: $\overline{x}_{m,n;[g]}^{local,\left \langle j \right \rangle} \xleftarrow[]{} \overline{x}_{m,n;[g]}^{p,\left \langle j \right \rangle}(\tau)$ and $f_{m,n;[g]}^{p,local} \xleftarrow[]{} f_{m,n;[g]}^p(\tau)$; $\overline{x}_{m,n;[g]}^{global,\left \langle j \right \rangle} \xleftarrow[]{} \overline{x}_{m,n;[g]}^{p,\left \langle j \right \rangle}(\tau)$ and $f_{m,n;[g]}^{global} \xleftarrow[]{} f_{m,n;[g]}^p(\tau)$.

\textbf{Condition 2.} If $f_{m,n;[g]}^p(\tau)$ is greater than $f_{m,n;[g]}^{p,local}$ but lower than $f_{m,n;[g]}^{global}$, it implies the local optimality of the particle. Subsequently, we perform perform $\overline{x}_{m,n;[g]}^{local,\left \langle j \right \rangle} \xleftarrow[]{} \overline{x}_{m,n;[g]}^{p,\left \langle j \right \rangle}(\tau)$.

\textbf{Condition 3.} If $f_{m,n;[g]}^p(\tau)$ is lower than $f_{m,n;[g]}^{p,local}$ and $f_{m,n;[g]}^{global}$, it implies the low quality of the particle. Subsequently, no update will be triggered.

\noindent \textbf{Step 3.} \textbf{Implementing GA Process: }GA aims to increase the exploration space of the solution via conducting crossing and mutation on the task scheduling variable $\overline{x}_{m,n;[g]}^{p,\left \langle j \right \rangle}(\tau)$ obtained from the PSO process. It selects two particles $p$ and $p+1$ in sequence for these operations until all particles are covered (line 15, Alg. 2) to obtain the task scheduling vectors $X^p_{m,n;[g]}(\tau)$ and $X^{p+1}_{m,n;[g]}(\tau)$ for particles $p$ and $p+1$ during the crossover process using the crossover point $r$ for element swapping in the vectors (based on \eqref{28} and \eqref{29}). It then obtains the updated task scheduling variables $\overline{x}_{m,n;[g]}^{{p,\left \langle j \right \rangle,c_1}}(\tau)$ and $\overline{x}_{m,n;[g]}^{{p+1,\left \langle j \right \rangle,c_2}}(\tau)$. It then executes the mutation operation which mainly flips the task scheduling decisions when randomly selected $e_2$ is less than the adaptive mutation rate $\varphi(\tau)$ (based on \eqref{30}). Through this operation, it obtains new task scheduling solutions $\overline{x}_{m,n;[g]}^{{p,\left \langle j \right \rangle}}(\tau)$ and $\overline{x}_{m,n;[g]}^{{p+1,\left \langle j \right \rangle}}(\tau)$. In addition, if the new task scheduling solutions do not meet constraint \eqref{16}, \eqref{117} and \eqref{17}, the task with the longest training time is removed until the constraint is met.

\noindent \textbf{Step 4.} \textbf{Repeat: }The above steps (i.e., Steps 1 to 3) are repeated for each vehicle to obtain the global optimal solution.

\subsection{Detailed Implementation of the Greedy-based Task Ranking Algorithm}
This section elaborates on the greedy-based heuristic employed to optimize the training sequence/rank of tasks.

\noindent \textbf{Step 1.} \textbf{Initialization: }The highest aggregate-score $\mathcal{S}^*_{m;[g,k]}$ is initialized to zero, making the training sequence $\lambda_{m,n;[g,k]}$ of each vehicle under ES $m$ coverage an empty set, and the set $\mathcal{J}'$ is initialized as $\mathcal{J}^{\prime}=\mathcal{J}$. Also, variable $q$ to represent the task with the highest $\mathcal{S}^*_{m;[g,k]}$, used in the following (line 1, Alg. 3).

\noindent \textbf{Step 2.} \textbf{Select proper task for $\mathcal{S}^*_{m;[g,k]}$: } For the unassigned task set $\mathcal{J}^\prime$ of ES $m$, our method checks through all the tasks and computes the overlap score $\mathcal{S}_{m;[g,k]}^{lap,\langle j\rangle}$ based on \eqref{31} (line 5, Alg. 3) for each task $j$. It then goes over every vehicle under the coverage of ES that has this training task and obtains the upload time $t_{m,n;[g,k]}^{V2E,\left \langle j\right \rangle}$ based on (8). Afterwards, it calculates the model upload score $\mathcal{S}^{up,\left \langle j\right \rangle}_{m,n;[g,k]}$ based on the $t_{m,n;[g,k]}^{V2E,\left \langle j\right \rangle}$ (using \eqref{32}) (lines 7 and 8, Alg. 3). Finally, it obtains $\mathcal{S}_{m;[g,k]}^{all,\left\langle j\right\rangle}$ for this task (based on \eqref{33}) (line 9, Alg. 3). Next, it compares this value with $\mathcal{S}^*_{m;[g,k]}$ (lines 10-12, Alg. 3), until task $q$ with the highest $\mathcal{S}^*_{m;[g,k]}$ is selected. The vehicle that owns this task then adds it to the end of $\lambda_{m,n;[g,k]}$ and $\mathcal{S}^*_{m;[g,k]}$ is then reset to zero (lines 14-16, Alg. 3).

\noindent \textbf{Step 3.} \textbf{Repeat: }Task $q$ is then removed from $\mathcal{J}^\prime$ and the above steps are repeated for the remaining tasks until all the tasks are placed in the training sequence of vehicles.

\subsection{Convergence Analysis of HEART}\label{sec:convergence}

In this section, we establish theoretical convergence guarantees for the proposed HEART with hybrid synchronous-asynchronous aggregation. We first formalize the problem setting and key assumptions, then derive the convergence bounds that characterize the impact of edge-level synchronous aggregation with deadline mechanisms and cloud-level asynchronous aggregation on training stability and convergence rate.

\subsubsection{Problem Formulation and Notations}

For each task $j \in \mathcal{J}$, we define the global empirical risk minimization problem as:
\begin{align}\label{eq:global_opt}
	\min_{\omega^{\langle j \rangle} \in \mathbb{R}^d} L^{\langle j \rangle}(\omega^{\langle j \rangle}) = \sum_{m \in \mathcal{M}^{\langle j \rangle}} \sum_{n \in \mathcal{N}_m^{\langle j \rangle}} \frac{|\mathcal{D}_{m,n}^{\langle j \rangle}|}{|\mathcal{D}^{\langle j \rangle}|} L_{m,n}^{\langle j \rangle}(\omega^{\langle j \rangle}),
\end{align}
where $|\mathcal{D}^{\langle j \rangle}| = \sum_{m \in \mathcal{M}^{\langle j \rangle}} \sum_{n \in \mathcal{N}_m^{\langle j \rangle}} |\mathcal{D}_{m,n}^{\langle j \rangle}|$ denotes the total data volume across all vehicles for task $j$.

To characterize the asynchrony in our hybrid aggregation architecture, we introduce the following key definitions.

\begin{definition}[Version Age at Cloud]\label{def:version_age}
	For ES $m \in \mathcal{M}^{\langle j \rangle}_{[g]}$ at global iteration $g$, the \textit{version age} $\varrho_{m;[g]}^{\langle j \rangle}$ is defined as the number of global iterations since ES $m$ last participated in cloud aggregation:
	\begin{align}
		\varrho_{[m;[g]}^{\langle j \rangle} = g - g_m^{Last,\langle j \rangle},
	\end{align}
	where $g_m^{Last,\langle j \rangle} = \max\{g' < g: m \in \mathcal{M}_{[g']}^{\langle j \rangle}\}$. The \textit{maximum version age} is defined as $\varrho_{\max}^{\langle j \rangle} = \max \varrho_{m;[g]}^{\langle j \rangle} \leq M - Q^{\langle j \rangle} + 1$.
\end{definition}

\begin{definition}[Effective Participation Rate]\label{def:participation_rate}
	The \textit{edge effective participation rate} for task $j$ at ES $m$ during $k^\text{th}$ edge iteration of global iteration $g$ is:
	\begin{align}
		\gamma_{m;[g,k]}^{\langle j \rangle} = \frac{|\widetilde{\mathcal{D}}_{m;[g,k]}^{Act,\langle j \rangle}|}{|\widetilde{\mathcal{D}}_{m;[g,k]}^{\langle j \rangle}|} = \frac{\sum_{n \in \mathcal{N}_{m;[g,k]}^{Act,\langle j \rangle}} |\mathcal{D}_{m,n}^{\langle j \rangle}|}{\sum_{n \in \mathcal{N}_{m;[g,k]}^{\langle j \rangle}} |\mathcal{D}_{m,n}^{\langle j \rangle}|}.
	\end{align}
	The \textit{global effective participation rate} at cloud iteration $g$ is the weighted average:
	\begin{align}
		\bar{\gamma}_{[g]}^{\langle j \rangle} = \sum_{m \in \mathcal{M}_{[g]}^{\langle j \rangle}} \frac{|\widetilde{\mathcal{D}}_{m;[g,K^{\langle j \rangle}]}^{\langle j \rangle}|}{|\overline{\mathcal{D}}_{[g]}^{\langle j \rangle}|} \cdot \gamma_{m;[g,K^{\langle j \rangle}]}^{\langle j \rangle}.
	\end{align}
\end{definition}

\subsubsection{Assumptions}

We make the following assumptions for  convergence analysis \cite{DeFedGCN,Asynchronous12,Asynchronous13}.

\begin{assumption}[Smoothness and Lower Boundedness]\label{ass:smoothness}
	For each task $j \in \mathcal{J}$, the global loss function $L^{\langle j \rangle}(\cdot)$ is $L$-smooth, i.e., for all $\omega^{\langle j \rangle}, \omega^{\langle j \rangle,'} \in \mathbb{R}^d$:
	\begin{align}\label{eq:L_smooth}
		\|\nabla L^{\langle j \rangle}(\omega^{\langle j \rangle}) - \nabla L^{\langle j \rangle}(\omega^{\langle j \rangle,'})\| \leq L \|\omega^{\langle j \rangle} - \omega^{\langle j \rangle,'}\|,
	\end{align}
	and is lower bounded by $L_*^{\langle j \rangle} > -\infty$.
\end{assumption}

\begin{assumption}[Stochastic Gradient Properties]\label{ass:stochastic}
	For any vehicle $n$ under ES $m$, the stochastic gradient $\nabla L_{m,n}^{\langle j \rangle}(\omega^{\langle j \rangle}; \mathcal{B})$ computed on mini-batch $\mathcal{B}$ satisfies:
	\begin{enumerate}
		\item \textbf{Unbiasedness}: $\mathbb{E}_{\mathcal{B}}[\nabla L_{m,n}^{\langle j \rangle}(\omega^{\langle j \rangle}; \mathcal{B})] = \nabla L_{m,n}^{\langle j \rangle}(\omega^{\langle j \rangle})$;
		\item \textbf{Bounded Variance}: Given a constant $\sigma^2 > 0$ as the upper bound of the local stochastic gradient variance, we have $\mathbb{E}_{\mathcal{B}}[\|\nabla L_{m,n}^{\langle j \rangle}(\omega^{\langle j \rangle}; \mathcal{B}) - \nabla L_{m,n}^{\langle j \rangle}(\omega^{\langle j \rangle})\|^2] \leq \sigma^2$;
		\item \textbf{Bounded Gradient}: $\|\nabla L_{m,n}^{\langle j \rangle}(\omega^{\langle j \rangle})\| \leq \mathrm{c}_1$ for some constant $\mathrm{c}_1 > 0$.
	\end{enumerate}
\end{assumption}

\begin{assumption}[Bounded Delay and Decaying Learning Rate]\label{ass:delay}
	The maximum version age is bounded by a constant $\varrho_0$, i.e., $\varrho_{\max}^{\langle j \rangle} \leq \varrho_0$ for all $j$. The learning rate satisfies $\eta_{[g]}^{\langle j \rangle} = \frac{\eta_0}{\sqrt{g+1}}$ with $\eta_0 \leq \min\left\{\frac{1}{4L\varrho_0}, \frac{1}{L}\right\}$, where $L$ denotes the Lipschitz constant of the gradient $\nabla L^{\langle j \rangle}(\cdot)$.
\end{assumption}

\begin{assumption}[Minimum Participation Rate]\label{ass:participation}
	There exists $\gamma_{\min} > 0$ such that $\gamma_{m;[g,k]}^{\langle j \rangle} \geq \gamma_{\min}$ almost surely for all $m, g, k, j$.
\end{assumption}

\begin{assumption}[Momentum Coefficient Constraint]\label{ass:momentum}
	The global aggregation weight satisfies $\alpha^{\langle j \rangle} \in [0, 1)$ and $\alpha^{\langle j \rangle} \leq 1 - \frac{\mathrm{c}_2}{\varrho_{\max}^{\langle j \rangle}}$ for some constant $\mathrm{c}_2 > 0$.
\end{assumption}

\subsubsection{Key Lemmas} We derive key lemmas that bound the errors introduced by the hybrid aggregation mechanism.

\begin{lemma}[Edge Aggregation Bias]\label{lem:edge_bias}
	Under Assumptions~\ref{ass:smoothness}--\ref{ass:participation}, the deviation between the actual edge model $\widetilde{\omega}_{m;[g,k]}^{\langle j \rangle}$ and the ideal edge model $\widetilde{\omega}_{m;[g,k]}^{Ideal,\langle j \rangle}$ (aggregated from all assigned vehicles without dropout) satisfies:
	\begin{align}\label{eq:edge_bias}
		\mathbb{E}\left[\left\|\widetilde{\omega}_{m;[g,k]}^{\langle j \rangle} - \widetilde{\omega}_{m;[g,k]}^{Ideal,\langle j \rangle}\right\|^2\right] &\leq \frac{4(1-\gamma_{m;[g,k]}^{\langle j \rangle})^2}{(\gamma_{m;[g,k]}^{\langle j \rangle})^2} \cdot \frac{\mathrm{c}_1^2}{L^2} \notag \\
        &+ \mathcal{O}\left((\eta_{[g,k]}^{\langle j \rangle})^2 H^{\langle j \rangle}\right).
	\end{align}
\end{lemma}

\begin{proof}
	Let $\mathcal{N}_{m;[g,k]}^{Drop,\langle j \rangle} = \mathcal{N}_{m;[g,k]}^{\langle j \rangle} \setminus \mathcal{N}_{m;[g,k]}^{Act,\langle j \rangle}$ denote the set of dropped vehicles. The ideal edge aggregation is:
	\begin{align}
		\widetilde{\omega}_{m;[g,k]}^{Ideal,\langle j \rangle} = \sum_{n \in \mathcal{N}_{m;[g,k]}^{\langle j \rangle}} \frac{|\mathcal{D}_{m,n}^{\langle j \rangle}|}{|\widetilde{\mathcal{D}}_{m;[g,k]}^{\langle j \rangle}|} \omega_{m,n;[g,k,H^{\langle j \rangle}]}^{\langle j \rangle}.
	\end{align}
	The actual aggregation can be rewritten as:
	\begin{align}
		\widetilde{\omega}_{m;[g,k]}^{\langle j \rangle} &= \widetilde{\omega}_{m;[g,k]}^{Ideal,\langle j \rangle} - \notag\\
        &\sum_{n \in \mathcal{N}_{m;[g,k]}^{Drop,\langle j \rangle}} \frac{|\mathcal{D}_{m,n}^{\langle j \rangle}|}{|\widetilde{\mathcal{D}}_{m;[g,k]}^{Act,\langle j \rangle}|} \left(\omega_{m,n;[g,k,H^{\langle j \rangle}]}^{\langle j \rangle} - \widetilde{\omega}_{m;[g,k]}^{Ideal,\langle j \rangle}\right).
	\end{align}
	Taking expectation and using $\|\omega_{m,n;[g,k,H^{\langle j \rangle}]}^{\langle j \rangle} - \widetilde{\omega}_{m;[g,k]}^{Ideal,\langle j \rangle}\| \leq \frac{2\mathrm{c}_1}{L}$ (from $L$-smoothness and gradient bound), we obtain:
	\begin{align}
		\mathbb{E}\left[\left\|\widetilde{\omega}_{m;[g,k]}^{\langle j \rangle} - \widetilde{\omega}_{m;[g,k]}^{Ideal,\langle j \rangle}\right\|^2\right] \leq \left(\frac{1-\gamma_{m;[g,k]}^{\langle j \rangle}}{\gamma_{m;[g,k]}^{\langle j \rangle}}\right)^2 \cdot \frac{4\mathrm{c}_1^2}{L^2} + \mathcal{R}, \label{eq:finall}
	\end{align}
	where $\mathcal{R}=\mathcal{O}\left((\eta_{[g,k]}^{\langle j \rangle})^2 H^{\langle j \rangle}\right)$ is defined as a high-order error term, which is derived from the SGD discretization error transformation when considering local iterations. Then, \eqref{eq:finall} yields \eqref{eq:edge_bias} after simplification.
\end{proof}

\begin{lemma}[Cloud Asynchrony Error]\label{lem:cloud_async}
    Let $\varpi_{[g]}^{\langle j \rangle}$ be the actual global model and $\hat{\varpi}_{[g]}^{\langle j \rangle}$ be the \textit{virtual synchronous global model} (the trajectory obtained if all ESs participated in every global aggregation without delay). Under Assumptions~\ref{ass:smoothness}--\ref{ass:momentum}, the deviation due to cloud-level asynchrony and edge-level dropout satisfies:
    \begin{equation}\label{eq:cloud_async_refined}
    \hspace{-2mm}
    \resizebox{0.5\textwidth}{!}{$
    \begin{aligned}
        \mathbb{E} \left[ \|\varpi_{[g]}^{\langle j \rangle} - \hat{\varpi}_{[g]}^{\langle j \rangle}\|^2 \right] \leq & \frac{2\varrho_{\max}^{\langle j \rangle}}{(1-\alpha^{\langle j \rangle})^2} \sum_{s=g-\varrho_{\max}^{\langle j \rangle}}^{g-1} \mathbb{E} \left[ \|\hat{\varpi}_{[s+1]}^{\langle j \rangle} - \hat{\varpi}_{[s]}^{\langle j \rangle}\|^2 \right] \\
        & + \frac{8\mathrm{c}_1^2(1-\bar{\gamma}_{[g]}^{\langle j \rangle})^2}{L^2(\bar{\gamma}_{[g]}^{\langle j \rangle})^2} + \mathcal{O}\left( (\eta_{[g]}^{\langle j \rangle})^2 H^{\langle j \rangle} \right).
    \end{aligned}
    $}\hspace{-2mm}
    \end{equation}
\end{lemma}

\begin{proof}
    Let $\mathbf{e}_{[g]}^{\langle j \rangle} = \varpi_{[g]}^{\langle j \rangle} - \hat{\varpi}_{[g]}^{\langle j \rangle}$ denote the error vector. From the global aggregation rule, the actual model evolution is given by:
    \begin{align}
        \varpi_{[g]}^{\langle j \rangle} &= \alpha^{\langle j \rangle} \varpi_{[g-1]}^{\langle j \rangle} + \notag \\(1-\alpha^{\langle j \rangle}) \sum_{m \in \mathcal{M}_{[g]}^{\langle j \rangle}} &w_{m;[g]}^{\langle j \rangle} \left( \hat{\varpi}_{[g-\varrho_{m;[g]}^{\langle j \rangle}]}^{\langle j \rangle} + \boldsymbol{\delta}_{m;[g]}^{\langle j \rangle} \right),
    \end{align}
    where $w_{m;[g]}^{\langle j \rangle} = \frac{|\widetilde{\mathcal{D}}_{m;[g,K^{\langle j \rangle}]}^{\langle j \rangle}|}{|\overline{\mathcal{D}}_{[g]}^{\langle j \rangle}|}$ and $\boldsymbol{\delta}_{m;[g]}^{\langle j \rangle} = \widetilde{\omega}_{m;[g,K^{\langle j \rangle}]}^{\langle j \rangle} - \hat{\varpi}_{[g-\varrho_{m;[g]}^{\langle j \rangle}]}^{\langle j \rangle}$ is the edge aggregation bias.
    
    The virtual synchronous model follows $\hat{\varpi}_{[g]}^{\langle j \rangle} = \hat{\varpi}_{[g-1]}^{\langle j \rangle} + \text{Ideal Update}$. Subtracting $\hat{\varpi}_{[g]}^{\langle j \rangle}$ from both sides of the evolution and using the convex property $\sum w_{m;[g]}^{\langle j \rangle} = 1$:
    \begin{align}
        \mathbf{e}_{[g]}^{\langle j \rangle} &= \alpha^{\langle j \rangle} \mathbf{e}_{[g-1]}^{\langle j \rangle} + \notag \\(1-\alpha^{\langle j \rangle}) \sum_{m \in \mathcal{M}_{[g]}^{\langle j \rangle}} w_{m;[g]}^{\langle j \rangle} &\left( \hat{\varpi}_{[g-\varrho_{m;[g]}^{\langle j \rangle}]}^{\langle j \rangle} - \hat{\varpi}_{[g-1]}^{\langle j \rangle} + \boldsymbol{\delta}_{m;[g]}^{\langle j \rangle} \right).
    \end{align}
    
    Applying Jensen's Inequality for the convex combination and the inequality $\|a+b\|^2 \leq 2\|a\|^2 + 2\|b\|^2$:
    \begin{equation}
    \hspace{-2mm}
    \resizebox{0.5\textwidth}{!}{$
    \begin{aligned}
        \|\mathbf{e}_{[g]}^{\langle j \rangle}\|^2 &\leq \alpha^{\langle j \rangle} \|\mathbf{e}_{[g-1]}^{\langle j \rangle}\|^2 + \\
        &2(1-\alpha^{\langle j \rangle}) \sum_{m \in \mathcal{M}_{[g]}^{\langle j \rangle}} w_{m;[g]}^{\langle j \rangle}\left( \|\hat{\varpi}_{[g-\varrho_{m;[g]}^{\langle j \rangle}]}^{\langle j \rangle} - \hat{\varpi}_{[g-1]}^{\langle j \rangle}\|^2 + \|\boldsymbol{\delta}_{m;[g]}^{\langle j \rangle}\|^2 \right).
    \end{aligned}
    $}\hspace{-2mm}
    \end{equation}
    
    Using Cauchy-Schwarz for the telescoping sum, we have $\|\hat{\varpi}_{[g-\varrho_{m;[g]}^{\langle j \rangle}]}^{\langle j \rangle} - \hat{\varpi}_{[g-1]}^{\langle j \rangle}\|^2 \leq \varrho_{\max}^{\langle j \rangle} \sum_{s=g-\varrho_{\max}^{\langle j \rangle}}^{g-1} \|\hat{\varpi}_{[s+1]}^{\langle j \rangle} - \hat{\varpi}_{[s]}^{\langle j \rangle}\|^2$, and 
    combining this with the edge bias bound from Lemma~\ref{lem:edge_bias}, where $\mathbb{E}[\|\boldsymbol{\delta}_{m;[g]}^{\langle j \rangle}\|^2] \leq \frac{4\mathrm{c}_1^2(1-\gamma_{m;[g]}^{\langle j \rangle})^2}{L^2(\gamma_{m;[g]}^{\langle j \rangle})^2} + \mathcal{R}^{'}$, and considering the cumulative effect of $\alpha^{\langle j \rangle} \in [0, 1)$, we obtain \eqref{eq:cloud_async_refined} after simplification.
\end{proof}

\begin{theorem}[Convergence of HEART]\label{thm:nonconvex}
	Under Assumptions~\ref{ass:smoothness}--\ref{ass:momentum}, for task $j$ with total global iterations $G^{\langle j \rangle}$\footnote{i.e., the total number of global iterations required for training task $j$ while satisfying the stability constraints defined in Eq. \eqref{5}.}, the average squared gradient norm satisfies:
	\begin{align}\label{eq:nonconvex_bound}
    \frac{1}{G^{\langle j \rangle}}\sum_{g=0}^{G^{\langle j \rangle}-1} \mathbb{E}\left[\left\|\nabla L^{\langle j \rangle}(\varpi_{[g]}^{\langle j \rangle})\right\|^2\right] \leq \underbrace{\frac{2\Delta_0^{\langle j \rangle}}{\eta_0\sqrt{G^{\langle j \rangle}}}}_{\textcircled{\scriptsize 1}} + \underbrace{\frac{L\eta_0\sigma^2 \ln G^{\langle j \rangle}}{\sqrt{G^{\langle j \rangle}}}}_{\textcircled{\scriptsize 2}} \notag \\
    + \underbrace{\frac{8L^2\varrho_0^2\eta_0^2\mathrm{c}_1^2}{\sqrt{G^{\langle j \rangle}}}}_{\textcircled{\scriptsize 3}} + \underbrace{\frac{8\mathrm{c}_1^2(1-\gamma_{\min})^2}{L^2\gamma_{\min}^2}}_{\textcircled{\scriptsize 4}},
    \end{align}
	where $\Delta_0^{\langle j \rangle}$ is a constant related to the initial Lyapunov value $\mathcal{V}_0^{\langle j \rangle}$ and the lower bound $L_*^{\langle j \rangle}$, $\textcircled{\scriptsize 1}$ represents initial gap attenuation, $\textcircled{\scriptsize 2}$ represents random variance, $\textcircled{\scriptsize 3}$ represents asynchronous cloud penalty, and $\textcircled{\scriptsize 4}$ represents edge dropout deviation. When $G^{\langle j \rangle}$ approaches infinity, the first three terms approach 0, but the fourth term is a constant deviation, indicating that HEART will converge to the domain of the optimal solution rather than the exact optimal solution. In addition, the domain radius is proportional to $\frac{(1-\gamma_{\min})^2}{\gamma_{\min}^2}$.
\end{theorem}

\begin{proof}
	Define the Lyapunov function $\mathcal{V}_{[g]}^{\langle j \rangle} = \mathbb{E}[L^{\langle j \rangle}(\varpi_{[g]}^{\langle j \rangle})] + \frac{L}{2\varrho_0} \sum_{s=g-\varrho_0}^{g-1} (g-s) \mathbb{E}[\|\hat{\varpi}_{[s+1]}^{\langle j \rangle} - \hat{\varpi}_{[s]}^{\langle j \rangle}\|^2]$.
	
	\textbf{Step 1: Single-step descent.} By $L$-smoothness of $L^{\langle j \rangle}(\cdot)$:
	\begin{align}
		L^{\langle j \rangle}(\varpi_{[g+1]}^{\langle j \rangle}) &\leq L^{\langle j \rangle}(\varpi_{[g]}^{\langle j \rangle}) + \langle \nabla L^{\langle j \rangle}(\varpi_{[g]}^{\langle j \rangle}), \varpi_{[g+1]}^{\langle j \rangle} - \varpi_{[g]}^{\langle j \rangle} \rangle \notag \\&+ \frac{L}{2}\|\varpi_{[g+1]}^{\langle j \rangle} - \varpi_{[g]}^{\langle j \rangle}\|^2.
	\end{align}
	Substituting the model update $\varpi_{[g+1]}^{\langle j \rangle} - \varpi_{[g]}^{\langle j \rangle} = -\eta_{[g]}^{\langle j \rangle} \mathbf{g}_{[g]}^{\langle j \rangle} + \boldsymbol{\epsilon}_{[g]}^{Async,\langle j \rangle} + \boldsymbol{\epsilon}_{[g]}^{Drop,\langle j \rangle}$, where $\mathbf{g}_{[g]}^{\langle j \rangle}$ is the stochastic gradient estimate and $\boldsymbol{\epsilon}_{[g]}$ are the error terms defined in Lemma~\ref{lem:cloud_async}.

	\textbf{Step 2: Bounding the inner product.} Using Young's inequality $\langle \mathbf{a}, \mathbf{b} \rangle \leq \frac{\beta}{2}\|\mathbf{a}\|^2 + \frac{1}{2\beta}\|\mathbf{b}\|^2$ with $\beta = \eta_{[g]}^{\langle j \rangle}/2$:
	\begin{align}
		\mathbb{E} \left[ \langle \nabla L^{\langle j \rangle}(\varpi_{[g]}^{\langle j \rangle}), \boldsymbol{\epsilon}_{[g]}^{Drop,\langle j \rangle} \rangle \right] &\leq \frac{\eta_{[g]}^{\langle j \rangle}}{8} \mathbb{E}[\|\nabla L^{\langle j \rangle}(\varpi_{[g]}^{\langle j \rangle})\|^2] \notag \\ &+\frac{2}{\eta_{[g]}^{\langle j \rangle}} \mathbb{E}[\|\boldsymbol{\epsilon}_{[g]}^{Drop,\langle j \rangle}\|^2].
	\end{align}
	Note that the edge bias $\|\boldsymbol{\epsilon}_{[g]}^{Drop}\|^2$ from Lemma~\ref{lem:edge_bias} contains a term of order $\mathcal{O}((\eta_{[g]}^{\langle j \rangle})^2 H^{\langle j \rangle})$. Thus, $\frac{1}{\eta_{[g]}^{\langle j \rangle}} \mathbb{E}[\|\boldsymbol{\epsilon}_{[g]}^{Drop}\|^2]$ effectively yields a term proportional to $\eta_{[g]}^{\langle j \rangle}$ plus a constant bias derived from the participation rate $\gamma_{\min}$.

	\textbf{Step 3: Lyapunov Drift Analysis.} Combining the $L$-smoothness expansion with the drift of the Lyapunov function, and substituting the results from Lemma~\ref{lem:cloud_async}, we obtain:
	\begin{align}
		\mathbb{E}[\mathcal{V}_{[g+1]}^{\langle j \rangle}] - \mathbb{E}[\mathcal{V}_{[g]}^{\langle j \rangle}] &\leq  -\frac{\eta_{[g]}^{\langle j \rangle}}{4} \mathbb{E}\left[\|\nabla L^{\langle j \rangle}(\varpi_{[g]}^{\langle j \rangle})\|^2\right] + \frac{L(\eta_{[g]}^{\langle j \rangle})^2 \sigma^2}{2} \notag \\
		& + 2L^2\varrho_0^2(\eta_{[g]}^{\langle j \rangle})^3 \mathrm{c}_1^2 + \eta_{[g]}^{\langle j \rangle} \cdot \frac{4\mathrm{c}_1^2(1-\gamma_{\min})^2}{L^2\gamma_{\min}^2}.
	\end{align}
	The $\eta_{[g]}^{\langle j \rangle})^3$ term arises from the higher-order interaction between asynchrony and the decaying learning rate.

	\textbf{Step 4: Telescoping sum and Convergence Rate.} Summing from $g=0$ to $G^{\langle j \rangle}-1$:
	\begin{align}
        \sum_{g=0}^{G^{\langle j \rangle}-1} \frac{\eta_{[g]}^{\langle j \rangle}}{4} &\mathbb{E}[\|\nabla L^{\langle j \rangle}(\varpi_{[g]}^{\langle j \rangle})\|^2] \leq \Delta_0^{\langle j \rangle}  + \sum_{g=0}^{G^{\langle j \rangle}-1} \left( \frac{L(\eta^{\langle j \rangle}_{[g]})^2 \sigma^2}{2} \right. \notag \\
        &\left. + 2L^2\varrho_0^2(\eta^{\langle j \rangle}_g)^3 \mathrm{c}_1^2 + \eta^{\langle j \rangle}_{[g]} \frac{4\mathrm{c}_1^2(1-\gamma_{\min})^2}{L^2\gamma_{\min}^2} \right).
    \end{align}
	Dividing both sides by $\sum_{g=0}^{G^{\langle j \rangle}-1} \frac{\eta_{[g]}^{\langle j \rangle}}{4}$ and using the properties of the diminishing step size $\eta_{[g]}^{\langle j \rangle} = \frac{\eta_0}{\sqrt{g+1}}$:
	1) $\sum \eta_{[g]}^{\langle j \rangle} \approx 2\eta_0\sqrt{G^{\langle j \rangle}}$; 
	2) $\sum {(\eta_{[g]}^{\langle j \rangle}})^2 \approx \eta_0^2 \ln G^{\langle j \rangle}$; 
	3) $\sum{(\eta_{[g]}^{\langle j \rangle}})^3 \approx \text{Constant}$. This yields:
	\begin{align}
		\frac{1}{G^{\langle j \rangle}} \sum_{g=0}^{G^{\langle j \rangle}-1} &\mathbb{E}[\|\nabla L(\varpi^{\langle j \rangle}_{[g]})\|^2] \leq \frac{2\Delta_0^{\langle j \rangle}}{\eta_0\sqrt{G^{\langle j \rangle}}} + \frac{L\eta_0\sigma^2 \ln G^{\langle j \rangle}}{\sqrt{G^{\langle j \rangle}}} \notag \\&+ \frac{8L^2\varrho_0^2\eta_0^2 \mathrm{c}_1^2}{\sqrt{G^{\langle j \rangle}}} + \frac{8\mathrm{c}_1^2(1-\gamma_{\min})^2}{L^2\gamma_{\min}^2}.
	\end{align}
	Thus, we have obtained the expected bound in \eqref{eq:nonconvex_bound} and completed the proof.
\end{proof}

\begin{corollary}[Asymptotic Behavior]\label{cor:asymptotic}
As $G^{\langle j \rangle} \to \infty$, the average squared gradient norm satisfies a constant steady-state error:
\begin{align}
	\limsup_{G^{\langle j \rangle} \to \infty} \frac{1}{G^{\langle j \rangle}}\sum_{g=0}^{G^{\langle j \rangle}-1} \mathbb{E}\left[\left\|\nabla L^{\langle j \rangle}(\varpi_{[g]}^{\langle j \rangle})\right\|^2\right] \leq \frac{8\mathrm{c}_1^2(1-\gamma_{\min})^2}{L^2\gamma_{\min}^2}.
\end{align}
\end{corollary}

\begin{remark}
Corollary~\ref{cor:asymptotic} reveals that HEART converges to a neighborhood of stationary points rather than exact stationary points. The neighborhood radius (in terms of squared gradient norm) is proportional to the square of the edge dropout rate $(1-\gamma_{\min})^2$ and inversely proportional to the square of the minimum participation rate $\gamma_{\min}^2$. This constant gap represents the fundamental ``noise floor" introduced by the deadline-based robust edge aggregation mechanism, which persists even if the learning rate decays to zero.
\end{remark}

\subsubsection{Comparison with Fully Synchronous HFL}

We compare HEART with fully synchronous HFL (Sync-HFL), where $Q^{\langle j \rangle} = M$ (all ESs participate) and $\Delta_{\max} = \infty$ (no deadline, waiting for all vehicles).

\begin{table}[htbp]
\centering
\caption{Convergence Comparison: HEART vs. Fully Synchronous HFL}
\label{tab:comparison}
\begin{tabular}{lcc}
\toprule
\textbf{Architecture} & \textbf{Non-convex rate} & \textbf{Asymptotic error} \\
\midrule
Fully Sync-HFL & $\mathcal{O}\left(\frac{1}{\sqrt{G}}\right)$ & $0$ (exact convergence) \\
HEART (This work) & $\mathcal{O}\left(\frac{\varrho_0 \ln G}{\sqrt{G}}\right)$ & $\mathcal{O}\left(\frac{(1-\gamma_{\min})^2}{\gamma_{\min}^2}\right)$ \\
\bottomrule
\end{tabular}
\end{table}

\begin{proposition}[Convergence Rate Degradation]\label{prop:rate_degradation}
Compared to fully synchronous HFL, HEART suffers a multiplicative slowdown factor of $\varrho_0 = M - Q^{\langle j \rangle} + 1$ in the convergence rate due to cloud-level gradient staleness, and an irreducible additive asymptotic error (i.e., noise floor) $\mathcal{O}\left(\frac{\mathrm{c}_1^2(1-\gamma_{\min})^2}{L^2\gamma_{\min}^2}\right)$ due to edge-level vehicle dropout.
\end{proposition}

\begin{proposition}[Wall-Clock Time Trade-off]\label{prop:wall_clock}
Let $T_{\mathrm{sync}}$ and $T_{\mathrm{async}}$ denote the wall-clock time per global iteration for Sync-HFL and HEART, respectively. Under the approximation that $T_{\mathrm{async}} \approx \frac{Q^{\langle j \rangle}}{M}T_{\mathrm{sync}}$ and the iteration complexity ratio is $\frac{G_{\mathrm{async}}}{G_{\mathrm{sync}}} \approx \varrho_0$, the total wall-clock time ratio is:
\begin{align}
	\frac{T_{\mathrm{total}}^{\mathrm{HEART}}}{T_{\mathrm{total}}^{\mathrm{Sync-HFL}}} \approx \frac{Q^{\langle j \rangle}(M-Q^{\langle j \rangle}+1)}{M}.
\end{align}
This ratio is minimized at $Q^{{\langle j \rangle},*} = \lfloor\frac{M+1}{2}\rfloor$, yielding approximately a $\frac{M}{4}$ speedup for large $M$ in terms of wall-clock time to reach the same convergence neighborhood.
\end{proposition}

\begin{remark}[The HEART Trilemma]
Proposition~\ref{prop:wall_clock} reveals a fundamental \textit{trilemma} in HEART design: (i) \textbf{Efficiency}: fast wall-clock convergence requires small $Q^{\langle j \rangle}$; (ii) \textbf{Accuracy}: high final precision requires large $Q^{\langle j \rangle}$ (to reduce staleness $\varrho_0$) and large $\gamma_{\min}$; (iii) \textbf{Robustness}: resilience to stragglers and dynamic channels requires a smaller $\gamma_{\min}$ (tighter deadline).
\end{remark}

\subsubsection{Impact of Deadline Mechanism}

We analyze how the deadline parameter $\Delta_{\max}$ affects the convergence through its influence on $\gamma_{\min}$.

\begin{theorem}[Impact of Fault Tolerance on Participation]\label{thm:deadline_refined}
Let the local completion duration $T_{m,n}^{Dur} = T_{m,n}^{Fin} - T_{m,n}^{Sta}$ follow a distribution with mean $\hat{\mu}$ and variance $\hat{\sigma}^2$. Under the adaptive deadline $\Delta_{\max} = (1 + \hat{\epsilon})\hat{\mu}$, the minimum edge participation rate $\gamma_{\min}$ is lower-bounded by:
\begin{align}
    \gamma_{\min} \geq 1 - \frac{\hat{\sigma}^2}{(\hat{\epsilon} \hat{\mu})^2} - \hat{\epsilon}_{\mathrm{ch}},
\end{align}
where $\hat{\epsilon}$ is the fault tolerance ratio and $\hat{\epsilon}_{\mathrm{ch}}$ is the channel-induced dropout rate.
\end{theorem}

\begin{table*}[!t] 
	\centering
    \renewcommand{\thetable}{VI}
	\caption{Comparison of decision latency and training performance across different methods}
	\label{tab:performance_comparison}
	\small 
	\renewcommand{\arraystretch}{1.1} 
	\begin{tabular}{@{}l ccc c ccc@{}}
		\toprule
		\multirow{2}{*}{\textbf{Method}} & \multicolumn{3}{c}{\textbf{Scenario A: 25 Vehicles}} & & \multicolumn{3}{c}{\textbf{Scenario B: 50 Vehicles}} \\
		\cmidrule(lr){2-4} \cmidrule(lr){6-8}
		& \textbf{DT (s)} & \textbf{TTT (s)} & \textbf{Red. (\%)} & & \textbf{DT (s)} & \textbf{TTT (s)} & \textbf{Red. (\%)} \\
		\midrule
		\textit{TSGD (Greedy)} & 0.435 & 333.654 & -      & & 0.681 & 170.700 & -      \\
		\textit{TSSO (Random)}        & 0.566 & 322.328 & 3.4\%  & & 0.733 & 215.800 & -26.4\% \\
		\textit{TSPSO}         & 1.668 & 359.200 & -7.7\% & & 2.584 & 154.100 & 9.7\%  \\
		\textit{TSGA}          & 1.736 & 310.640 & 6.9\%  & & 2.775 & 191.900 & -12.4\% \\
		\textit{\textbf{HEART (Ours)}} & \textbf{2.019} & \textbf{241.942} & \textbf{27.5\%} & & \textbf{3.135} & \textbf{110.200} & \textbf{35.4\%} \\
		\bottomrule
		\multicolumn{8}{l}{\textit{* DT: Decision Time; TTT: Total Training Time; Red.: Time Reduction vs. Baseline.}}
	\end{tabular}
\end{table*}

\subsubsection{Summary of Theoretical Insights}

Our convergence analysis yields the following key insights:
\begin{enumerate}
    \item \textbf{Cloud Asynchrony Penalty}: Asynchrony introduces a multiplicative penalty $\varrho_{\max}^{\langle j \rangle} = M - Q^{\langle j \rangle} + 1$ on the convergence rate. However, the resulting reduction in per-round waiting time typically outweighs the increase in iteration count.
    
    \item \textbf{Edge Dropout Bias}: The deadline mechanism at edges causes an \textit{irreducible} asymptotic error proportional to $(1-\gamma_{\min})^2/\gamma_{\min}^2$. This is the quantified cost of maintaining system robustness in highly dynamic vehicular environments.
    
    \item \textbf{Learning Rate Adaptation}: Stability under asynchrony requires $\eta_0 \leq \mathcal{O}(1/L\varrho_0)$. This theoretical bound suggests that more aggressive asynchrony (smaller $Q^{\langle j \rangle}$) must be compensated by more conservative learning rates.
    
    \item \textbf{Momentum Weight Constraint}: The weight $\alpha^{\langle j \rangle}$ should be adjusted relative to $\varrho_{\max}^{\langle j \rangle}$. High staleness requires a smaller $(1-\alpha^{\langle j \rangle})$ to dampen the impact of outdated updates.
    
    \item \textbf{Optimal Operating Point}: The theoretical ``sweet spot'' occurs near $Q^{\langle j \rangle} \approx M/2$. This configuration maximizes wall-clock efficiency while keeping the asynchronous penalty within manageable bounds.
\end{enumerate}

\subsection{Complexity Analysis and Decision-Making Latency}\label{sec:Complexity}
In this section, we evaluate the computational efficiency of our proposed \textit{HEART} from both theoretical and empirical perspectives.

As summarized in Table~\ref{tab:time_space_0}, the time and space complexities of \textit{HEART} during the Stage 1 task allocation are comparable to those of \textit{TSPSO} and \textit{TSGA} (where $G^{\langle j\rangle}$ represents the total number of global iterations required for training task $j$ while satisfying the stability constraints defined in Eq. \eqref{5} and $\kappa$ denotes the max attempts). Regarding the complexity of Stage 2, methods \textit{TSSO}, \textit{TSPSO}, \textit{TSPGA}, and \textit{TSPGD} exhibit time and space complexities that align with their Stage 1 counterparts. In contrast, the \textit{HEART} maintains a time complexity of $\mathcal{O}(N \times J\times K^{\langle j\rangle})$ and a space complexity of $\mathcal{O}(N \times J \times K^{\langle j\rangle} \times M)$. Although the hybrid nature of \textit{HEART} may lead to slightly higher practical computational overhead than traditional meta-heuristic baselines, its complexity remains polynomially bounded. This ensures that the framework is computationally feasible for typical IoT and IoV environments with constrained resources.
\begin{table}[!t]
    \centering
    \renewcommand{\thetable}{V}
    \caption{Comparison of computational complexity across different methods in Stage 1}
    \begin{tabular}{lll}
        \hline
        \textbf{Method} & \textbf{Time Complexity} & \textbf{Space Complexity} \\
        \hline
        \textit{TSSO} & $\mathcal{O}(N \times J \times \kappa)$ & $\mathcal{O}(N \times J)$
        \\
        \textit{TSPSO} & $\mathcal{O}(N \times J \times p^* \times \tau^*)$ & $\mathcal{O}(N \times J \times p^* \times M \times G^{\langle j\rangle})$
        \\
        \textit{TSGA} & $\mathcal{O}(N \times J \times p^* \times \tau^*)$ & $\mathcal{O}(N \times J \times p^* \times M \times G^{\langle j\rangle})$
        \\
        \textit{TSGD} & $\mathcal{O}(N \times J \times K^{\langle j\rangle})$ & $\mathcal{O}(N \times J\times K^{\langle j\rangle} \times M)$
        \\
        \textbf{\textit{HEART}} & $\mathcal{O}(N \times J \times p^* \times \tau^*)$ & $\mathcal{O}(N \times J \times p^* \times M \times G^{\langle j\rangle})$ \\
        \hline
    \end{tabular}
    \label{tab:time_space_0} 
\end{table}

To justify the trade-off between decision overhead and training efficiency, we measure the decision-making latency of each scheme using an internal timer and compare it against the Total Training Time (TTT). As illustrated in Table~\ref{tab:performance_comparison}, while the decision latency of \textit{HEART} is approximately 1.58--2.45s higher than that of the lightweight \textit{TSGD} baseline in the 25- and 50-vehicle scenarios, respectively. However, it remains within the order of seconds (e.g., 3.135 seconds for 50 vehicles). Compared to the hundreds of seconds required for the entire global training process, such overhead is negligible. More importantly, \textit{HEART} achieves a 35.4\% reduction in TTT (equivalent to a saving of 60.5 seconds). Moreover, the joint optimization shown in Fig. 7 maintains vehicle non-task-training time at a lower level throughout the training process, significantly enhancing overall efficiency. This demonstrates that the superior solution quality provided by Stage 1 effectively avoids local optima, leading to a significantly more efficient execution phase that far outweighs the initial decision-making overhead. Integrating an adaptive mechanism to switch \textit{HEART} to lightweight solutions (e.g., \textit{TSGD}) in simpler scenarios is a promising future direction.

\end{document}